\documentclass[10pt,twocolumn,letterpaper]{article}

\usepackage{cvpr}
\usepackage{times}
\usepackage{epsfig}
\usepackage{graphicx}
\usepackage{amsmath}
\usepackage{amssymb}
\usepackage{csquotes}
\usepackage{algorithm}
\usepackage{algorithmic}
\usepackage[british]{babel}
\usepackage{multirow}
\usepackage{color}
\usepackage{xcolor}
\usepackage{bm}
\usepackage{xcolor}
\usepackage{subcaption}
\usepackage{booktabs} 

\DeclareMathOperator*{\argmin}{argmin}
\DeclareMathOperator*{\argmax}{argmax}
\usepackage[toc,page]{appendix}

\newenvironment{packed_enum}{
\begin{enumerate}
  \setlength{\itemsep}{1pt}
  \setlength{\parskip}{0pt}
  \setlength{\parsep}{0pt}
}{\end{enumerate}}

\usepackage[pagebackref=true,breaklinks=true,letterpaper=true,colorlinks,bookmarks=false]{hyperref}

\cvprfinalcopy 

\def\cvprPaperID{958} 

\begin{document}

\title{{Joint Unsupervised Learning of Deep Representations and Image Clusters}}

\author{Jianwei Yang, Devi Parikh, Dhruv Batra\\
Virginia Tech\\
{\tt\small \{jw2yang, parikh, dbatra\}@vt.edu}
}

\maketitle

\begin{abstract}
In this paper, we propose a recurrent framework for \textbf{J}oint \textbf{U}nsupervised \textbf{LE}arning (\textbf{JULE}) of deep representations and image clusters. In our framework, {successive operations in a clustering algorithm} are expressed as \emph{{steps} in {a recurrent process}}, stacked on top of representations output by a Convolutional Neural Network (CNN). {During training, image clusters and representations are updated jointly: image clustering is conducted in the forward pass, while representation learning in the backward pass. Our key idea behind this framework is that good representations are beneficial to image clustering and clustering results provide supervisory {signals} to representation learning.} By integrating two processes into a single model with a unified {weighted triplet} loss and optimizing it end-to-end, we can obtain not only more powerful representations, but also more precise image clusters. Extensive experiments show that our method {outperforms} the state-of-the-art on image clustering across a variety of image datasets. Moreover, the learned representations generalize well when transferred to other tasks. {The source code can be downloaded from \url{https://github.com/jwyang/joint-unsupervised-learning}}.
\end{abstract}

\section{Introduction}
\label{Sec_Introduction}
We are witnessing an explosion in visual content. Significant recent advances in machine learning and computer vision, especially via deep neural networks, have relied on supervised learning and availability of copious annotated data. However, manually labelling data is a time-consuming, laborious, and often expensive process. In order to make better use of available unlabeled images, clustering and/or unsupervised learning is a promising direction.

In this work, we aim to {address} image clustering and representation learning on unlabeled images in a {unified} framework. {It is a natural idea to leverage cluster ids of images as supervisory signals to learn representations and in turn the representations would be beneficial to image clustering.} At a high-level view, given a collection of $n_s$ unlabeled images $\bm{I} = \{I_1, ... ,I_{n_s}\}$, the global objective function {for learning image representations and clusters} can be written as:
\begin{equation}
\small
\argmin_{\bm{y}, \bm{\theta}} \mathcal{L}(\bm{y, \bm{\mathcal{\theta}}} | \bm{I})
\label{F_General_Objective_Function}
\end{equation}
where $\mathcal{L}(\cdot)$ is a loss function, $\bm{y}$ denotes the cluster ids for all images, and $\bm{\theta}$ denotes the parameters for representations. If we hold one in $\{\bm{y}, \bm{\theta}\}$ to be fixed, the optimization can be decomposed into two alternating steps:
\begin{subequations}
\small
\begin{equation}
\argmin_{\bm{y}} \mathcal{L}(\bm{y} | \bm{I}, \bm{\theta})
\label{F_Decomposed_General_Objective_FunctionA}
\end{equation}
\begin{equation}
\argmin_{\bm{\theta}} \mathcal{L}(\bm{\theta} | \bm{I}, \bm{y})
\label{F_Decomposed_General_Objective_FunctionB}
\end{equation}
\end{subequations}

\begin{figure}[t]
   \begin{subfigure}{0.32\linewidth}
   \centering
    \includegraphics[trim=6cm 5cm 5cm 4cm, clip=true, scale=0.15]{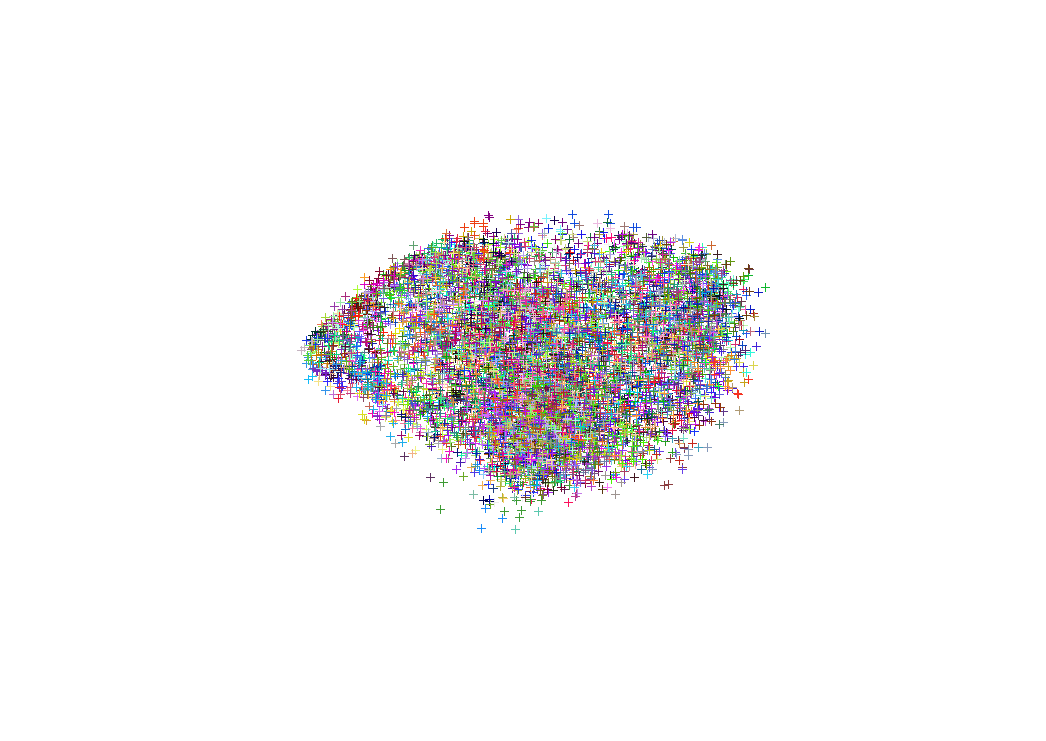}
    \vspace{-5pt}
    \caption{Initial stage}
   \end{subfigure}
   \begin{subfigure}{0.32\linewidth}
   \centering
    \includegraphics[trim=6cm 5cm 5cm 4cm, clip=true, scale=0.15]{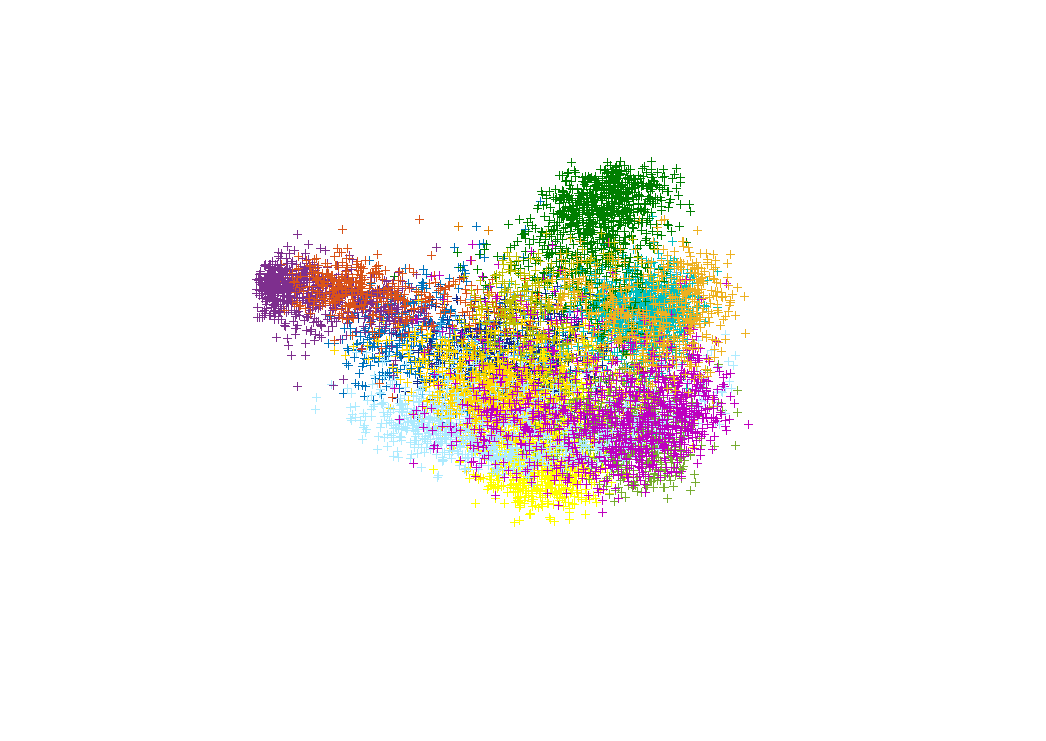}
    \vspace{-5pt}
    \caption{Middle stage}
   \end{subfigure}
   \begin{subfigure}{0.32\linewidth}
   \centering
       \includegraphics[trim=6cm 5cm 5cm 4cm, clip=true, scale=0.15]{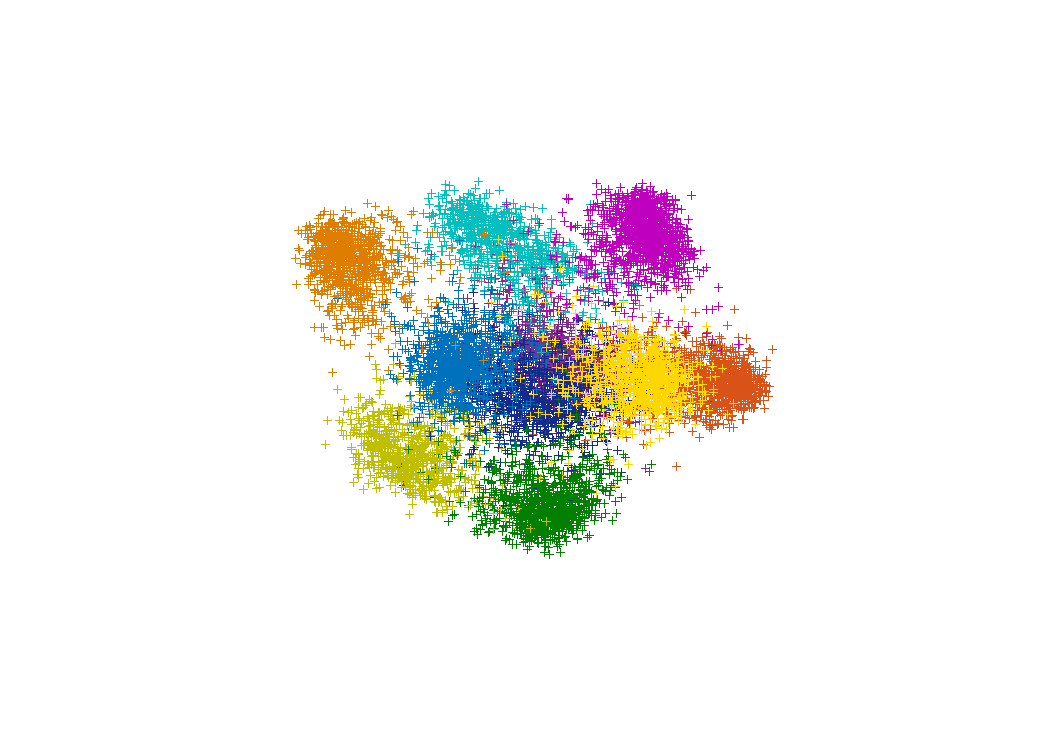}
   \vspace{-5pt}
   \caption{Final stage}
   \end{subfigure}
   \vspace{-5pt}
   \caption{{Clustering outputs for MNIST \cite{lecun1998gradient} test set at different stages of the proposed method. We conduct PCA on the {image} representations and then choose the first three dimensions for {visualization}. Different colors correspond to different clusters. Samples are grouped together gradually and more discriminative representations are obtained.}}
   \label{Fig_Introduction}
\end{figure}
{Intuitively}, \eqref{F_Decomposed_General_Objective_FunctionA} can be cast as a {conventional} clustering problem based on fixed representations, while \eqref{F_Decomposed_General_Objective_FunctionB} is a standard supervised representation learning process. 

{In this paper, we propose an approach that alternates between the two steps -- updating the cluster ids given the current representation parameters and updating the representation parameters given the current clustering result. Specifically, we cluster images using agglomerative clustering\cite{gowda1978agglomerative} and represent images via activations of a Convolutional Neural Network (CNN).}

{The reason to choose agglomerative clustering is three-fold: 1) it begins with an over-clustering, which is more reliable in the beginning when a good representation has not yet been learned. Intuitively, clustering with representations from a CNN initialized with random weights are not reliable, but nearest neighbors and over-clusterings are often acceptable; 2) These over-clusterings can be merged as better representations are learned; 3) Agglomerative clustering is a recurrent process and can naturally be interpreted in a recurrent framework.}

{Our final algorithm is farily intuitive. We start with an intial over-clustering, update CNN parameters (2b) using image cluster labels as supervisory signals, then merge clusters (2a) and iterate until we reach a stopping criterion.} An outcome of the proposed framework is illustrated in Fig.~\ref{Fig_Introduction}. Initially, there are 1,762 clusters for MNIST test set (10k samples), and the representations (image intensities) are not that discriminative. After several iterations, we obtain 17 clusters and more discriminative representations. Finally, we obtain 10 clusters which are well-separated by the learned representations and {interestingly correspond primarily to the groundtruth category labels in the dataset, even though the representation is learnt in an unsupervised manner}. To summarize, the major contributions of our work are:
\begin{packed_enum}
\item[1] We propose a {simple but effective} end-to-end learning framework to {jointly learn deep representations and image clusters} {from an} unlabeled image set;
\item[2] We formulate the joint learning in a {recurrent framework}, where {merging} operations of {agglomerative} clustering are expressed as a forward pass, {and} representation learning of CNN as a backward pass;
\item[3] We derive \emph{a single loss function} to guide agglomerative clustering and deep representation learning, {which makes optimization {over the} two tasks seamless};
\item[4] Our experimental results show that the proposed framework outperforms previous methods on image clustering and learns deep representations that can be transferred to other tasks and datasets.
\end{packed_enum}

\section{Related Work}
\vspace{-5pt}
\noindent \textbf{Clustering}
Clustering algorithms can be {broadly} categorized into hierarchical and {partitional} approaches \cite{jain1999data}. {Agglomerative clustering is a hierarchical clustering algorithm that begins with many {small} clusters, and then merges clusters gradually \cite{gowda1978agglomerative,kurita1991efficient, gdalyahu2001self}}. As for partitional clustering methods, the most well-known is K-means \cite{macqueen1967some}, which minimizes the sum {of} square errors between data points and their nearest cluster centers. {Related ideas form the basis of a number of methods, such as expectation maximization (EM) \cite{dempster1977maximum, mclachlan2004finite}, spectral clustering \cite{ng2002spectral, zelnik2004self, shi2000normalized}, {and} non-negative matrix factorization (NMF) based clustering \cite{ding2010convex, cai2009locality, zafeiriou2010nonlinear}.}

\noindent \textbf{Deep Representation Learning}
Many works {use} raw image intensity or hand-crafted features \cite{singh2012unsupervised,doersch2013mid,hariharan2012discriminative,han2015unsupervised,rematas2015dataset,huang2012affinity} combined with conventional clustering methods. Recently, representations learned using deep neural networks have presented {significant improvements over} {hand-designed} features on many computer vision tasks, such as image classification \cite{krizhevsky2012imagenet,sermanet2013overfeat,simonyan2014very,russakovsky2014imagenet}, object detection \cite{girshick2014rich,girshick2015fast,he2014spatial,ren2015faster}, etc. However, these approaches rely on {supervised learning with large amounts of labeled data to learn rich representations}. {A number of works have} focused on learning representations from unlabled image data. One class of approaches cater to reconstruction tasks, such as auto-encoders \cite{ranzato2007unsupervised, hinton2006reducing, vincent2008extracting, krizhevsky2011using, lee2009convolutional}, deep belief networks (DBN) \cite{le2013building}, etc. Another group of techniques learn discriminative representations after fabricating supervisory signals for images, and then finetune them supervisedly for downstream applications \cite{dosovitskiy2014discriminative, doersch2015unsupervised, wang2015unsupervised}. {Unlike our approach, the fabricated supervisory signal in these previous works is not} updated during representation learning.

\noindent \textbf{Combination}
A number of works have explored combining image clustering with representation learning. In \cite{tian2014learning}, the authors proposed to learn a non-linear embedding of the undirected affinity graph using stacked autoencoder, and then ran K-means in the embedding space to obtain clusters. In \cite{trigeorgis2014deep}, a deep semi-NMF model was used to factorize the input into multiple stacking factors which are initialized and updated layer by layer. Using the representations on the top layer, K-means was implemented to get the final results. Unlike our work, they do not jointly optimize for the representation learning and clustering.

To connect image clustering and representation learning more closely, \cite{xie2015integrating} conducted image clustering and codebook learning iteratively. However, they learned codebook over SIFT feature \cite{lowe1999object}, {and did not learn deep representations}. Instead of using hand-crafted features, Chen \cite{chen2015deep} used DBN to learn representations, and then conducted a nonparametric maximum margin clustering upon the outputs of DBN. {Afterwards, they fine-tuned the top layer of DBN based on clustering results. A {more} recent work on jointly optimizing two tasks is found in \cite{wang2015learning}, where the authors trained a task-specific deep architecture for clustering. The deep architecture is composed of sparse coding modules which can be jointly trained through back propagation from a cluster-oriented loss. However, {they used sparse coding to extract representations for images, while we use a CNN. Instead of fixing the number of clusters to be the number of categories and predicted labels based on softmax outputs, we predict the labels using agglomerative clustering based on the learned representations. In our experiments we show that our approach outperforms \cite{wang2015learning}}.

\section{Approach}
\vspace{-5pt}
\subsection{Notation}
\begin{figure}[t]
   \begin{minipage}{1.0\linewidth}
   \centering
   \epsfig{file=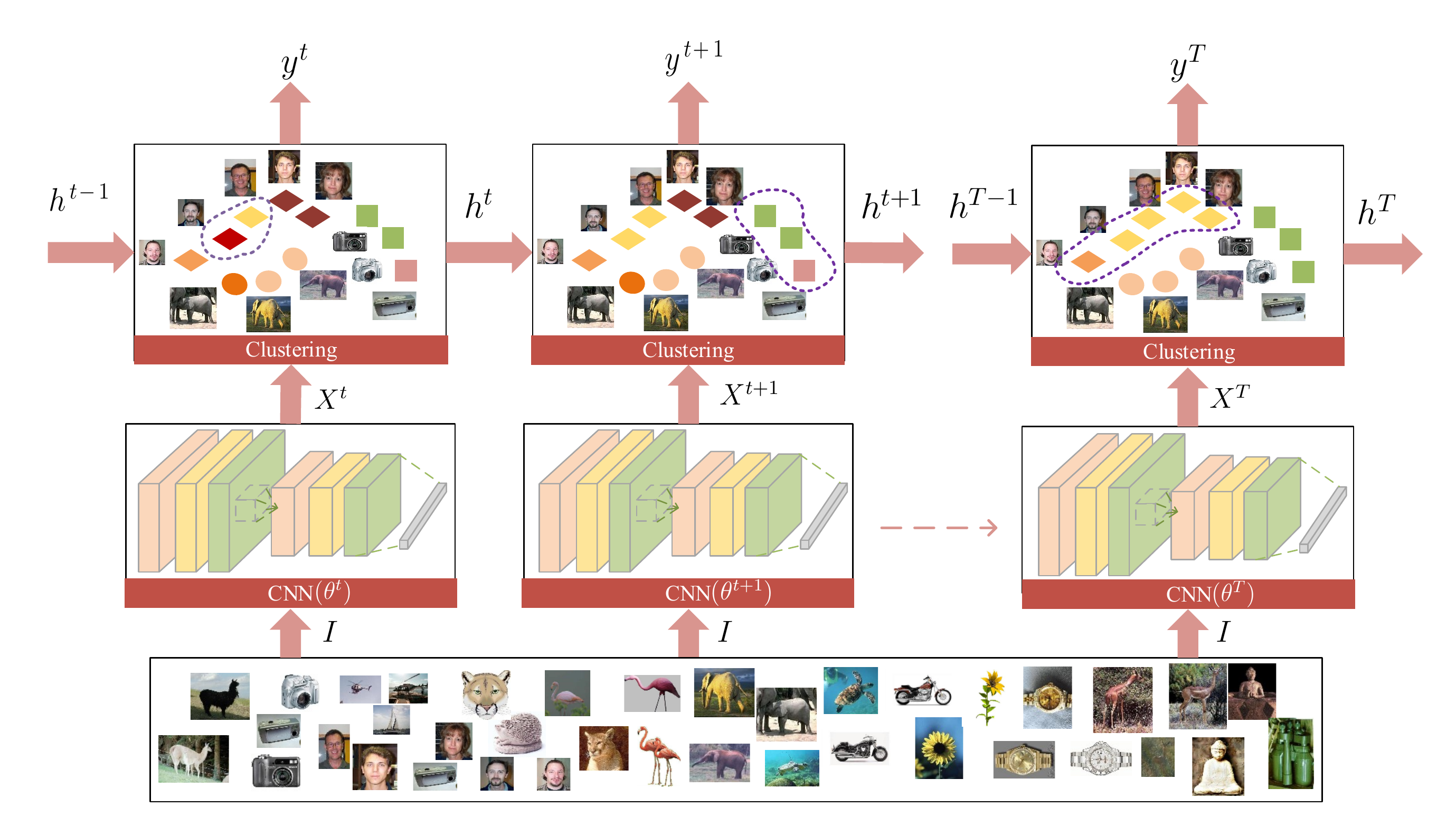, scale=0.3}
   \end{minipage}
   \vspace{-10pt}
   \caption{Proposed recurrent framework for unsupervised learning of deep representations and image clusters.}
   \label{Fig_Framework}
\end{figure}
We denote an image set with $n_s$ images by $\bm{I} = \{I_1,...,I_{n_s}\}$. {The cluster labels for this image set are $\bm{y} = \{y_1,...,y_{n_s}\}$.} $\bm{\theta}$ are the CNN parameters, based on which we obtain deep representations $\bm{X} = \{\bm{x}_1,...,\bm{x}_{n_s}\}$ from $\bm{I}$. Given the predicted {image cluster labels}, we organize them into $n_c$ clusters $\bm{\mathcal{C}} = \{\mathcal{C}_1, ..., \mathcal{C}_{n_c}\}$, {where $\mathcal{C}_i = \{\bm{x}_k | y_k = i,\ \forall k \in {1,...,n_s}\}$}. {$\mathcal{N}_i^{K_s}$ are the $K_s$ nearest neighbours of $\bm{x}_i$, and $\mathcal{N}_{\mathcal{C}_i}^{K_c}$ is the set of $K_c$ nearest neighbour clusters of $\mathcal{C}_i$. For convenience, we sort clusters in $\mathcal{N}_{\mathcal{C}_i}^{K_c}$ in descending order of affinity with $\mathcal{C}_i$ so that the nearest neighbour $\argmax_{C \in \bm{\mathcal{C}^t}}{\bm{\mathcal{A}}}(\mathcal{C}_i, \mathcal{C})$ is the first entry $\mathcal{N}_{\mathcal{C}_i}^{K_c}[1]$. Here, $\mathcal{A}$ is a function to measure the affinity (or similarity) between two clusters. We add a superscript $t$ to \{$\bm{\theta}$, $\bm{X}$, $\bm{y}$, $\bm{\mathcal{C}}$\} to refer to their states at timestep $t$. We use $\bm{\mathcal{Y}}$ to denote the {sequence} $\{\bm{y}^1, ...,\bm{y}^T\}$ with $T$ timesteps.

\subsection{Agglomerative Clustering}
As background, we first briefly describe conventional agglomerative clustering \cite{gowda1978agglomerative,kurita1991efficient}. The core idea in agglomerative clustering is {to merge} two clusters at each step until some stopping conditions. Mathematically, it tries to find two clusters $\mathcal{C}_a$ and $\mathcal{C}_b$ by
\begin{equation}
\small
\{\mathcal{C}_a, \mathcal{C}_b\} = \argmax _{\mathcal{C}_i, \mathcal{C}_j \in \bm{\mathcal{C}}, i \neq j} \bm{\mathcal{A}}(\mathcal{C}_i, \mathcal{C}_j)
\label{Eq_AgglClustering_Canonical}
\end{equation}

There are many methods to compute the affinity between two clusters \cite{gowda1978agglomerative, kurita1991efficient,navarro1997universal,zhao2009cyclizing,zhang2012graph}. More details can be found in \cite{jain1999data}. We now describe how the affinity is measured by $\mathcal{A}$ in our approach.

\subsection{Affinity Measure}
{First}, we build a directed graph {$G = <\mathcal{V}, \mathcal{E}>$}, where $\mathcal{V}$ is the set of vertices corresponding to {deep representations $\bm{X}$ for $\bm{I}$}, and $\mathcal{E}$ is the set of edges connecting vertices. We define an affinity matrix ${\bm{W}} \in \mathbb{R}^{n_s \times n_s}$ {corresponding to the edge set}. The weight from vertex $\bm{x}_i$ to $\bm{x}_j$ is defined by
\begin{equation}
\bm{W}(i, j)=
\begin{cases}
    exp(-\frac{||\bm{x}_i-\bm{x}_j||_2^2}{\sigma^2}), & \text{if } \bm{x}_j \in \mathcal{N}_i^{K_s} \\ 
    0,              & \text{otherwise}
\end{cases}
\label{Eq_Sample_Affinity}
\end{equation}
where $\sigma^2 = \frac{a}{n_sK_s} \sum_{\bm{x}_i \in \bm{X}} \sum_{\bm{x}_j \in \mathcal{N}_i^{K_s}} ||\bm{x}_i - \bm{x}_j||_2^2$. This way to build up a directed graph can be found in many previous works {such as} \cite{zhao2009cyclizing,zhang2012graph}. Here, $a$ and $K_s$ are two predefined parameters {(their values are listed in Table \ref{TB_AC_Param})}. After constructing a directed graph for samples, we then adopt the graph degree linkage in \cite{zhang2012graph} to measure the affinity {between cluster $\mathcal{C}_i$ and $\mathcal{C}_j$, denoted by $\mathcal{A}(\mathcal{C}_i, \mathcal{C}_j)$.}

\subsection{A Recurrent Framework}
{Our key insight is that agglomerative} clustering can be interpreted as a recurrent process in {the} sense that it merges clusters over multiple timesteps. Based on this insight, we propose a recurrent framework to combine the image clustering and representation learning processes. 

As shown in Fig.~\ref{Fig_Framework}, at the timestep $t$, images $\bm{I}$ are first fed into the CNN to get representations $\bm{X}^t$ and then used in conjunction with previous hidden state $\bm{h}^{t - 1}$ to predict current hidden state $\bm{h}^t$, i.e, the image cluster labels at timestep $t$. {In our context, the output at timestep $t$ is $\bm{y}^t = \bm{h}^t$}. {Hence, at timestep $t$}
\begin{subequations}
\begin{equation}
\small
\bm{X}^t = f_r(\bm{I} | \bm{\theta}^t)
\label{Eq_Recurrent_X}
\end{equation}
\begin{equation}
\small
\bm{h}^t = f_m(\bm{X}^t, \bm{h}^{t-1})
\label{Eq_Recurrent_h_t}
\end{equation}
\begin{equation}
\small
\bm{y}^t = f_o(\bm{h}^t) = \bm{h}^t
\label{Eq_Recurrent_y_t}
\end{equation}
\end{subequations}
where $f_r$ is a function to extract deep representations $\bm{X}^t$ for input $\bm{I}$ {using the CNN parameterized by $\bm{\theta}^t$}, and {$f_m$ is a merging process for generating $\bm{h}^{t}$ based on $\bm{X}^t$ and $\bm{h}^{t-1}$}. 

{In a typical Recurrent Neural Network, {one would unroll all timesteps at each training iteration}. {In our case, that would involve performing agglomerative clustering} until we obtain the desired number of clusters, and then update the CNN parameters by back-propagation.}

{In this work, we introduce a \emph{partial unrolling} strategy, \ie, we split the overall $T$ timesteps into multiple periods, and unroll one period at a time. The intuitive reason we unroll partially is that the representation of the CNN at the beginning is not reliable. We need to update CNN parameters to obtain more discriminative representations for {the} following merging processes. In each period, we merge a number of clusters and update CNN parameters for a fixed number of iterations at the end of the period. An extreme case would be one timestep per period, but it involves updating the CNN parameters too frequently and is thus time-consuming. Therefore, the number of timesteps per period (and thus the number of clusters merged per period) is determined by a parameter in our approach. We elaborate on this more in Sec.~\ref{Sec_Optimization}}.

\subsection{Objective Function}
In our recurrent framework, we accumulate the losses from all timesteps, which is formulated as
\begin{equation}
\small
\mathcal{L}(\{\bm{y}^1, ...,\bm{y}^T\}, \{\bm{\theta}^1,...,\bm{\theta}^T\} | \bm{I}) = \sum_{t = 1}^{T} \mathcal{L}^t(\bm{y}^t, \bm{\theta}^t | \bm{y}^{t-1}, \bm{I})
\label{Eq_Loss_Overall}
\end{equation}

Here, $\bm{y}^0$ takes each image as a cluster. At timestep $t$, we find two clusters to merge given $\bm{y}^{t-1}$. In conventional agglomerative clustering, the two clusters are determined by finding the maximal affinity over all pairs of clusters. In this paper, we introduce a criterion that considers not only the affinity between two clusters but also the local structure surrounding the clusters. Assume from $\bm{y}^{t-1}$ to $\bm{y}^t$, we merged a cluster $\mathcal{C}^t_i$ and its nearest neighbour. Then the loss at timestep $t$ is a combination of negative affinities, that is,
\begin{subequations}
\begin{equation}
\begin{aligned}
\mathcal{L}^t(\bm{y}^t, \bm{\theta}^t |\bm{y}^{t-1}, \bm{I}) = -\bm{\mathcal{A}}(\mathcal{C}^t_i,  \mathcal{N}_{\mathcal{C}^t_i}^{K_c}[1]) &  &  &  & &&&&&&&&&&
\end{aligned}
\label{Eq_similarity_term_a_timestep_t}
\end{equation}
\begin{equation}
\begin{aligned}
&&&& - \frac{\lambda}{(K_c - 1)} \sum_{k = 2}^{K_c} \left(\bm{\mathcal{A}}(\mathcal{C}^t_i,  \mathcal{N}_{\mathcal{C}^t_i}^{K_c}[1]) - \bm{\mathcal{A}}(\mathcal{C}^t_i,  \mathcal{N}_{\mathcal{C}^t_i}^{K_c}[k])\right)
\end{aligned}
\label{Eq_similarity_term_b_timestep_t}
\end{equation}
\end{subequations}
where $\lambda$ weighs \eqref{Eq_similarity_term_a_timestep_t} and \eqref{Eq_similarity_term_b_timestep_t}. Note that $\bm{y}^t$, $\bm{y}^{t-1}$ and $\bm{\theta}^t$ are not explicitly presented at the right side, but they {determine} the loss via the image cluster labels and affinities among clusters. {On the right side of the above equation}, there are two terms: 1) \eqref{Eq_similarity_term_a_timestep_t} measures the affinity between cluster $\mathcal{C}_i$ and its nearest neighbour, which follows conventional agglomerative clustering; 2) \eqref{Eq_similarity_term_b_timestep_t} measures the {difference} between affinity of $\mathcal{C}_i$ to its nearest neighbour cluster and affinities of $\mathcal{C}_i$ to its other neighbour clusters. This term takes the local structure into account. See Sec.~\ref{Sec_Forward_Pass} for detailed explanation.

It is hard to simultaneously derive the optimal $\{\bm{y}^1, ...,\bm{y}^T\}$ and $\{\bm{\theta}^1,...,\bm{\theta}^T\}$ that minimize the overall loss in Eq.~\eqref{Eq_Loss_Overall}. As aforementioned, we optimize iteratively in a recurrent process. We divide $T$ timesteps into $P$ partially unrolled periods. In each period,  we fix $\bm{\theta}$ and search optimal $\bm{y}$ in the forward pass, and then in the backward pass we derive optimal $\bm{\theta}$ given the optimal $\bm{y}$. Details will be explained in the following sections.

\subsubsection{Forward Pass}
\label{Sec_Forward_Pass}
{In forward pass of the $p$-th ($p \in \{1,...,P\}$) partially unrolled period}, we update the cluster labels {with $\bm{\theta}$ fixed to $\bm{\theta}^p$}, and the overall loss in period $p$ is
\begin{equation}
\small
\mathcal{L}^p(\bm{\mathcal{Y}}^p | \bm{\theta}^p, \bm{I}) = \sum_{t = t_p^s}^{t_p^e} \mathcal{L}^t(\bm{y}^t | \bm{\theta}^p, \bm{y}^{t-1}, \bm{I})
\label{Eq_Loss_Overall_wst_label}
\end{equation}
where $\bm{\mathcal{Y}}^p$ is the sequence of image labels in period $p$, and $[t_p^s, t_p^e]$ is the corresponding timesteps in period $p$.} For optimization, we follow a greedy search {similar to} conventional agglomerative clustering. {Starting from the time step $t_p^s$}, it finds one cluster and its nearest neighbour to merge so that $\mathcal{L}^t$ is minimized over all possible cluster pairs.

In Fig.~\ref{Fig_Approach_ToyExample}, we present a toy example to explain the reason why we employ the term \eqref{Eq_similarity_term_b_timestep_t}. As shown, it is often the case that {the clusters are densely populated {in some regions} while sparse in some other regions}. In conventional agglomerative clustering, it will choose two clusters with largest affinity (or smallest loss) at each time no mater where the clusters are located. {In this specific case}, it will choose cluster $\mathcal{C}_b$ and its nearest neighbour to merge. In contrast, as shown in Fig.~\ref{Fig_Approach_ToyExample}(b), our algorithm by adding \eqref{Eq_similarity_term_b_timestep_t} will find cluster $\mathcal{C}_e$, because it is not only close to it nearest neighbour, but also relatively far away from its other neighbours, i.e., the local structure is considered around one cluster. {Another merit of introducing \eqref{Eq_similarity_term_b_timestep_t} is that it will allow us to write the loss in terms of triplets as explained next.}

\subsubsection{Backward Pass}
\begin{figure}[t]
   \begin{subfigure}{0.48\linewidth}
   \centering
    \includegraphics[trim=1cm 1cm 1cm 1cm, clip=true, scale=0.5]{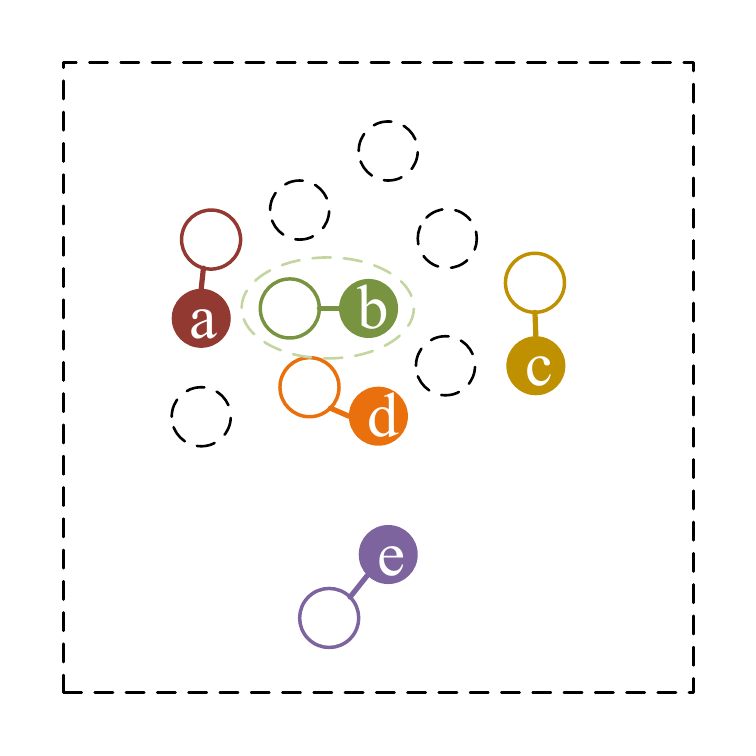}
    \caption{}
   \end{subfigure}
   \begin{subfigure}{0.48\linewidth}
   \centering
    \includegraphics[trim=1cm 1cm 1cm 1cm, clip=true, scale=0.5]{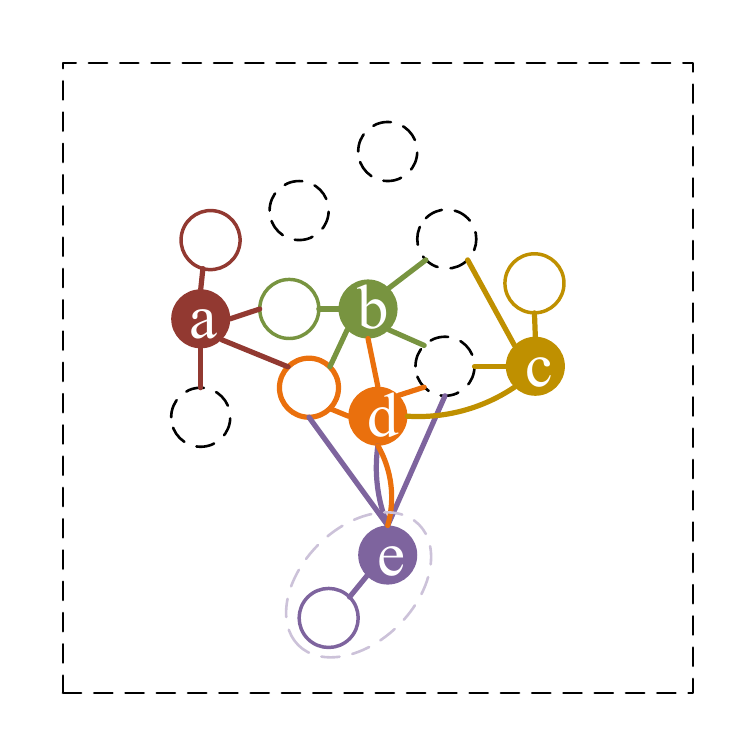}
    \caption{}
   \end{subfigure}
   \caption{A toy illustration of (a) conventional agglomerative clustering strategy and (b) the proposed one. For simplification, we use a single circle to represent a cluster/sample. In conventional agglomerative clustering, node \textit{b} and its nearest neighbour are chosen to merge because they are closest to each other; while node \textit{e} is chosen {in our proposed strategy} considering the local structure.} 
\label{Fig_Approach_ToyExample}
\end{figure}
{In forward pass of the $p$-th partially unrolled period, we have merged a number of clusters. {Let the sequence of optimal image cluster labels be given by} $\bm{\mathcal{Y}}_*^p = \{\bm{y}_*^{t}\}$, and clusters merged in forward pass are denoted by $\{[\mathcal{C}^t_*, \mathcal{N}_{{\mathcal{{C}}}^t_*}^{K_c}[1]]\}$, $t \in \{t_p^s,...,t_p^e\}$.} In the backward pass, we aim to derive the optimal $\bm{\theta}$ to minimize the losses generated in forward pass. Because the clustering in current period is conditioned on the clustering results of all previous periods, {we {accumulate} the losses of all $p$ periods}}, i.e.,
\begin{equation}
\small
\mathcal{L}(\bm{\theta} | \{\bm{\mathcal{Y}}_*^1,...,\bm{\mathcal{Y}}_*^p\}, \bm{I}) = \sum_{k = 1}^{p} \mathcal{L}^k(\bm{\theta} | \bm{\mathcal{Y}}_*^k,  \bm{I})
\label{Eq_Loss_Overall_wst_theta}
\end{equation}

Minimizing \eqref{Eq_Loss_Overall_wst_theta} w.r.t $\bm{\theta}$ leads to representation learning on $\bm{I}$ supervised by $\{\bm{\mathcal{Y}}_*^1,...,\bm{\mathcal{Y}}_*^p\}$ or $\{\bm{y}_*^1, ...,\bm{y}_*^{t_p^e}\}$. Based on \eqref{Eq_similarity_term_a_timestep_t} and \eqref{Eq_similarity_term_b_timestep_t}, {the loss in Eq.~\ref{Eq_Loss_Overall_wst_theta} is reformulated to}
\begin{equation}
\small
\begin{aligned}
-\frac{\lambda}{K_c - 1} \sum_{t = 1}^{t_p^e} \sum_{k = 2}^{K_c} \left(\lambda' \bm{\mathcal{A}}(\mathcal{C}^t_*, \mathcal{N}_{\mathcal{C}^t_*}^{K_c}[1]) - \bm{\mathcal{A}}(\mathcal{C}^t_*, \mathcal{N}_{\mathcal{C}^t_*}^{K_c}[k])\right)
\end{aligned}
\label{Eq_Loss_triplet_time_step}
\end{equation}
where $\lambda' = (1 +1 / \lambda)$. {\eqref{Eq_Loss_triplet_time_step} is a loss defined on clusters of points, which needs the entire dataset to estimate, making it difficult to use batch-based optimization.} However, we show that this loss can be approximated by a sample-based loss, enabling us to compute unbiased estimators for the gradients using batch-statistics. 

The intuition behind reformulation of the loss is that agglomerative clustering starts with each datapoint as a cluster, and clusters at a higher level in the hierarchy are formed by merging lower level clusters. Thus, affinities between clusters can be expressed in terms of affinities between datapoints. We show in the supplement that the loss in \eqref{Eq_Loss_triplet_time_step} can be approximately reformulated as
\begin{equation}
\small
\mathcal{L}(\bm{\theta} |\bm{y}^{t_p^e}_*, \bm{I}) = -\frac{\lambda}{K_c - 1} \sum_{i,j,k}  \left(\gamma \bm{\mathcal{A}}(\bm{x}_i, \bm{x}_j)  - \bm{\mathcal{A}}(\bm{x}_i, \bm{x}_k)\right)
\label{Eq_Loss_triplet_wst_theta_sample}
\end{equation}
{where $\gamma$ is a weight whose value depends on $\lambda'$ and how clusters are merged during the forward pass.} $\bm{x}_i$ and $\bm{x}_j$ are from the same cluster, while $\bm{x}_k$ is from the neighbouring clusters, {and their cluster labels are merely determined by the final clustering result $\bm{y}_*^{t_p^e}$. To further simplify the optimization, {we instead search $\bm{x}_k$ in at most $K_c$ neighbour samples of $x_i$ from other clusters in a training batch.}} Hence, the batch-wise optimization can be performed using conventional stochastic gradient {descent} method. {Note that such triplet losses have appeared in other works \cite{wang2014learning, schroff2015facenet}.} Because it is associated with a weight, we call \eqref{Eq_Loss_triplet_wst_theta_sample} the weighted triplet loss.

\subsection{Optimization}
\label{Sec_Optimization}
\begin{algorithm}[t]
\caption{Joint Optimization on $\bm{y}$ and $\bm{\theta}$}
\begin{algorithmic}[1]
\REQUIRE ~~\\
      $\bm{I}$: = collection of image data; \\
      $n^*_c$: = target number of clusters; \\
\ENSURE ~~\\
      $\bm{y}^*, \bm{\mathcal{\theta}}^*$: = final image labels and CNN parameters;\\
\STATE $t \gets 0$; $p \gets 0$
\STATE Initialize $\bm{\theta}$ and $\bm{y}$
\REPEAT
\STATE Update $\bm{y}^t$ to $\bm{y}^{t+1}$ by merging two clusters
\IF{$t=t_p^e$} 
\STATE {Update $\bm{\theta}^p$ to $\bm{\theta}^{p+1}$ by training CNN}
\STATE $p \gets (p + 1)$
\ENDIF
\STATE $t \gets t + 1$
\UNTIL Cluster number reaches $n_c^*$
\STATE $\bm{y}^* \gets \bm{y}^{t}$; $\bm{{\theta}}^* \gets \bm{{\theta}}^{p}$
\end{algorithmic}
\label{Alg_Optimization}
\end{algorithm}
{{Given an image dataset with $n_s$ samples, we assume the number of desired clusters $n_c^*$ is given to us {as is standard in clustering}. Then we can build up a recurrent process with  $T = n_s - n_c^*$ timesteps, starting by regarding each sample as a cluster}. However, such initialization makes the optimization time-consuming, especially when datasets contain a large number of samples. To address this problem, we can first run a fast clustering algorithm to get the initial clusters. Here, we adopt the initialization algorithm proposed in \cite{zhang2013agglomerative} for fair comparison with their experiment results. Note that other kind of initializations {can also be used}, e.g. K-means. {Based on the algorithm in \cite{zhang2013agglomerative}}, we obtain a number of clusters which contain a few samples for each (average is about 4 in our experiments). {Given these initial clusters, our optimization algorithm learns deep representations and clusters}. The algorithm is outlined in Alg.~\ref{Alg_Optimization}. In each {partially unrolled period}, we perform forward and backward passes to update $\bm{y}$ and $\bm{\theta}$, respectively. Specifically, in the forward pass, we merge two clusters at each timestep. {In the backward pass}, we run about 20 epochs to update $\bm{\theta}$, {and the affinity matrix $W$ is also updated based on the new representation}. The duration of the $p$-th period is $n_p = ceil(\eta \times n^s_c)$ timesteps, where $n^s_c$ is the number of clusters at the beginning of current period, and $\eta$ is a parameter called \emph{unrolling rate} to control the number of timesteps. The less $\eta$ is, the more frequently we update $\bm{\theta}$.}

\section{Experiments}
\vspace{-5pt}
\subsection{Image Clustering}
We compare our approach with 12 clustering algorithms, including K-means \cite{macqueen1967some}, NJW spectral clustering (SC-NJW) \cite{ng2002spectral}, self-tuning spectral clustering (SC-ST)\cite{zelnik2004self}, large-scale spectral clustering (SC-LS) \cite{chen2011large}, agglomerative clustering with average linkage (AC-Link)\cite{jain1999data}, Zeta function based agglomerative clustering (AC-Zell) \cite{zhao2009cyclizing}, graph degree linkage-based agglomerative clustering (AC-GDL) \cite{zhang2012graph}, agglomerative clustering via path integral (AC-PIC) \cite{zhang2013agglomerative}, normalized cuts (N-Cuts) \cite{shi2000normalized}, locality preserving non-negative matrix factorization (NMF-LP) \cite{cai2009locality}, NMF with deep model (NMF-D) \cite{trigeorgis2014deep}, task-specific clustering with deep model (TSC-D) \cite{wang2015learning}.

For evaluation, we use a commonly used metric: normalized mutual information (NMI) \cite{xu2003document}. It ranges in $[0, 1]$. Larger value indicates more precise clustering results. 

\subsubsection{Datasets}
\begin{table*}[!ht]
\caption{{Datasets used in our experiments.}}
\vspace{-15pt}
\center
\small
\begin{tabular}{lccccccccc}
\toprule
  Dataset & \textit{MNIST} & \textit{USPS} & \textit{COIL20} & \textit{COIL100} & \textit{UMist}  & \textit{FRGC-v2.0} & \textit{CMU-PIE}  & \textit{YTF} \\  
  \midrule
  \#Samples    & 70000 &11000 & 1440   & 7200    & 575  & 2462 & 2856    & 10000           \\
  \#Categories  & 10    & 10   & 20     & 100    & 20    & 20  & 68      & 41                   \\  
  Image Size  & 28$\times$28 & 16$\times$16   &128$\times$128     & 128$\times$128     & 112$\times$92    & 32$\times$32  & 32$\times$32      & 55$\times$55                  \\  
\bottomrule
\end{tabular}
\label{TB_Datasets_Info}
\end{table*}
We evaluate the clustering performance on two hand-written digit image datasets (MNIST \cite{lecun1998gradient} and USPS\footnote{\url{http://www.cs.nyu.edu/~roweis/data.html}}), two multi-view object image datasets (COIL20 and COIL100 \cite{nene1996columbia}), and four face image datasets (UMist \cite{graham1998characterising}, FRGC-v2.0\footnote{\url{http://www3.nd.edu/~cvrl/CVRL/Data_Sets.html}}, CMU-PIE \cite{sim2002cmu}, Youtube-Face (YTF)) \cite{wolf2011face}. The number of samples and categories, and image size are listed in Table~\ref{TB_Datasets_Info}. MNIST consists of training set (60,000) and testing set (10,000). To compare with different approaches, we experiment on the full set (MNIST-full) and testing set (MNIST-test), separately. For face image datasets such as UMist, CMU-PIE, we use the images provided {as is} without any changes. For FRGC-v2.0 and YTF datasets, we first crop faces and then resize them to a constant size. In FRGC-v2.0 dataset, we randomly choose 20 subjects. As for YTF dataset, we choose the first 41 subjects which are sorted by their names in alphabet order.

\subsubsection{Experimental Setup}
\begin{table}
\caption{{{Hyper-parameters in our approach.}}}
\vspace{-15pt}
\small
\center
\begin{tabular}{@{} l  c  c  c c  c  c  c @{}}
\toprule 
  Hyper-parameter & $K_s$ & $a$ & $K_c$ & $\lambda$ & $\gamma$ & $\eta$   \\
  \midrule
  Value  & 20    & 1.0   & 5  & 1.0 & 2.0  & 0.9 {or} 0.2    \\
  \bottomrule
\end{tabular}
\label{TB_AC_Param}
\end{table}

{All the {hyper-parameters} and their values {for our approach} are listed in Table \ref{TB_AC_Param}. {In our experiments, $K_s$ is set to 20, the same value to \cite{zhang2012graph}. $a$ and $\lambda$ are simply set to 1.0. We search the values of $K_c$ and $\gamma$ for best performance on MNIST-test set.} The unrolling rate $\eta$ for first four datasets is 0.9; and 0.2 for face datasets. The target cluster number $n_c^*$ is set to be the number of categories in each dataset.}

We use Caffe \cite{jia2014caffe} to implement our approach. We stacked multiple combinations of convolutional layer, batch normalization layer, ReLU layer and pooling layer. For all the convolutional layers, the number of channels is 50, and filter size is $5 \times 5$ with stride = 1 and padding = 0. For pooling layer, its kernel size is 2 and stride is 2. To deal with varying image sizes across datasets, the number of stacked convolutional layers for each dataset is chosen so that the size of the output feature map is about $10 \times 10$. On the top of all CNNs, we append an inner product (\textit{ip}) layer whose dimension is 160. \textit{ip} layer is followed by a L2-normalization layer before {being} fed to the weighted triplet loss layer or used for clustering. For each partially unrolled period, the base learning rate is set to 0.01, momentum 0.9, and weight decay $5 \times 10^{-5}$. {We use the \textit{inverse} learning rate decay policy, with \textit{Gamma}=0.0001 and \textit{Power}=0.75. {Stochastic gradient descent (SGD)} is adopted for optimization.}

\subsubsection{Quantitative Comparison}
\begin{table*}[!ht]
\caption{{Quantitative clustering performance (NMI) for different algorithms using image intensities as input.}}
\vspace{-15pt}
\center
\small
\begin{tabular}{lccccccccc}
\toprule 
   Dataset & \textit{COIL20} & \textit{COIL100} & \textit{USPS} & \textit{MNIST-test} & \textit{MNIST-full} & \textit{UMist} & \textit{FRGC} & \textit{CMU-PIE} & \textit{YTF} \\   
     
     \midrule
  K-means \cite{macqueen1967some}  &0.775 &0.822 &0.447 &0.528 & 0.500
  &0.609 &0.389 &0.549 & 0.761 \\
  
  SC-NJW \cite{ng2002spectral}     &0.860/0.889 &0.872/0.854  &0.409/0.690  &0.528/0.755 & 0.476 & 0.727    & 0.186     & 0.543  &0.752      \\
      
  SC-ST \cite{zelnik2004self}      & 0.673/0.895 &0.706/0.858   & 0.342/0.726    & 0.445/0.756 & 0.416 & 0.611    & 0.431   &   0.581        &0.620  \\
  
  SC-LS \cite{chen2011large}    & 0.877       &0.833   & 0.681    & 0.756  & 0.706 & 0.810       &  0.550   & 0.788       &0.759 \\
  
  N-Cuts \cite{shi2000normalized}  & 0.768/0.884 &0.861/0.823  & 0.382/0.675   & 0.386/0.753 & 0.411 & 0.782 & 0.285   &  0.411     &0.742 \\
  
  AC-Link \cite{jain1999data}      & 0.512       &0.711  & 0.579    & 0.662  & 0.686 &0.643      &   0.168  & 0.545  &0.738 \\
    
  AC-Zell \cite{zhao2009cyclizing} & 0.954/0.911 &0.963/0.913  & 0.774/0.799    & 0.810/0.768  & 0.017 &0.755    & 0.351    &   0.910  &0.733  \\
  
  AC-GDL \cite{zhang2012graph}     & 0.945/0.937 &0.954/0.929   & 0.854/0.824   & 0.864/0.844  & 0.017  &0.755    & 0.351   &  0.934  &0.622  \\   
  
  AC-PIC \cite{zhang2013agglomerative}   & 0.950 &0.964   & 0.840    & 0.853 & 0.017 &0.750      &  0.415  & 0.902 & 0.697 \\
  
  NMF-LP \cite{cai2009locality}    & 0.720       &0.783   & 0.435    & 0.467  & 0.452 & 0.560    & 0.346      &0.491 & 0.720 \\
  
  NMF-D \cite{trigeorgis2014deep} &0.692 & 0.719   &0.286    &0.243  &0.148 &0.500 & 0.258 & 0.983/0.910  & 0.569\\
    
  TSC-D \cite{wang2015learning}   & -/0.928 &- & -   & -  & -/0.651  & -  & - & - &- \\  
 {OURS-SF} &\textbf{1.000} & {0.978} & {0.858}   & {0.876} & 0.906 & \textbf{0.880} & {0.566}        & {0.984}    &  \textbf{0.848}   \\
 {OURS-RC} &\textbf{1.000} & \textbf{0.985} & \textbf{0.913}   & \textbf{0.915} &\textbf{0.913}  & {0.877} & \textbf{0.574}        & \textbf{1.00}    &  \textbf{0.848}   \\
\bottomrule 
\end{tabular}
\label{TB_Quantitative_Results}
\end{table*}

\begin{table*}[!ht]
\caption{{Quantitative clustering performance (NMI) for different algorithms using our learned representations as inputs.}}
\vspace{-15pt}
\center
\small
\begin{tabular}{lccccccccccc}
  \toprule
   Dataset & \textit{COIL20} & \textit{COIL100} & \textit{USPS} & \textit{MNIST-test} & \textit{MNIST-full} & \textit{UMist} & \textit{FRGC} & \textit{CMU-PIE} & \textit{YTF} \\   
  \midrule
  K-means \cite{macqueen1967some}  &0.926 &0.919 &0.758 &0.908 &0.927 &0.871 &0.636 &0.956       &0.835 \\
  
  SC-NJW \cite{ng2002spectral}     &0.915 &0.898 &0.753 &0.878 &0.931 &0.833 &0.625 &0.957        &0.789 \\
      
  SC-ST \cite{zelnik2004self}      &0.959 &0.922 &0.741 &0.911 &0.906 &0.847 &\textbf{0.651} &0.938 &0.741 \\   
  
  SC-LS \cite{chen2011large}    &0.950 &0.905 &0.780 &0.912 &\textbf{0.932} &\textbf{0.879}       &0.639   & 0.950 &0.802 \\
  
  N-Cuts \cite{shi2000normalized}  &0.963 &0.900 &0.705 &0.910 &0.930 &0.877 &0.640 &0.995     &0.823   \\
  
  AC-Link \cite{jain1999data}      &0.896 &0.884 &0.783 &0.901 &0.918 &0.872 &0.621 &0.990  &{0.803} \\
    
  AC-Zell \cite{zhao2009cyclizing} &\textbf{1.000} &0.989 &0.910 &0.893 &0.919 &0.870 &0.551 &\textbf{1.000}  &0.821  \\
  
  AC-GDL \cite{zhang2012graph}     &\textbf{1.000} &0.985 &0.913 &\textbf{0.915} &0.913  & 0.870 &0.574 &\textbf{1.000} &\textbf{0.84}2  \\   
  
  AC-PIC \cite{zhang2013agglomerative}   &\textbf{1.000} &\textbf{0.990} &\textbf{0.914} &0.909 &0.907  &0.870 &0.553  &\textbf{1.000}  &0.829 \\
  
  NMF-LP \cite{cai2009locality}    &0.855  &0.834 &0.729 &0.905 &0.926 & 0.854    & 0.575      &0.690 &0.788 \\
  \bottomrule
\end{tabular}
\label{TB_Quantitative_Results_withRep}
\end{table*}
%
%
%
%
%
%
%
%
%

We report NMI for different methods on various datasets. {Results are averaged from 3 runs}. We report the results by re-running the code released by original papers. For those that did not release the code, the corresponding results are borrowed from {the papers}. We find the results we obtain are somewhat different from the one reported in original papers. We suspect that these differences may be caused by the different experimental settings or the released code is changed from the one used in the original paper. For all test algorithms, we conduct L2-normalization on the image intensities since it empirically improves the clustering performance. We report our own results in two cases: 1) the straight-forward clustering results obtained when the recurrent process finish, denoted by OURS-SF; 2) the clustering results obtained by re-running clustering algorithm after obtaining the final representation, denoted by OURS-RC. The quantitative results are shown in Table~\ref{TB_Quantitative_Results}. In the table cells, the value before '/' is obtained by re-running code while the value after '/' is that reported in previous papers. 

As we can see from Table~\ref{TB_Quantitative_Results}, both OURS-SF and OURS-RC outperform previous methods on all datasets with noticeable margin. {Interestingly}, we achieved perfect results (NMI = 1) on COIL20 and CMU-PIE datasets, which means that all samples in the same category are clustered into the same group. {The agglomerative clustering algorithms, such as AC-Zell, AC-GDL and AC-PIC perform better than other algorithms generally. However, on MNIST-full test, they all perform {poorly}. The possible reason is that MNIST-full has 70k samples, and these methods cannot cope with such large-scale dataset when using image intensity as representation. However, this problem is addressed by our learned representation. {We show} that we achieved {analogous} performance on MNIST-full to MNIST-test set}. In most cases, we can find OURS-RC performs better on datasets that have room for improvement. {We believe the reason is that} OURS-RC uses the final learned representation over the entire clustering process, while OURS-SF starts with {image intensity}, which indicates that the learned representation is more discriminative than {image intensity}.  
\footnote{We experimented with hand-crafted features such as HOG, LBP, spatial pyramid on a subset of the datasets with some of the better clustering algorithms from Table \ref{TB_Quantitative_Results}, and found that they performed worse.}

\subsubsection{Generalization Across Clustering Algorithms}

{We now evaluate if the representations learned by our joint agglomerative clustering and representation learning approach generalize to other clustering techniques.} We re-run all the clustering algorithms without any changes of parameters, but using our learned deep representations as features. The results are shown in Table~\ref{TB_Quantitative_Results_withRep}. It can be seen that {all clustering algorithms obtain more precise image clusters by using our learned representation}. {Some algorithms like K-means, AC-Link that performed very poorly with raw intensities perform much better with our learned representations}, and the variance in performance across all clustering algorithms is much lower. These results clearly demonstrate that our learned representation is not over-fitting to a single clustering algorithm, but generalizes well across various algorithms. Interestingly, using our learned representation, some of the clustering algorithms perform even better than AC-GDL we build on in our approach.

\subsection{Transferring Learned Representation}
\subsubsection{Cross-Dataset Clustering}
\begin{table}[!ht]
\caption{{NMI performance across COIL20 and COIL100.}}
\vspace{-15pt}
\center
\small
\begin{tabular}{lcccc}
  \toprule
  Layer  & \textit{data} & {top(\textit{ip})} & {top-1} & {top-2}  \\   
  \midrule
  COIL20 $\rightarrow$ COIL100  & 0.924 & 0.927 & \textbf{0.939} & 0.934 \\
  COIL100 $\rightarrow$ COIL20  & 0.944 & 0.949 & \textbf{0.957} & 0.951 \\  
  \bottomrule
\end{tabular}
\label{TB_Cross_Data_Group1}
\end{table}

\begin{table}[!ht]
\caption{{NMI performance across MNIST-test and USPS.}}
\vspace{-15pt}
\center
\small
\begin{tabular}{lcccc}
  \toprule
  Layer  & \textit{data} & {top(\textit{ip})} & {top-1} & {top-2} \\   
  \midrule  
  MNIST-test $\rightarrow$ USPS   & 0.874 & 0.892 & 0.907 & \textbf{0.908} \\
  USPS $\rightarrow$ MNIST-test   & 0.872 & 0.873 & \textbf{0.886} & - \\  
  \bottomrule
\end{tabular}
\label{TB_Cross_Data_Group2}
\end{table}
{In this section, we study whether our learned representations generalize across datasets.} We train a CNN based on our approach on one dataset, and then cluster images from another (but related) dataset using the image features extracted via the CNN. Specifically, we experiment on two dataset pairs: 1) multi-view object datasets (COIL20 and COIL100); 2) hand-written digit datasets (USPS and MNIST-test). We use the representation learned from one dataset to represent another dataset, followed by agglomerative clustering. Note that because the image sizes or channels are different across datasets, we resize the input images {and/or expand the channels} before feeding them to CNN. The experimental results are shown in Table~\ref{TB_Cross_Data_Group1} and \ref{TB_Cross_Data_Group2}. {We use the representations from top \textit{ip} layer and also the \textit{convolutional} or \textit{pooling} layers (top-1, top-2) close to top layer for image clustering}. In two tables, compared with directly using raw image from the \textit{data} layer, the clustering performance based on learned representations from all layers improve, which indicates that the learned representations can be transferred across these datasets. {As perhaps expected}, the performance on target datasets {is worse compared to learning on the target dataset directly}. For COIL20 and COIL100, a possible reason is that they have different image categories. As for MNIST and USPS, the performance beats OURS-SF, but worse than OURS-RC. We find transferring representation learned on MNIST-test to USPS gets close performance to OURS-RC {learned on USPS}.

\subsubsection{Face Verification}

We now evaluate the performance of our approach by applying {it to} face verification. {In particular, the representation {is} learned on Youtube-Face dataset and evaluated on LFW dataset \cite{huang2007labeled} under the restricted protocol. For training, we randomly choose about {10k, 20k, 30k, 50k, 100k} samples from YTF dataset. All these subsets have 1446 categories. We implement our approach to train CNN model and cluster images on the training set. Then, we remove the L2-normalization layer and append a softmax layer to fine-tune our unsupervised CNN model \emph{based on the predicted image cluster labels}.} Using the same training samples and CNN architecture, we also train a CNN model with a softmax loss {supervised by the groundtruth labels of the training set}. {According to the evaluation protocol in \cite{huang2007labeled}}, we run 10-fold cross-validation. The cosine similarity is used {to compute the similarity between samples. In each of 10 cross-validations, nine folds are used to find the optimal threshold, and the remaining one fold is used for evaluation. The average accuracy is reported in Table.~\ref{TB_LFW_Perf}}. As shown, though no {groundtruth labels are used for representation learning in our approach}, we obtain analogous performance to the supervised learning approach. {Our approach even (slightly) beats the supervised learning method in one case.}

\begin{table}
\caption{{Face verification results on LFW.}}
\vspace{-15pt}
\center
\small
\begin{tabular}{lcccccc}
  \toprule
  \#Samples      & 10k    & 20k    & 30k     & 50k   & 100k     \\   
  \midrule 
   Supervised    & \textbf{0.73}7  & \textbf{0.746}  &  0.748  & \textbf{0.764} & 0.770 \\  
   OURS          & 0.728  & 0.743  & \textbf{0.750}  & 0.762 & 0.767    \\
   \bottomrule
\end{tabular}
\label{TB_LFW_Perf}
\end{table}

\subsection{Image Classification}
Recently, unsupervised representation learning methods are starting to achieve promising results for a variety of recognition tasks \cite{coates2011analysis, coates2011selecting, jia2012beyond, lin2014stable}. We are interested in knowing whether the proposed method can also learn useful representation for image classification. We experiment with CIFAR-10 \cite{krizhevsky2009learning}. We follow the pipeline in \cite{coates2011analysis}, and base our experiments on their publicly available code. In this pipeline, codebook with 1600 codes is build upon $6 \times 6$ ZCA-whitened image patches, and then used to code the training and testing samples by extracting 1,600-d feature from each of 4 image quadrants. Afterwards, a linear SVM \cite{cortes1995support} is applied for image classification on 6,400-d feature. In our approach, the only difference is that we learn a new representation from $6 \times 6$ patches, and then use these new representations to build the codebook with 1,600 codes. The CNN architecture we use contains two convolutional layers, each of which is combined with a ReLu and a pooling layer, followed by an inner product layer. Both convolutional layers have 50 $3 \times 3$ filters with pad = 1. The kernel size of pooling layer is 2, and the stride is 2. To save on training time, 40k randomly extracted patches are extracted from 50k training set and used in all the experiments.

Classification accuracies on test set with different settings are shown in Table~\ref{TB_CIFAR10_Perf}. We vary the number of training samples and evaluate the performance for representations from different layers. As we can see, the combination of representations from the first and second convolutional layer achieve the best performance. {We also use the representation output by inner product layer to learn the codebook. However, it performs poorly. {A} possible reason is that it discards spatial information of image patches, which may be important for learning a codebook}. When using 400k randomly extracted patches to learn the codebook, \cite{coates2011analysis} achieved 77.9\%. However, it is still lower than what we achieved. This performance also beats several other methods listed in \cite{coates2011selecting, goodfellow2012spike, jia2012beyond, lin2014stable}.

\begin{table}
\caption{{Image classification accuracy on CIFAR-10.}}
\vspace{-15pt}
\center
\small
\begin{tabular}{lccccc}
  \toprule
  \#Samples & K-means \cite{coates2011analysis} & conv1 & conv2 & conv1\&2 \\   
  \midrule
  
  5k        &62.81\% &63.05\% &63.10\% &\textbf{63.50}\%\\
  
  10k       &68.01\% &68.30\% &68.46\% &\textbf{69.11}\% \\  
  
  25k       &74.01\% &72.83\% &72.93\% &\textbf{75.11}\%\\
  
  50k (full set)&76.59\% &74.68\% &74.68\% &\textbf{78.55}\% \\
  \bottomrule
\end{tabular}
\label{TB_CIFAR10_Perf}
\end{table}

%
%
%
%

\section{Conclusion}
{In this paper, we have proposed an approach to {jointly learn deep representations and image clusters. In our approach, we combined agglomerative clustering with CNNs and formulate them as a recurrent process. We used a partially unrolling strategy to divide the timesteps into multiple periods. In each period, we merged clusters step by step during the forward pass and learned representation in the backward pass, which are guided by a single weighted triplet-loss function.} The extensive experiments on image clustering, deep representation transfer learning and image classification demonstrate that our approach can obtain more precise image clusters and discriminative representations that generalize well across many datasets and tasks.}

\section{Acknowledgements}
This work was supported in part by the Paul G. Allen Family Foundation, Google, and Institute for Critical Technology and Applied Science (ICTAS) at Virginia Tech through awards to D. P.; and by a National Science Foundation CAREER award, an Army Research Office YIP award, an Office of Naval Research grant, an AWS in Education Research Grant, and GPU support by NVIDIA to D. B. The views and conclusions contained herein are those of the authors and should not be interpreted as necessarily representing the official policies or endorsements, either expressed or implied, of the U.S. Government or any sponsor.

\bibliographystyle{ieee}
{\small
\bibliography{egbib}
}
\appendix
\section{Appendix}

\subsection{Affinity Measure for Clusters}
\label{Ap_Affinity_Measure_Cluster}
In this paper, we employ the affinity measure in \cite{zhang2012graph}
\begin{equation}
\begin{aligned}
\bm{\mathcal{A}}(\mathcal{C}_i, \mathcal{C}_j) & = \bm{\mathcal{A}}(\mathcal{C}_j \rightarrow \mathcal{C}_i) + \bm{\mathcal{A}}(\mathcal{C}_i \rightarrow \mathcal{C}_j) \\
& = \frac{1}{|\mathcal{C}_i|^2} \bm{1}^{\bm{T}}_{|\mathcal{C}_i|} \bm{W}_{\mathcal{C}_i, \mathcal{C}_j} \bm{W}_{\mathcal{C}_j, \mathcal{C}_i}\bm{1}_{|\mathcal{C}_i|} \\
& + \frac{1}{|\mathcal{C}_j|^2} \bm{1}^{\bm{T}}_{|\mathcal{C}_j|} \bm{W}_{\mathcal{C}_j, \mathcal{C}_i} \bm{W}_{\mathcal{C}_i, \mathcal{C}_j}\bm{1}_{|\mathcal{C}_j|}
\end{aligned}
\label{EQ_GDL_Affinity}
\end{equation}
where $\bm{W}$ is the affinity matrix for samples, and $\bm{W}_{\mathcal{C}_i, \mathcal{C}_j} \in \mathbb{R}^{|\mathcal{C}_i| \times |\mathcal{C}_j|}$ is the submatrix in $\bm{W}$ pointing from samples in $\mathcal{C}_i$ to samples in $\mathcal{C}_j$, and $\bm{W}_{\mathcal{C}_j, \mathcal{C}_i} \in \mathbb{R}^{|\mathcal{C}_j| \times |\mathcal{C}_i|} $ is the one pointing from $\mathcal{C}_j$ to $\mathcal{C}_i$. $\bm{1}_{|\mathcal{C}_i|}$ and $\bm{1}_{|\mathcal{C}_j|}$ are two vectors with all $|\mathcal{C}_i|$ and $|\mathcal{C}_j|$ elements be 1, respectively. Therefore, we have $\bm{\mathcal{A}}(\mathcal{C}_i, \mathcal{C}_j) = \bm{\mathcal{A}}(\mathcal{C}_j, \mathcal{C}_i)$.

According to \eqref{EQ_GDL_Affinity}, we can derive 
\begin{equation}
\bm{\mathcal{A}}((\mathcal{C}_m \cup \mathcal{C}_n) \rightarrow \mathcal{C}_i) = \bm{\mathcal{A}}(\mathcal{C}_m \rightarrow \mathcal{C}_i) + \bm{\mathcal{A}}(\mathcal{C}_n \rightarrow \mathcal{C}_i)
\label{Eq_Affinity_fromMergedCs}
\end{equation}
which has also been shown in \cite{zhang2012graph}. Meanwhile,
\begin{equation}
\begin{aligned}
& \bm{\mathcal{A}}(\mathcal{C}_i \rightarrow (\mathcal{C}_m \cup \mathcal{C}_n)) \\
& = \beta \bm{1}^{\bm{T}}_{|\mathcal{C}_m| + |\mathcal{C}_n|} \bm{W}_{\mathcal{C}_m \cup \mathcal{C}_n, \mathcal{C}_i} \bm{W}_{\mathcal{C}_i, \mathcal{C}_m \cup \mathcal{C}_n}\bm{1}_{|\mathcal{C}_m| + |\mathcal{C}_n|} \hfill \\
& = \beta  \bm{1}^{\bm{T}}_{|\mathcal{C}_m|} \bm{W}_{\mathcal{C}_m, \mathcal{C}_i} \bm{W}_{\mathcal{C}_i, \mathcal{C}_m}\bm{1}_{|\mathcal{C}_m|}
+ \beta  \bm{1}^{\bm{T}}_{|\mathcal{C}_n|} \bm{W}_{\mathcal{C}_n, \mathcal{C}_i} \bm{W}_{\mathcal{C}_i, \mathcal{C}_n}\bm{1}_{|\mathcal{C}_n|} \\
& + \beta  \bm{1}^{\bm{T}}_{|\mathcal{C}_m|} \bm{W}_{\mathcal{C}_m, \mathcal{C}_i} \bm{W}_{\mathcal{C}_i, \mathcal{C}_n}\bm{1}_{|\mathcal{C}_n|}
+ \beta \bm{1}^{\bm{T}}_{|\mathcal{C}_n|} \bm{W}_{\mathcal{C}_n, \mathcal{C}_i} \bm{W}_{\mathcal{C}_i, \mathcal{C}_m}\bm{1}_{|\mathcal{C}_m|} \\
\end{aligned}
\phantom{\hspace{50cm}} 
\label{Eq_Affinity_toMergedCs}
\end{equation}
where $\beta = {1}/{(|\mathcal{C}_m| +  |\mathcal{C}_n|)^2}$.

\subsection{Approximated Affinity Measure}
\label{Ap_Affinity_Measure}
\begin{figure*}[t]
   \begin{minipage}{0.5\linewidth}
   \centering
   \epsfig{file=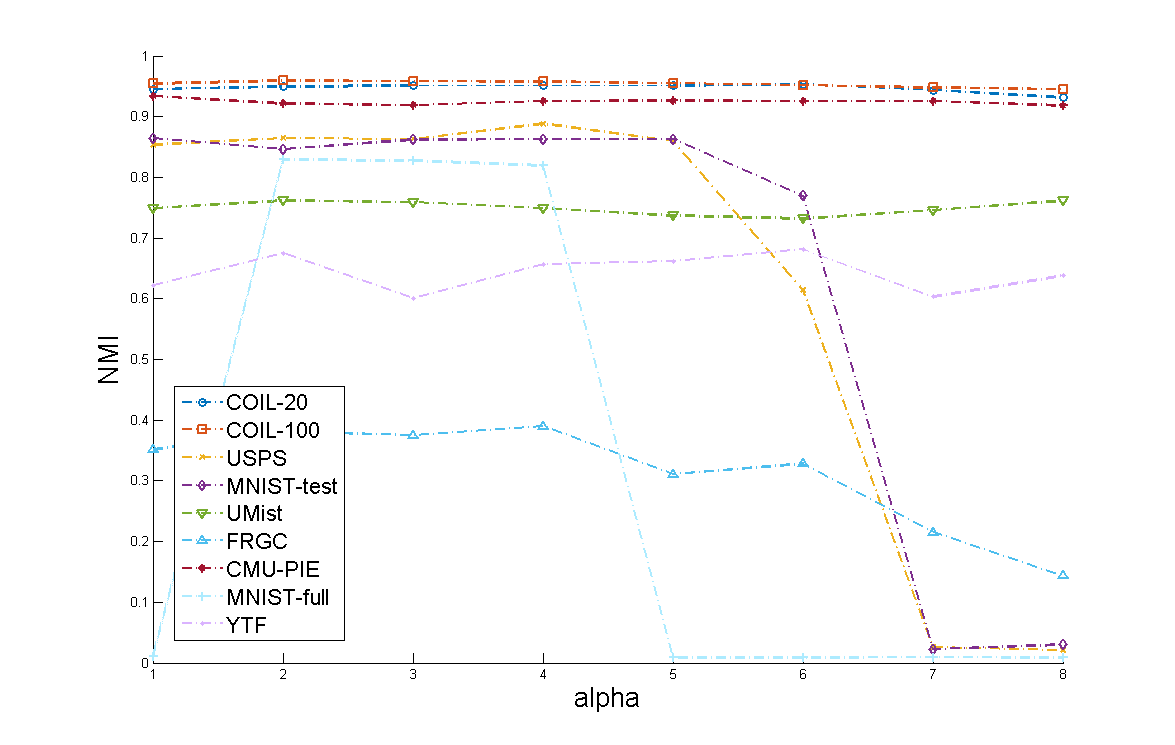, scale=0.3}
   \end{minipage}
   \begin{minipage}{0.5\linewidth}
   \centering
   \epsfig{file=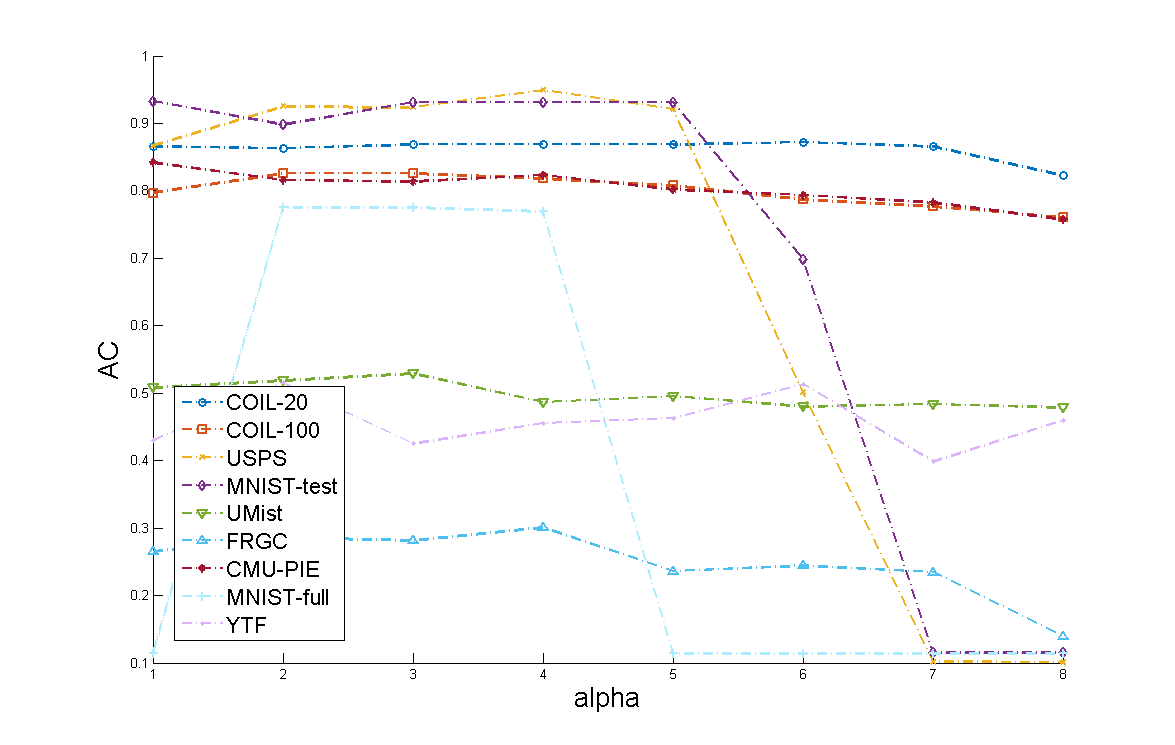, scale=0.3}
   \end{minipage}        
   \caption{Performance of agglomerative clustering with approximations. Left one is NMI metric, and right one is AC metric. The first column is without acceleration. For the other columns from left to right, $\alpha = \{-0.2, -0.1, 0, 0.1, 0.2, 0.3, 0.5\}$.}
   \label{Fig_Approximation_EvalPerf}
\end{figure*}
During agglomerative clustering, we need to re-compute the affinity between the merged cluster to all other clusters based on \ref{Eq_Affinity_fromMergedCs} and \ref{Eq_Affinity_toMergedCs} repeatedly. It is simple to compute \ref{Eq_Affinity_fromMergedCs}. However, to get $\bm{\mathcal{A}}(\mathcal{C}_i \rightarrow (\mathcal{C}_m \cup \mathcal{C}_n))$, we need a lot of computations. These time costs become dominant and remarkable when we have a large-scale dataset. To accelerate the computations, we introduce an approximation method. At the right side of \eqref{Eq_Affinity_toMergedCs}, we assume samples in $\mathcal{C}_m$ and $\mathcal{C}_n$ have similar affinities to $\mathcal{C}_i$. This assumption is mild because the condition to merge $\mathcal{C}_m$ and $\mathcal{C}_n$ is that they are similar to each other. In this case, the ratio between $\bm{W}_{\mathcal{C}_i, \mathcal{C}_m} \bm{1}_{|\mathcal{C}_m|}$ and $\bm{W}_{\mathcal{C}_i, \mathcal{C}_n} \bm{1}_{|\mathcal{C}_n|}$ is analogy to the  ratio between the number of samples in two set, i.e.,
\begin{equation}
\bm{W}_{\mathcal{C}_i, \mathcal{C}_m} \bm{1}_{|\mathcal{C}_m|} = \frac{|\mathcal{C}_m|}{|\mathcal{C}_n|}\bm{W}_{\mathcal{C}_i, \mathcal{C}_n} \bm{1}_{|\mathcal{C}_n|}
\label{Eq_Affinity_Approximation}
\end{equation}

Based on \eqref{Eq_Affinity_Approximation}, we have
\begin{subequations}
\begin{equation}
\bm{1}^{\bm{T}}_{|\mathcal{C}_m|} \bm{W}_{\mathcal{C}_m, \mathcal{C}_i} \bm{W}_{\mathcal{C}_i, \mathcal{C}_n}\bm{1}_{|\mathcal{C}_n|} = \frac{|\mathcal{C}_n|}{|\mathcal{C}_m|} \bm{1}^{\bm{T}}_{|\mathcal{C}_m|} \bm{W}_{\mathcal{C}_m, \mathcal{C}_i} \bm{W}_{\mathcal{C}_i, \mathcal{C}_m}\bm{1}_{|\mathcal{C}_m|} \\
\end{equation}
\begin{equation} 
\bm{1}^{\bm{T}}_{|\mathcal{C}_n|} \bm{W}_{\mathcal{C}_n, \mathcal{C}_i} \bm{W}_{\mathcal{C}_i, \mathcal{C}_m}\bm{1}_{|\mathcal{C}_m|} = \frac{|\mathcal{C}_m|}{|\mathcal{C}_n|} \bm{1}^{\bm{T}}_{|\mathcal{C}_n|} \bm{W}_{\mathcal{C}_n, \mathcal{C}_i} \bm{W}_{\mathcal{C}_i, \mathcal{C}_n}\bm{1}_{|\mathcal{C}_n|}
\end{equation}
\end{subequations}

As a result, we can re-formulate \eqref{Eq_Affinity_toMergedCs} to
\begin{equation}
\begin{aligned}
& \bm{\mathcal{A}}(\mathcal{C}_i \rightarrow (\mathcal{C}_m \cup \mathcal{C}_n))\\
& = \frac{1}{(|\mathcal{C}_m|^2 +  |\mathcal{C}_m||\mathcal{C}_n|)}  \bm{1}^{\bm{T}}_{|\mathcal{C}_m|} \bm{W}_{\mathcal{C}_m, \mathcal{C}_i} \bm{W}_{\mathcal{C}_i, \mathcal{C}_m}\bm{1}_{|\mathcal{C}_m|} & \\ 
& + \frac{1}{(|\mathcal{C}_m||\mathcal{C}_n| +  |\mathcal{C}_n|^2)}  \bm{1}^{\bm{T}}_{|\mathcal{C}_n|} \bm{W}_{\mathcal{C}_n, \mathcal{C}_i} \bm{W}_{\mathcal{C}_i, \mathcal{C}_n}\bm{1}_{|\mathcal{C}_n|} & \\
\end{aligned}
\phantom{\hspace{50cm}} 
\label{Eq_Affinity_toMergedCs_Simplified}
\end{equation}

Therefore, we have
\begin{equation}
\begin{aligned}
\bm{\mathcal{A}}(\mathcal{C}_i \rightarrow (\mathcal{C}_m \cup \mathcal{C}_n)) & = \frac{|\mathcal{C}_m|}{|\mathcal{C}_m| +  |\mathcal{C}_n|} \bm{\mathcal{A}}(\mathcal{C}_i \rightarrow \mathcal{C}_m) \\
& + \frac{|\mathcal{C}_n|}{|\mathcal{C}_m| +  |\mathcal{C}_n|} \bm{\mathcal{A}}(\mathcal{C}_i \rightarrow \mathcal{C}_n)
\end{aligned}
\label{Eq_Affinity_toMergedCs_Simplified}
\end{equation}

Consequently, we have
\begin{equation}
\begin{aligned}
\bm{\mathcal{A}}(\mathcal{C}_m \cup \mathcal{C}_n, \mathcal{C}_i) & = \bm{\mathcal{A}}(\mathcal{C}_m \rightarrow \mathcal{C}_i) + \bm{\mathcal{A}}(\mathcal{C}_n \rightarrow \mathcal{C}_i) \\
& + \frac{|\mathcal{C}_m|}{|\mathcal{C}_m| +  |\mathcal{C}_n|} \bm{\mathcal{A}}(\mathcal{C}_i \rightarrow \mathcal{C}_m) \\
& + \frac{|\mathcal{C}_n|}{|\mathcal{C}_m| +  |\mathcal{C}_n|} \bm{\mathcal{A}}(\mathcal{C}_i \rightarrow \mathcal{C}_n)
\end{aligned}
\label{Eq_Affinity_Mutual_Simplified}
\end{equation}

Above approximation provides us a potential way to reduce the computational complexity of agglomerative clustering. Though we computed $\bm{\mathcal{A}}(\mathcal{C}_i \rightarrow (\mathcal{C} \cup \mathcal{C}_n))$ based on Eq.~\eqref{Eq_Affinity_toMergedCs} in all our experiments, we found the approximation version achieves analogy performance while costs much less time than the original one. We further simplify the computation by assuming a constant ratio $\alpha$ between the terms at the right side of Eq. \eqref{Eq_Affinity_toMergedCs}:
\begin{subequations}
\begin{equation}
\bm{1}^{\bm{T}}_{|\mathcal{C}_m|} \bm{W}_{\mathcal{C}_m, \mathcal{C}_i} \bm{W}_{\mathcal{C}_i, \mathcal{C}_n}\bm{1}_{|\mathcal{C}_n|} = \alpha \bm{1}^{\bm{T}}_{|\mathcal{C}_m|} \bm{W}_{\mathcal{C}_m, \mathcal{C}_i} \bm{W}_{\mathcal{C}_i, \mathcal{C}_m}\bm{1}_{|\mathcal{C}_m|} \\
\end{equation}
\begin{equation} 
\bm{1}^{\bm{T}}_{|\mathcal{C}_n|} \bm{W}_{\mathcal{C}_n, \mathcal{C}_i} \bm{W}_{\mathcal{C}_i, \mathcal{C}_m}\bm{1}_{|\mathcal{C}_m|} = \alpha \bm{1}^{\bm{T}}_{|\mathcal{C}_n|} \bm{W}_{\mathcal{C}_n, \mathcal{C}_i} \bm{W}_{\mathcal{C}_i, \mathcal{C}_n}\bm{1}_{|\mathcal{C}_n|}
\end{equation}
\end{subequations}

Based on above assumption,
\begin{equation}
\begin{aligned}
\bm{\mathcal{A}}(\mathcal{C}_m \cup \mathcal{C}_n, \mathcal{C}_i) & = \bm{\mathcal{A}}(\mathcal{C}_m \rightarrow \mathcal{C}_i) + \bm{\mathcal{A}}(\mathcal{C}_n \rightarrow \mathcal{C}_i) \\
& + \frac{(1 + \alpha)|\mathcal{C}_m|^2}{(|\mathcal{C}_m| +  |\mathcal{C}_n|)^2} \bm{\mathcal{A}}(\mathcal{C}_i \rightarrow \mathcal{C}_m) \\
& + \frac{(1 + \alpha)|\mathcal{C}_n|^2}{(|\mathcal{C}_m| +  |\mathcal{C}_n|)^2} \bm{\mathcal{A}}(\mathcal{C}_i \rightarrow \mathcal{C}_n)
\end{aligned}
\label{Eq_Affinity_Mutual_Simplified_alpha}
\end{equation}

\begin{figure}[t]
   \begin{minipage}{1\linewidth}
   \center
   \epsfig{file=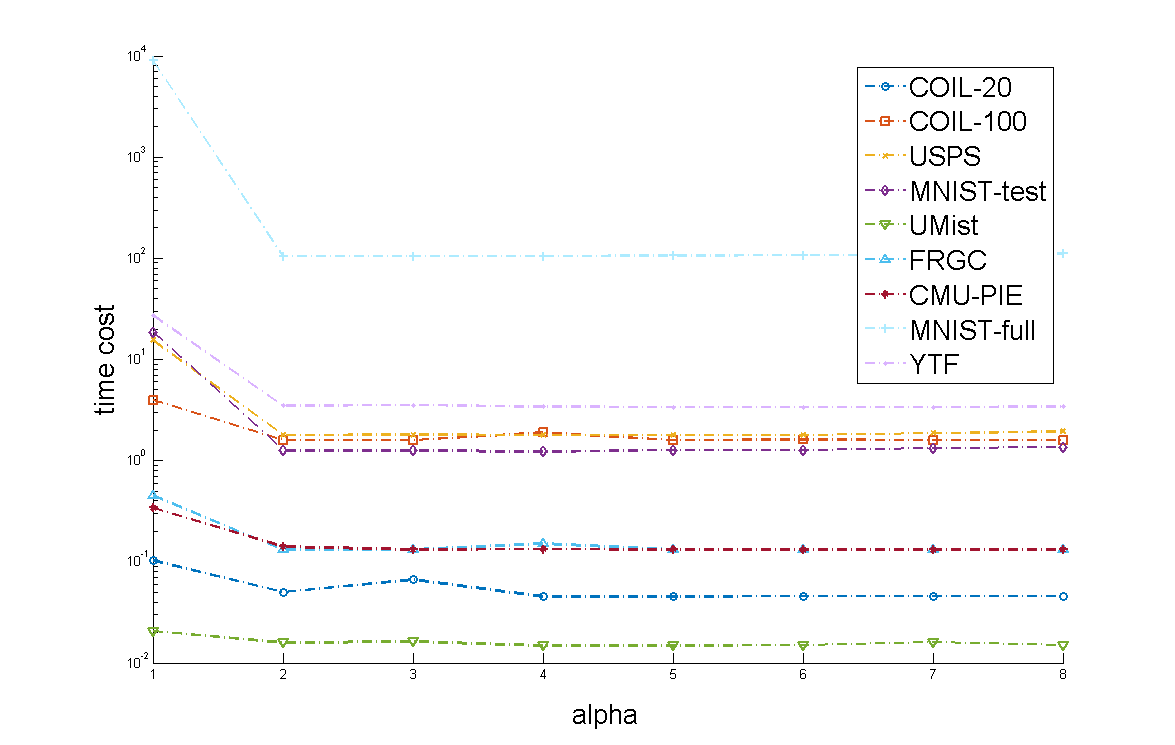, trim=2cm 0 0 0, clip, scale=0.28}
   \end{minipage}       
   \caption{Time cost for different values of $\alpha$. The first column is the time cost without acceleration. For the other columns from left to right, $\alpha = \{-0.2, -0.1, 0, 0.1, 0.2, 0.3, 0.5\}$.}
   \label{Fig_Approximation_EvalTime}
\end{figure}

We test various values for $\alpha$, which are $\{-0.2, -0.1, 0, 0.1, 0.2, 0.3, 0.5\}$. We show the quantitative comparison in Fig.~\ref{Fig_Approximation_EvalPerf}. We use image intensities as input to rule out all random factors. The original AC-GDL algorithm is used as the baseline. By conducting experiments on various datasets, we find a valid range $[-0.2, 0.1]$ for $\alpha$ which helps achieve analogous or even better performance to the one without acceleration. These results indicate that we may do not need to compute the explicit value of affinities to obtain equivalent level performance. Also, to measure how much time we can save by using our approximation, we compare the time cost between original AC-GDL algorithm and accelerated one in Fig.~\ref{Fig_Approximation_EvalTime}. It is clear that our approximation algorithm has much lower computational complexity.

\subsection{Cluster-based to Sample-based Loss}
\begin{figure}[!ht]
   \begin{minipage}{1.0\linewidth}
   \centering
   \epsfig{file=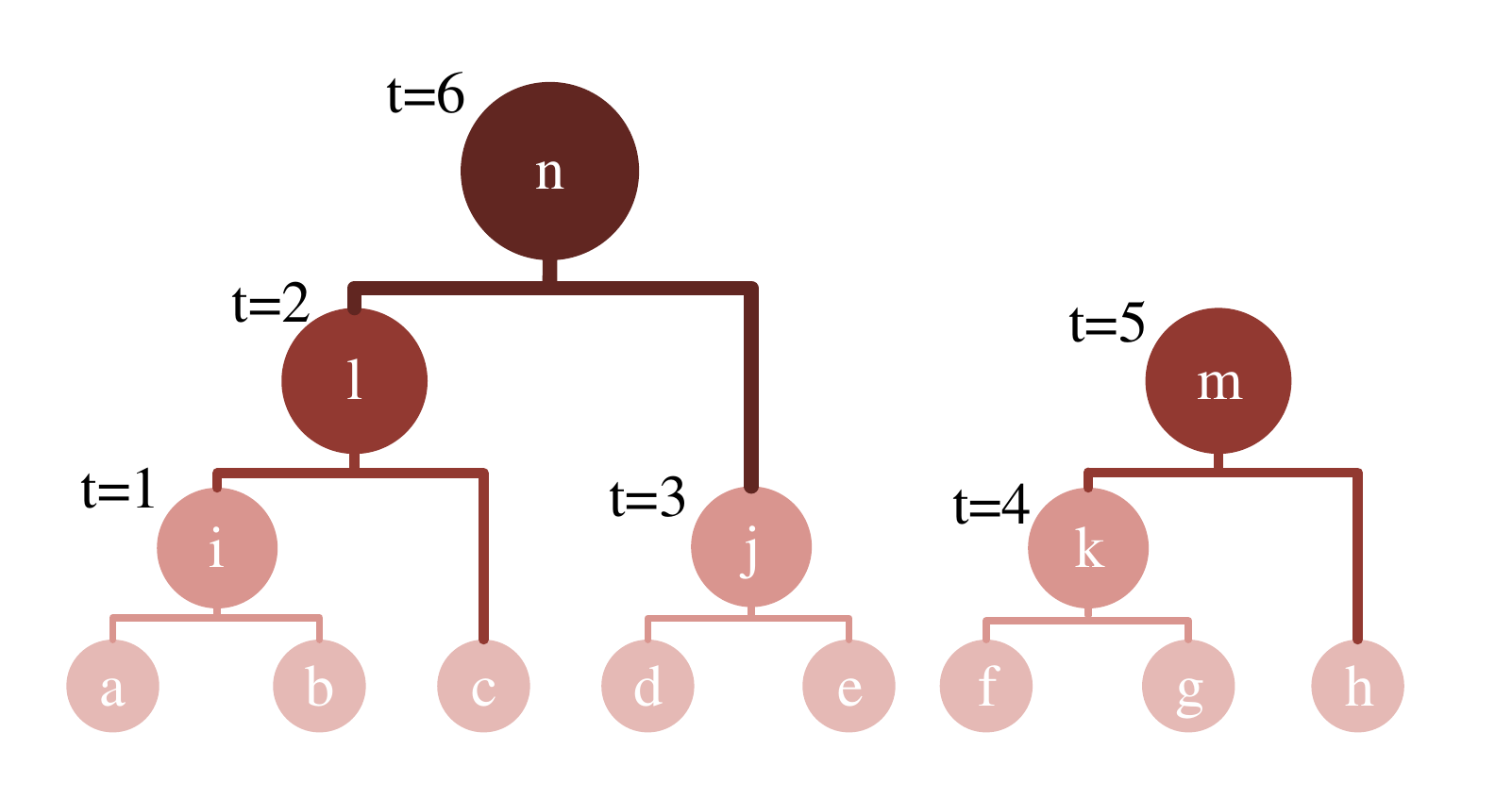, scale=0.51}
   \end{minipage}
   \caption{A illustration of agglomerative clustering.}
   \label{Fig_AgglomerativeCLustering}
\end{figure}

In this part, we explain how to convert cluster-based loss to sample-based loss. Because it depends on specific agglomerative clustering processes, we use a toy example in Fig.~\ref{Fig_AgglomerativeCLustering} for illustration. We set $K_c$ be 2 for simplicity. In Fig.~\ref{Fig_AgglomerativeCLustering}, there are six time steps, and thus $T=6$. We assume they are in a single partial unrolled period. The leaf nodes represent single samples. For simplicity, we omit $\frac{\lambda}{K_c-1}$ in \eqref{Eq_Loss_triplet_time_step}, obtaining the overall loss
\begin{equation}
\small
\begin{aligned}
\mathcal{L}(\bm{\theta} | \bm{\mathcal{Y}}_*, \bm{I}) = -\sum_{t = 1}^6 \left(\lambda' \bm{\mathcal{A}}(\mathcal{C}^t_*, \mathcal{N}_{\mathcal{C}^t_*}^{K_c}[1]) - \bm{\mathcal{A}}(\mathcal{C}^t_*, \mathcal{N}_{\mathcal{C}^t_*}^{K_c}[2])\right)
\end{aligned}
\label{Eq_Loss_triplet_time_step_nogamma}
\end{equation}

Given above loss function, we decompose it from first time step ($t = 1$) to the most recent time step ($t = 6$):
\begin{itemize}
\item \textbf{t=1}: $\mathcal{C}^1_* = \mathcal{C}_a$, $\mathcal{N}_{\mathcal{C}^1_*}^{2}[1] = \mathcal{C}_b$ and $\mathcal{N}_{\mathcal{C}^1_*}^{2}[2] = \mathcal{C}_c$. We have
\begin{equation}
\mathcal{L}(\bm{\theta}|\bm{y}^1_*, I) = -\left(\lambda'\bm{\mathcal{A}}(\mathcal{C}_a, \mathcal{C}_b) - \bm{\mathcal{A}}(\mathcal{C}_a, \mathcal{C}_c)\right)
\end{equation}

Clearly, above is sample-based weighted triplet loss function, where samples $\mathcal{C}_a$ and $\mathcal{C}_b$ are positive pair and $\mathcal{C}_a$ and $\mathcal{C}_c$ are negative pair.

\item \textbf{t=2}: $\mathcal{C}^2_* = \mathcal{C}_i$, $\mathcal{N}_{\mathcal{C}^2_*}^{2}[1] = \mathcal{C}_c$ and $\mathcal{N}_{\mathcal{C}^2_*}^{2}[2] = \mathcal{C}_d$. We have
\begin{align}
\mathcal{L}(\bm{\theta}|\{\bm{y}^1_*, \bm{y}^2_*\}, I) 
& = \mathcal{L}(\bm{\theta}|\bm{y}^1_*, I) \nonumber \\ 
& - \left(\lambda'\bm{\mathcal{A}}(\mathcal{C}_i, \mathcal{C}_c) - \bm{\mathcal{A}}(\mathcal{C}_i, \mathcal{C}_d)\right)
\end{align}

Since $\mathcal{C}_i = \mathcal{C}_a \cup \mathcal{C}_b$, we base on Eq.~\eqref{Eq_Affinity_Mutual_Simplified} for approximation
\begin{equation}
\begin{aligned}
\bm{\mathcal{A}}(\mathcal{C}_i, \mathcal{C}_c) &= \bm{\mathcal{A}}(\mathcal{C}_a \rightarrow \mathcal{C}_c) + \bm{\mathcal{A}}(\mathcal{C}_b \rightarrow \mathcal{C}_c) \\
& + \frac{1}{2} \bm{\mathcal{A}}(\mathcal{C}_c \rightarrow \mathcal{C}_a) + \frac{1}{2} \bm{\mathcal{A}}(\mathcal{C}_c \rightarrow \mathcal{C}_b)
\end{aligned}
\end{equation}
\begin{equation}
\begin{aligned}
\bm{\mathcal{A}}(\mathcal{C}_i, \mathcal{C}_d) &= \bm{\mathcal{A}}(\mathcal{C}_a \rightarrow \mathcal{C}_d) + \bm{\mathcal{A}}(\mathcal{C}_b \rightarrow \mathcal{C}_d) \\
& + \frac{1}{2} \bm{\mathcal{A}}(\mathcal{C}_d \rightarrow \mathcal{C}_a) + \frac{1}{2} \bm{\mathcal{A}}(\mathcal{C}_d \rightarrow \mathcal{C}_b)
\end{aligned}
\end{equation}

Thus, 
\begin{equation}
\begin{aligned}
& \mathcal{L}(\bm{\theta}|\{\bm{y}^1_*, \bm{y}^2_*\}, I) \\
& = -\lambda'\bm{\mathcal{A}}(\mathcal{C}_a, \mathcal{C}_b) - (\lambda' - 1) \bm{\mathcal{A}}(\mathcal{C}_a \rightarrow \mathcal{C}_c) - \lambda' \bm{\mathcal{A}}(\mathcal{C}_b \rightarrow \mathcal{C}_c) \\
& - (\frac{\lambda'}{2} - 1) \bm{\mathcal{A}}(\mathcal{C}_c \rightarrow \mathcal{C}_a) - \frac{\lambda'}{2}\bm{\mathcal{A}}(\mathcal{C}_c \rightarrow \mathcal{C}_b) \\
& + \bm{\mathcal{A}}(\mathcal{C}_a \rightarrow \mathcal{C}_d) + \bm{\mathcal{A}}(\mathcal{C}_b \rightarrow \mathcal{C}_d) \\
& + \frac{1}{2}\bm{\mathcal{A}}(\mathcal{C}_d \rightarrow \mathcal{C}_a) + \frac{1}{2}
\bm{\mathcal{A}}(\mathcal{C}_d \rightarrow \mathcal{C}_b)
\end{aligned}
\label{Eq_Loss_t_2}
\end{equation}

At current time step, sample \textit{a}, \textit{b} and \textit{c} belong to the same cluster $\mathcal{C}_l$, while sample \textit{d} is from another cluster. \eqref{Eq_Loss_t_2} computes the sample-based weighted triplet loss for samples in $\mathcal{C}_l$ and sample \textit{d}. Except for $\mathcal{C}_l$, the other clusters all have merely one sample. No need to compute triplet loss for them. It should be pointed out that $\lambda'$ in above loss function should be not less than 2 so that the affinities for all pairs in $\mathcal{C}_l$ are enlarged.

\item \textbf{t=3}: $\mathcal{C}^3_* = \mathcal{C}_d$, $\mathcal{N}_{\mathcal{C}^3_*}^{2}[1] = \mathcal{C}_e$ and $\mathcal{N}_{\mathcal{C}^3_*}^{2}[2] = \mathcal{C}_f$. We have
\begin{equation}
\begin{aligned}
\mathcal{L}(\bm{\theta}|\{\bm{y}^1_*, \bm{y}^2_*, \bm{y}^3_*\}, I) & = \mathcal{L}(\bm{\theta}|\{\bm{y}^1_*, \bm{y}^2_*\}, I) \\
& - \lambda'\left(\bm{\mathcal{A}}(\mathcal{C}_d, \mathcal{C}_e) - \bm{\mathcal{A}}(\mathcal{C}_d, \mathcal{C}_f)\right)
\end{aligned}
\label{Eq_Loss_t_3}
\end{equation}

Besides the loss $\mathcal{L}(\bm{\theta}|\{\bm{y}^1_*, \bm{y}^2_*\}, I)$ for $\mathcal{C}_l$, we also compute the loss for $\mathcal{C}_j$ in \eqref{Eq_Loss_t_3} because it contains two samples, \textit{d} and \textit{e}.

\item \textbf{t=4}: $\mathcal{C}^4_* = \mathcal{C}_f$, $\mathcal{N}_{\mathcal{C}^4_*}^{2}[1] = \mathcal{C}_g$ and $\mathcal{N}_{\mathcal{C}^4_*}^{2}[2] = \mathcal{C}_h$. We have
\begin{equation}
\begin{aligned}
\mathcal{L}(\bm{\theta}|\{\bm{y}^1_*,...,\bm{y}^4_*\}, I) 
& = \mathcal{L}(\bm{\theta}|\{\bm{y}^1_*, \bm{y}^2_*, \bm{y}^3_*\}, I) \\
& - \left(\lambda'\bm{\mathcal{A}}(\mathcal{C}_f, \mathcal{C}_g) - \bm{\mathcal{A}}(\mathcal{C}_f, \mathcal{C}_h)\right)
\end{aligned}
\end{equation}

Here, we additionally compute the weighted triplet loss for cluster $\mathcal{C}_k$ since it contains two samples.

\item \textbf{t=5}: $\mathcal{C}^5_* = \mathcal{C}_k$, $\mathcal{N}_{\mathcal{C}^5_*}^{2}[1] = \mathcal{C}_h$ and $\mathcal{N}_{\mathcal{C}^5_*}^{2}[2] = \mathcal{C}_j$. We have
\begin{equation}
\begin{aligned}
& \mathcal{L}(\bm{\theta}|\{\bm{y}^1_*,...,\bm{y}^5_*\}, I) \\ 
& = \mathcal{L}(\bm{\theta}|\{\bm{y}^1_*,...,\bm{y}^4_*\}, I) - \left(\lambda'\bm{\mathcal{A}}(\mathcal{C}_k, \mathcal{C}_h) - \bm{\mathcal{A}}(\mathcal{C}_k, \mathcal{C}_j)\right)
\end{aligned}
\end{equation}

Because $\mathcal{C}_k  = \mathcal{C}_f \cup \mathcal{C}_g$, we have
\begin{equation}
\begin{aligned}
\bm{\mathcal{A}}(\mathcal{C}_k, \mathcal{C}_h) & = \bm{\mathcal{A}}(\mathcal{C}_f \rightarrow \mathcal{C}_h) + \bm{\mathcal{A}}(\mathcal{C}_g \rightarrow \mathcal{C}_h) \\
& + \frac{1}{2}\bm{\mathcal{A}}(\mathcal{C}_h \rightarrow \mathcal{C}_f) + \frac{1}{2}\bm{\mathcal{A}}(\mathcal{C}_h \rightarrow \mathcal{C}_g)
\end{aligned}
\end{equation}
\begin{equation}
\begin{aligned}
\bm{\mathcal{A}}(\mathcal{C}_k, \mathcal{C}_j) &= \bm{\mathcal{A}}(\mathcal{C}_f \rightarrow \mathcal{C}_j) + \bm{\mathcal{A}}(\mathcal{C}_g \rightarrow \mathcal{C}_j) \\
& + \frac{1}{2}\bm{\mathcal{A}}(\mathcal{C}_j \rightarrow \mathcal{C}_f) + \frac{1}{2}\bm{\mathcal{A}}(\mathcal{C}_j \rightarrow \mathcal{C}_g)
\end{aligned}
\end{equation}

Since $\mathcal{C}_j  = \mathcal{C}_d \cup \mathcal{C}_e$, we further transform above equation to
\begin{equation}
\begin{aligned}
\bm{\mathcal{A}}(\mathcal{C}_k, \mathcal{C}_j) & = \frac{1}{2}\bm{\mathcal{A}}(\mathcal{C}_f \rightarrow \mathcal{C}_d) + \frac{1}{2}\bm{\mathcal{A}}(\mathcal{C}_f \rightarrow \mathcal{C}_e) \\
& + \frac{1}{2}\bm{\mathcal{A}}(\mathcal{C}_g \rightarrow \mathcal{C}_d) + \frac{1}{2}\bm{\mathcal{A}}(\mathcal{C}_g \rightarrow \mathcal{C}_e) \\
& + \frac{1}{2}\bm{\mathcal{A}}(\mathcal{C}_d \rightarrow \mathcal{C}_f) 
+ \frac{1}{2}\bm{\mathcal{A}}(\mathcal{C}_e \rightarrow \mathcal{C}_f) \\
& + \frac{1}{2}\bm{\mathcal{A}}(\mathcal{C}_d \rightarrow \mathcal{C}_g)
+ \frac{1}{2}\bm{\mathcal{A}}(\mathcal{C}_e \rightarrow \mathcal{C}_g)
\end{aligned}
\end{equation}

Similar to the relation between sample $a$ and $c$ at time steps $t = 1,2$, sample $f$ and $h$ belong to the same cluster $\mathcal{C}_m$ at current time step while they are from different clusters at time step $t = 4$. Based on the approximation, the terms $\bm{\mathcal{A}}(\mathcal{C}_f \rightarrow \mathcal{C}_h)$ and $\bm{\mathcal{A}}(\mathcal{C}_h \rightarrow \mathcal{C}_f)$ in two time steps will be merged. As a result, the final loss is computed on intra-cluster pairs and inter-cluster pairs sampled from three clusters $\mathcal{C}_l$, $\mathcal{C}_j$ and $\mathcal{C}_m$. 

\item \textbf{t=6}: $\mathcal{C}^6_* = \mathcal{C}_l$, $\mathcal{N}_{\mathcal{C}^6_*}^{2}[1] = \mathcal{C}_j$ and $\mathcal{N}_{\mathcal{C}^6_*}^{2}[2] = \mathcal{C}_m$. Thus
\begin{equation}
\begin{aligned}
\mathcal{L}(\bm{\theta}|\{\bm{y}^1_*,...,\bm{y}^6_*\}, I) & = \mathcal{L}(\bm{\theta}|\{\bm{y}^1_*,...,\bm{y}^5_*\}, I) \\
& - \left(\lambda'\bm{\mathcal{A}}(\mathcal{C}_l, \mathcal{C}_j) - \bm{\mathcal{A}}(\mathcal{C}_l, \mathcal{C}_m)\right)
\end{aligned}
\end{equation}

Similar to the decomposition procedures above, both $\bm{\mathcal{A}}(\mathcal{C}_l, \mathcal{C}_j)$ and $\bm{\mathcal{A}}(\mathcal{C}_l, \mathcal{C}_m)$ can be transformed to sample-based affinities. Because $\mathcal{C}_l$ and $\mathcal{C}_j$ are regarded as different clusters previously, sample pairs from both of them are with positive weights in the loss function. However, it will be diminished by positive pairs (with negative weights) at current time step. 
\end{itemize}

Though we use a toy example to show that the cluster-based loss can be transformed to sample-based loss above, the reduction is general to any possible agglomerative clustering processes because the loss for clusters at high-level can always be decomposed to the losses on clusters at low-level until it reaches to single samples. The difference among various processes lies on the different weights associated with sample-based affinities. We should know that sample pairs from the same cluster may be with positive weights. One way to avoid this is increase $\lambda'$. In our implementation, we aim to increase affinities between samples from the same clusters, while decrease the affinities between samples from different clusters. And the clusters are determined by cluster ids at current step. Therefore, we assign a consistent weight $\gamma$ to any affinities from the same cluster and 1 to any affinities from different clusters. Because we use SGD for batch optimization, the scales for affinities do not affect much on the performance. It is the signs affect much. Accordingly, at any given time step $T$, the overall loss is approximated to
\begin{equation}
\mathcal{L}(\bm{\theta} |\bm{y}^T_*, \bm{I}) = -\frac{\lambda}{K_c - 1}\sum_{i,j,k}  \left(\gamma \bm{\mathcal{A}}(\bm{x}_i, \bm{x}_j)  - \bm{\mathcal{A}}(\bm{x}_i, \bm{x}_k)\right)
\label{Eq_Loss_triplet_wst_theta_sample}
\end{equation}

Note that we replace $\bm{\mathcal{Y}}_*$ in \eqref{Eq_Loss_triplet_wst_theta_sample} by $\bm{y}^T_*$ in \eqref{Eq_Loss_triplet_wst_theta_sample} because it is merely determined by current $\bm{y}^T$, regardless of $\{\bm{y}^1_*,...,\bm{y}^{T-1}_*\}$. As a result, we do not need to record $\{\bm{y}^1_*,...,\bm{y}^{T-1}_*\}$. This simplifies the batch optimization for CNN. Concretely, given a sample $\bm{x}_i$, we randomly select a sample $\bm{x}_j$ which belongs to the same cluster, while select neighbours of $\bm{x}_i$ that from other clusters to be $\bm{x}_k$. To omit the case that $\bm{\mathcal{A}}(\bm{x}_i, \bm{x}_j)$ is much larger than $\bm{\mathcal{A}}(\bm{x}_i, \bm{x}_k)$, we also add a margin threshold like the triplet loss function used in \cite{wang2014learning, schroff2015facenet}.

\subsection{Detailed CNN Architectures in our Paper}
\label{Ap_Detailed_CNN}

In this paper, the CNN architectures vary from dataset to dataset. As we mentioned in the main paper, we stacked different number of layers for different datasets so that the size of most top layer response map is about 10$\times$10. In Table \ref{TB_CNN_Archs}, we list the architectures for the datasets used in our paper. "conv" means convolutional layer. "bn" means batch normalization layer. "wt-loss" means weighted triplet loss layer. $\checkmark$ means the layer is used, while $-$ means the layer is not used.

\begin{table*}[t]
\caption{CNN architectures for different datasets in our paper.}
\vspace{-5pt}
\center
\small
\begin{tabular}{lccccccccccc}
  \toprule
   Dataset & \textit{COIL20} & \textit{COIL100} & \textit{USPS} & \textit{MNIST-test} & \textit{MNIST-full} & UMist & \textit{FRGC} & \textit{CMU-PIE} & \textit{YTF} \\   
\midrule
  conv1 & \checkmark & \checkmark & \checkmark & \checkmark & \checkmark & \checkmark & \checkmark & \checkmark & \checkmark\\
  
  bn1 &  \checkmark & \checkmark & \checkmark & \checkmark & \checkmark & \checkmark & \checkmark & \checkmark& \checkmark\\
  
  relu1 & \checkmark & \checkmark & \checkmark & \checkmark & \checkmark & \checkmark & \checkmark & \checkmark& \checkmark\\
  
  pool1 & \checkmark & \checkmark & \checkmark & \checkmark & \checkmark & \checkmark & \checkmark & \checkmark& \checkmark\\
  
  conv2 & \checkmark & \checkmark & $-$ & \checkmark & \checkmark & \checkmark & \checkmark & \checkmark& \checkmark\\
  
  bn2 &  \checkmark & \checkmark & $-$ & \checkmark & \checkmark & \checkmark & \checkmark & \checkmark & \checkmark\\
  
  relu2 & \checkmark & \checkmark & $-$ & \checkmark & \checkmark & \checkmark & \checkmark & \checkmark& \checkmark\\
  
  pool2 & \checkmark & \checkmark & $-$ & $-$ & $-$ & \checkmark & \checkmark  & \checkmark& \checkmark\\
  
  conv3 & \checkmark & \checkmark & $-$ & $-$ & $-$ & \checkmark & $-$ & $-$& $-$\\ 
  
  bn3 & \checkmark & \checkmark & $-$ & $-$ & $-$ & \checkmark & $-$ & $-$& $-$\\
  
  relu3 & \checkmark & \checkmark & $-$ & $-$ & $-$ & \checkmark & $-$ & $-$& $-$ \\
  
  pool3 & \checkmark & \checkmark & $-$ & $-$ & $-$ & \checkmark & $-$ & $-$ & $-$\\
  
  conv4 & \checkmark & \checkmark & $-$ & $-$ & $-$ & $-$ & $-$ & $-$& $-$\\
  
  bn4 & \checkmark & \checkmark & $-$ & $-$ & $-$ & $-$ & $-$ & $-$& $-$\\
  
  relu4 & \checkmark & \checkmark & $-$ & $-$ & $-$ & $-$  & $-$ & $-$& $-$\\
  
  pool4 & \checkmark & \checkmark & $-$ & $-$ & $-$ & $-$ & $-$ & $-$& $-$\\
  
  ip1 & \checkmark & \checkmark & \checkmark & \checkmark & \checkmark & \checkmark & \checkmark & \checkmark& \checkmark\\
  
  l2-norm & \checkmark & \checkmark & \checkmark & \checkmark & \checkmark & \checkmark & \checkmark & \checkmark& \checkmark\\
  
  wt-loss & \checkmark & \checkmark & \checkmark & \checkmark & \checkmark & \checkmark & \checkmark & \checkmark& \checkmark\\
  \bottomrule
\end{tabular}
\label{TB_CNN_Archs}
\end{table*}

\subsection{Performance Evaluated by Accuracy}
In this section, we evaluate the performance of different algorithms based on clustering accuracy (AC) metric, as a supplement to the NMI metric used in our main paper. As we can see from table~\ref{TB_Quantitative_Results_AC}, the proposed method outperform other methods on all datasets, which has similar trend as evaluated using NMI. Meanwhile, according to table~\ref{TB_Quantitative_Results_withRep_AC}, all other clustering algorithms are boosted after using the learned representation as evaluated on AC. These results further prove the proposed method is superior to other clustering algorithms and also learns powerful deep representations that generalize well across different clustering algorithms.

\begin{table*}[!ht]
\caption{{Quantitative clustering performance (AC) for different algorithms using image intensities as input.}}
\vspace{-5pt}
\center
\small
\begin{tabular}{@{}lccccccccc@{}}
  \toprule
   Dataset & \textit{COIL20} & \textit{COIL100} & \textit{USPS} & \textit{MNIST-test} & \textit{MNIST-full} & \textit{UMist} & \textit{FRGC} & \textit{CMU-PIE} & \textit{YTF} \\   
     
  \midrule
  K-means \cite{macqueen1967some}  &0.665 &0.580 &0.467 &0.560 &0.564 &0.419 &0.327 &0.246 &0.548 \\
  
  SC-NJW \cite{ng2002spectral}     &0.641 &0.544 &0.413  &0.220 &0.502 &0.551 &0.178 &0.255 &0.551 \\
      
  SC-ST \cite{zelnik2004self}      &0.417 &0.300 &0.308    
&0.454 &0.311 &0.411 &0.358 &0.293 &0.290  \\
  
  SC-LS \cite{chen2011large}    &0.717 &0.609 &0.659    
&0.740 &0.714 &0.568 &0.407 &0.549 &0.544 \\
  
  N-Cuts \cite{shi2000normalized}  &0.544 &0.577 &0.314   
&0.304 &0.327 &0.550 &0.235 &0.155 &0.536 \\
  
  AC-Link \cite{jain1999data}      &0.251 &0.269  &0.421    
&0.693 &0.657 &0.398 &0.175 &0.201 &0.547 \\
    
  AC-Zell \cite{zhao2009cyclizing} &0.867 &0.811  &0.575    
&0.693 &0.112 &0.517 &0.266 &0.765 &0.519  \\
  
  AC-GDL \cite{zhang2012graph}     &0.865 &0.797  &0.867   
&0.933 &0.113 &0.563 &0.266 &0.842 &0.430  \\   
  
  AC-PIC \cite{zhang2013agglomerative} &0.855 &0.840 &0.855    &0.920 &0.115 &0.576 &0.320 &0.797 &0.472 \\
  
  NMF-LP \cite{cai2009locality}    &0.621 &0.553 &0.522    
&0.479 &0.471 &0.365 &0.259 &0.229 &0.546 \\  
 {OURS-SF} &\textbf{1.000} & {0.894} & {0.922}   & {0.940} & 0.959 & \textbf{0.809} & \textbf{0.461}        & {0.980}    &  \textbf{0.684}   \\
 
 {OURS-RC} &\textbf{1.000} & \textbf{0.916} & \textbf{0.950}   & \textbf{0.961} &\textbf{0.964}  & \textbf{0.809} & \textbf{0.461}        & \textbf{1.000}    &  \textbf{0.684}   \\
  \bottomrule
  
\end{tabular}
\label{TB_Quantitative_Results_AC}
\end{table*}

\begin{table*}[!ht]
\caption{{Quantitative clustering performance (AC) for different algorithms using our learned representations as inputs.}}
\vspace{-5pt}
\center
\small
\begin{tabular}{@{}lccccccccccc}
  \toprule
   Dataset & \textit{COIL20} & \textit{COIL100} & \textit{USPS} & \textit{MNIST-test} & \textit{MNIST-full} & \textit{UMist} & \textit{FRGC} & \textit{CMU-PIE} & \textit{YTF} \\   
  \midrule
  K-means \cite{macqueen1967some}  &0.821 &0.751 &0.776 &0.957 &0.969 &0.761 &0.476 &0.834 &0.660 \\
  
  SC-NJW \cite{ng2002spectral}     &0.738 &0.659 &0.716 &0.868 &0.972 &0.707 &0.485 &0.776 &0.521 \\
      
  SC-ST \cite{zelnik2004self}      &0.851 &0.705 &0.661 &0.960 &0.958 &0.697 &0.496 &0.896 &0.575 \\   
  
  SC-LS \cite{chen2011large}    &0.867 &0.735 &0.792 &0.960 &\textbf{0.973} &0.733 &0.502 & 0.802 &0.571 \\
  
  N-Cuts \cite{shi2000normalized}  &0.888 &0.626 &0.634 &0.959 &0.971 &\textbf{0.798} &\textbf{0.504} &0.981 &0.441   \\
  
  AC-Link \cite{jain1999data}      &0.678 &0.539 &0.773 &0.955 &0.964 &0.795 &0.495 &0.947 &0.602 \\
    
  AC-Zell \cite{zhao2009cyclizing} &\textbf{1.000} &0.931 &0.879 &0.879 &0.969 &0.790 &0.449 &\textbf{1.000} &0.644  \\
  
  AC-GDL \cite{zhang2012graph}     &\textbf{1.000} &0.920 &0.949 &\textbf{0.961} &0.878 & 0.790 &0.461 &\textbf{1.000} &\textbf{0.677}  \\   
  
  AC-PIC \cite{zhang2013agglomerative}&\textbf{1.000} &\textbf{0.950} &\textbf{0.955} &0.958 &0.882 &0.790 &0.438  &\textbf{1.000} &0.652 \\
  
  NMF-LP \cite{cai2009locality}    &0.769 &0.603 &0.778 &{0.955} &0.970 & 0.725 & 0.481 &0.504 &0.575 \\
  \bottomrule
\end{tabular}
\label{TB_Quantitative_Results_withRep_AC}
\end{table*}

\subsection{Robustness Analysis}
\begin{figure*}[t]
\begin{minipage}{0.49\linewidth}
\centering
\includegraphics[trim=0.5in 0.0in 0.5in 0.0in, clip=true, scale=0.36]{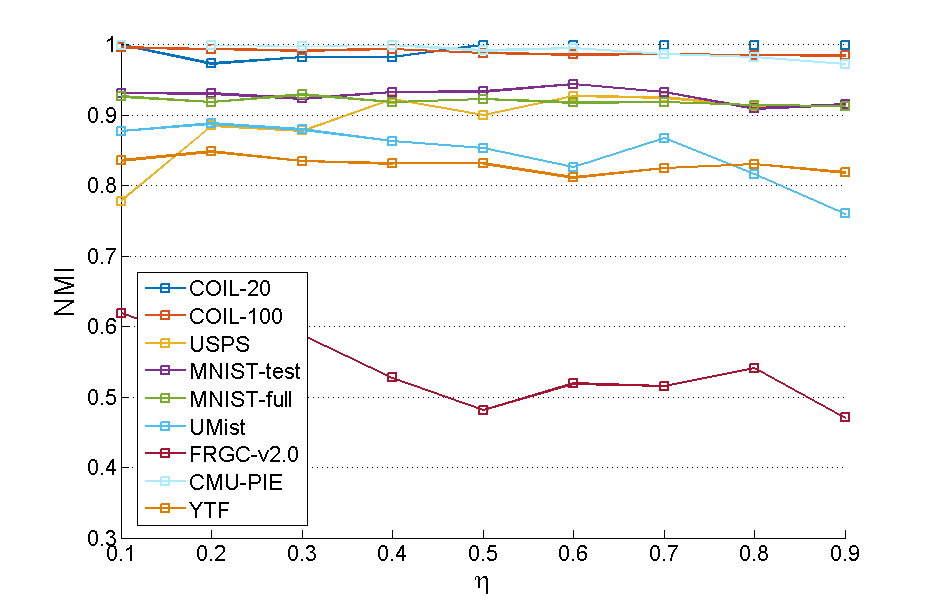}
\end{minipage}
\begin{minipage}{0.49\linewidth}
\centering
\includegraphics[trim=0.5in 0.0in 0.5in 0.0in, clip=true, scale=0.36]{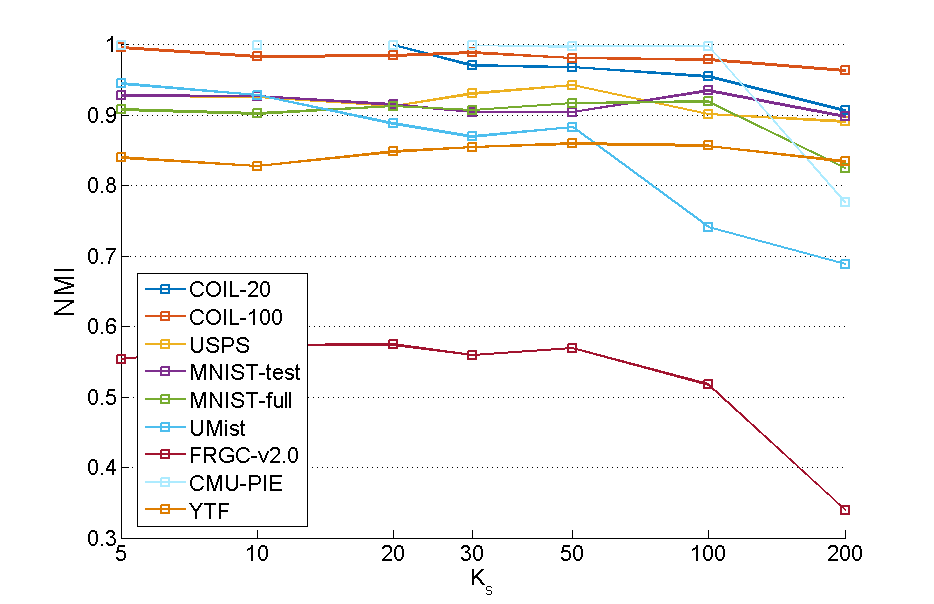}
\end{minipage}
\vspace{-1pt}
\caption{Clustering performance (NMI) with different $\eta$ (left) and $K_s$ (right).}
\label{Fig_NNI_parameters}
\end{figure*}
We choose the two most important parameters: unfolding rate $\eta$ and $K_s$ for evaluating the robustness of our approach to variations in these parameters. In these experiments, we set all the other parameters except for the target one to default values listed in Table 2 in the main paper. As we can see from Fig.~\ref{Fig_NNI_parameters}, when the unfolding rate increases, the performance is not affected much for most of the datasets. For $K_s$, the performance is stable when $K_s <= 50$ for all datasets. It drops with larger values of $K_s$ for a few datasets. Increasing $K_s$ also result in similar degradation in the agglomerative clustering algorithms we compare to. This suggests that $K_s$ should not be set to very large value in general.

\subsection{Reliability Analysis}
\begin{figure*}[t]
\begin{minipage}{0.49\linewidth}
\centering
\includegraphics[trim=0.5in 0.0in 0.5in 0.0in, clip=true, scale=0.34]{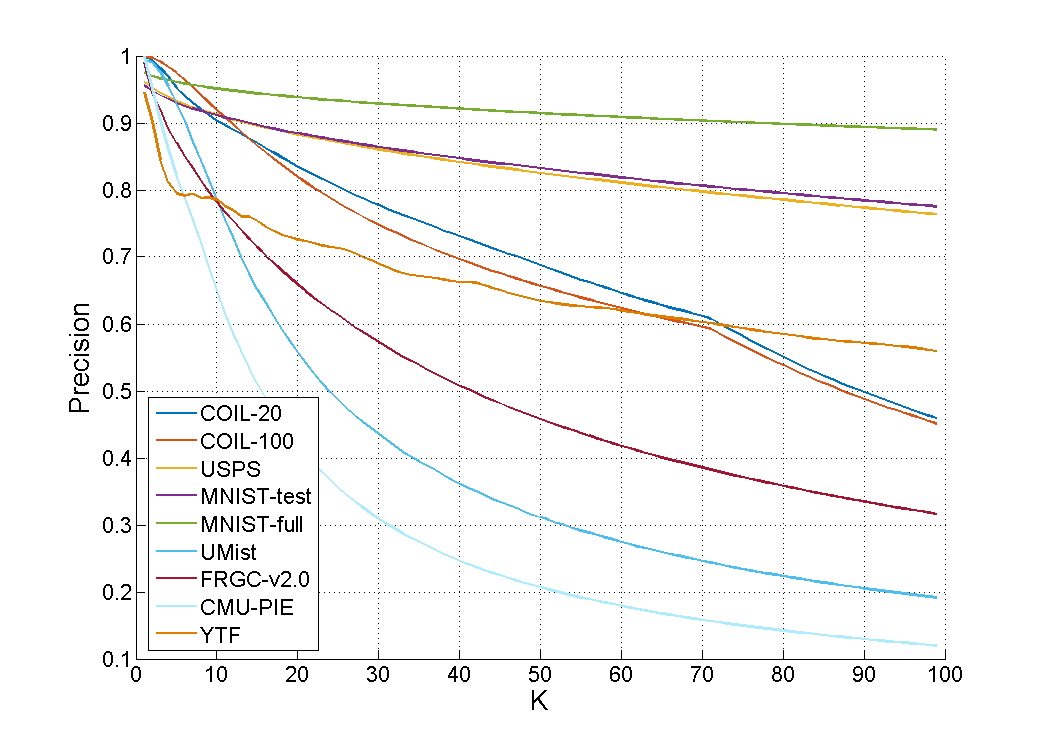}
\end{minipage}
\begin{minipage}{0.49\linewidth}
\centering
\includegraphics[trim=0.5in 0.0in 0.5in 0.0in, clip=true, scale=0.34]{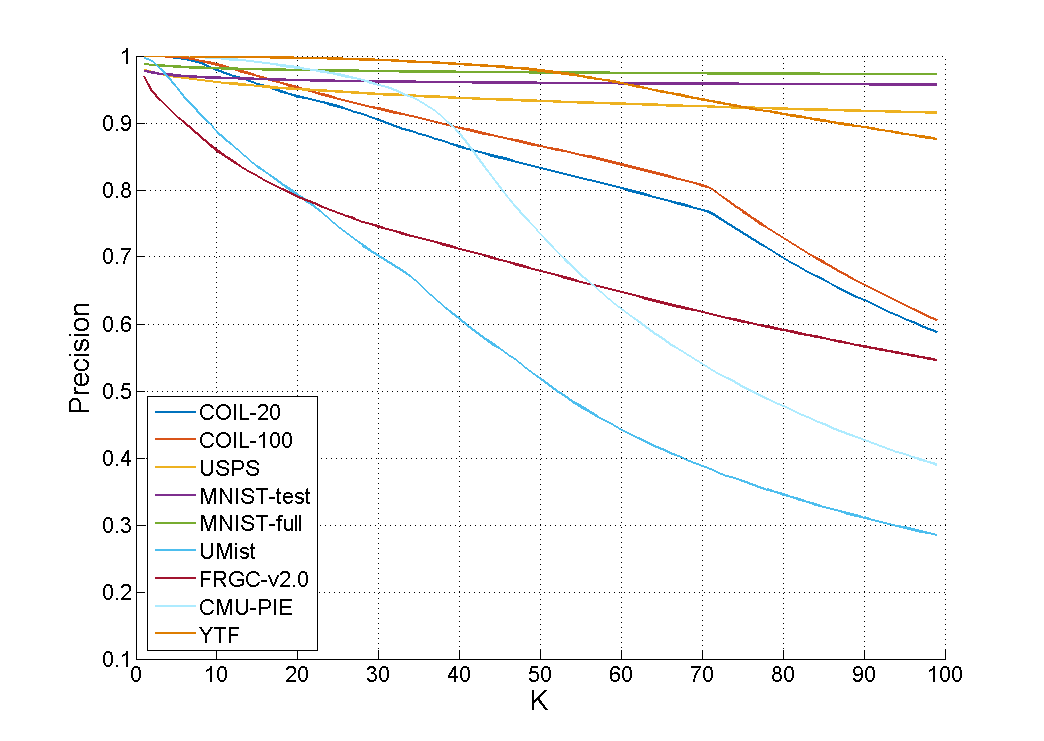}
\end{minipage}
\vspace{-3pt}
\caption{Average purity of K-nearest neighbour for varying values of $K$. Left is computed using raw image data, while right is computed using our learned representation.}
\label{Fig_K_Precision}
\end{figure*}
We evaluate the reliability by measuring the purity of samples at the beginning of our algorithm. Because we use agglomerative clustering, there are very few samples in each cluster at the beginning (average is about 4 in our experiments). Most samples in the same cluster tend to belong to the same category. Quantitatively, for each sample in a dataset, we count the number of samples ($K_{m}$) that belong to the same category within its $K$ nearest neighbours, and then compute the precision $K_{m} / K$ for it. In Fig.~\ref{Fig_K_Precision}, we report the average precision across all samples. As we can see, based on raw image data, all datasets have high ratios when $K$ is smaller, and the ratios increase further when using our learned deep representations. Consequently, when $K$ is small, the pseudo-labels are reliable enough to learn plausible deep representations.

\subsection{Clustering based on Hand-Crafted Features}
\label{Ap_Cluster_Hand_Crafted}
We also evaluate the performance of clustering based on image features, instead of image intensities. We choose three different types of datasets for testing: COIL100, MNIST-test and UMist, and three types of clustering algorithms including SC-LS \cite{chen2011large}, N-Cuts \cite{shi2000normalized} and AC-PIC \cite{zhang2013agglomerative} for comparison since their better performance among all the algorithms. For these three datasets, we use spatial pyramid descriptor \cite{lazebnik2006beyond}\footnote{\url{http://slazebni.cs.illinois.edu/research/SpatialPyramid.zip}}, histogram of oriented gradient (HOG) \cite{dalal2005histograms}\footnote{\url{http://www.robots.ox.ac.uk/~vgg/research/caltech/phog.html}} and local binary pattern (LBP) \cite{ojala1996comparative} for representation, respectively. We report the results in Table~\ref{TB_CLustering_Results_Feats}. $\downarrow$ means performance become worse, and $\uparrow$ means it become better. Almost all algorithms perform worse than using original image as input. It indicates hand-crafted features should be designed dataset by dataset. In contrast, directly learning from image intensities is more straightforward and also achieves better performance.
\begin{table}
\caption{{Clustering performance (NMI) based on hand-crafted features.}}
\vspace{-5pt}
\center
\small
\begin{tabular}{@{}lcccc@{}}
  \toprule
  Dataset                              &COIL100 &MNIST-test &UMist &FRGC    \\   
  \midrule  
  SC-LS \cite{chen2011large}        &0.733$\downarrow$   &0.625$\downarrow$       &{0.752}$\downarrow$ &0.338$\downarrow$ \\
  
  N-Cuts \cite{shi2000normalized}      &0.722$\downarrow$   &0.423$\uparrow$      &{0.420}$\downarrow$ &0.238$\downarrow$ \\  
  
  AC-PIC \cite{zhang2013agglomerative} &0.878$\downarrow$   &0.735$\downarrow$      &{0.734}$\downarrow$ &0.322$\downarrow$ \\  
  \bottomrule
\end{tabular}
\label{TB_CLustering_Results_Feats}
\end{table}
\newpage
\subsection{Visualizing Data in Low Dimension}
\begin{figure*}[t]
   \begin{subfigure}{0.24\linewidth}
   \centering
    \includegraphics[trim=0cm 0cm 0cm 0cm, clip=true, scale=0.2]{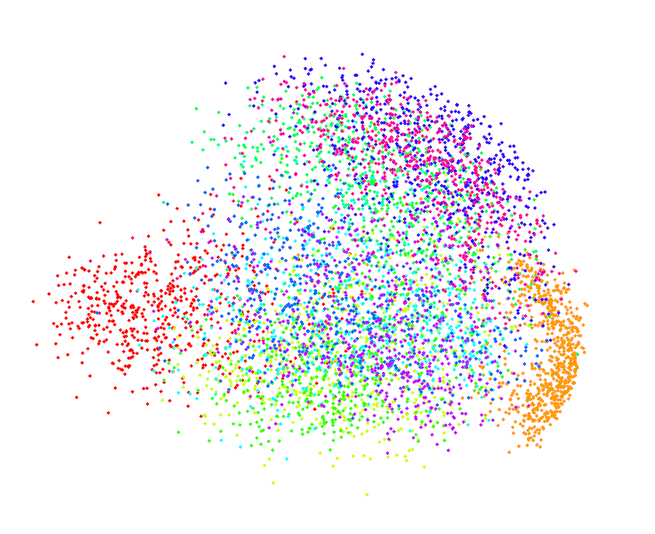}
    \vspace{-5pt}
    \caption{PCA.}
       \vspace{20pt}
   \end{subfigure}
   \begin{subfigure}{0.24\linewidth}
   \centering
    \includegraphics[trim=1cm 0cm 0cm 1cm, clip=true, scale=0.2]{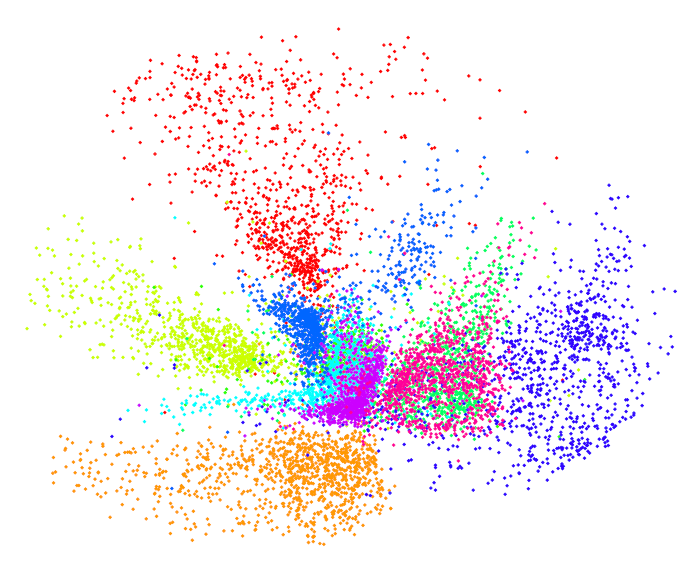}
    \vspace{-5pt}
    \caption{Autoencoder.}
       \vspace{20pt}
   \end{subfigure}   
   \begin{subfigure}{0.24\linewidth}
   \centering
    \includegraphics[trim=1cm 0cm 0cm 1cm, clip=true, scale=0.2]{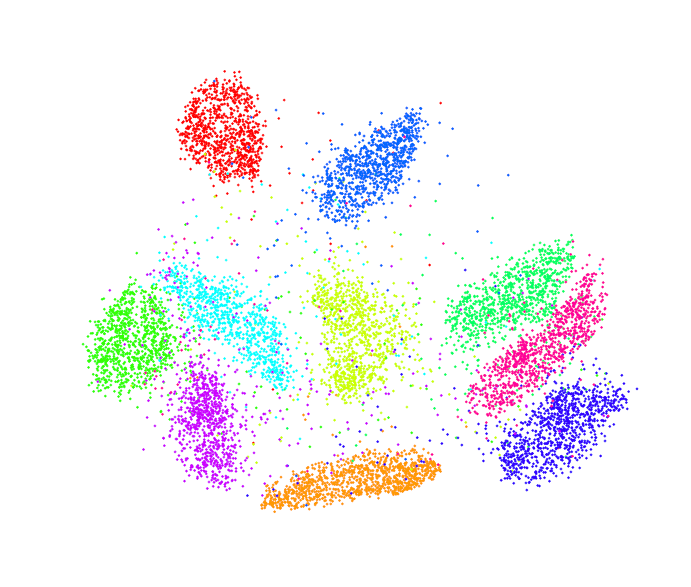}
    \vspace{-5pt}
    \caption{Parametric t-SNE.}
       \vspace{20pt}
   \end{subfigure}
   \begin{subfigure}{0.24\linewidth}
   \centering
       \includegraphics[trim=4cm 3cm 5cm 5cm, clip=true, scale=0.12]{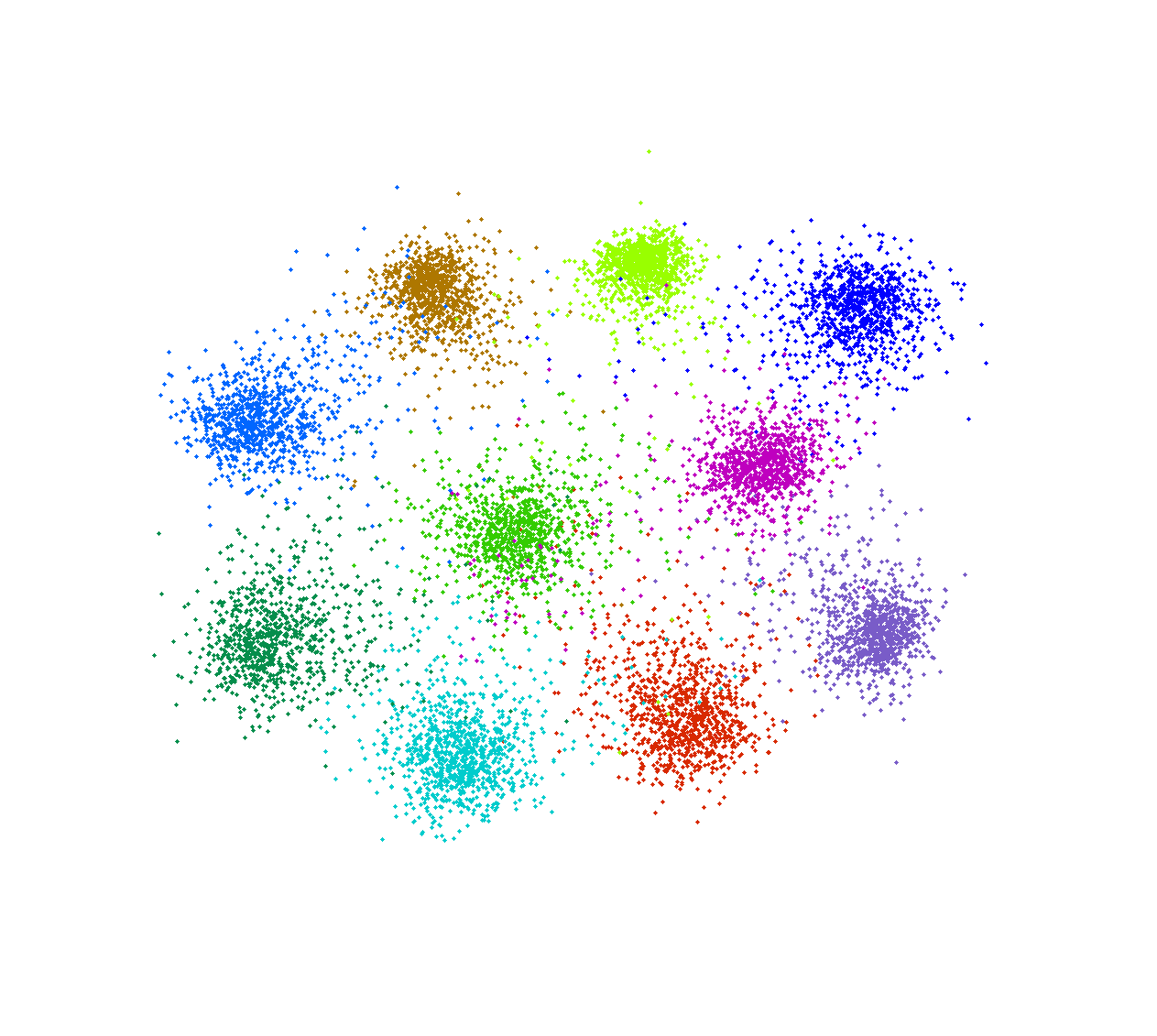}
   \vspace{-5pt}
   \caption{Visualization by our method.}
      \vspace{20pt}
   \end{subfigure}
      \vspace{-5pt}
   \caption{Visualization of 10,000 MNIST test samples in different embedding spaces.}
   \label{Fig_PCA_Display_6}
\end{figure*}
Projecting high-dimensional data into low-dimensional space can help people to intuitively understand the data. Though the proposed method is aimed to learn deep representations and image clusters, we note that it can be naturally converted to a parametric visualization method for an image dataset by slightly alternating the objective. Instead of updating the affinities among samples based on the learned representations gradually, we consistently use the affinities among raw image data to perform the agglomerative cluster, which then guides representation learning in low-dimensional space. By this way, we can obtain a low-dimensional space (2D or 3D) which can retain the structure of the original data. 

We compare three dimension reduction techniques, principle component analysis (PCA) \cite{wold1987principal}, neighbourhood components analysis (NCA) \cite{roweis2004neighbourhood}, and parametric t-SNE \cite{maaten2009learning}. Though both \cite{maaten2009learning} and our visualization method are based on neural networks, there are two main differences: 1) In \cite{maaten2009learning}, a Kullback-Leibler divergence between the joint distributions of original data  and the embedded data is considered. However, in our method, we employ a weighted triplet loss that directly takes the local structure of embedded data into account; 2) In \cite{maaten2009learning}, the authors need to pretrain a stack of RBMs layer-by-layer, and then fine-tune the neural network. Nevertheless, we directly train the neural network from scratch end-to-end. 

We perform experiments on MNIST dataset. In MNIST, 60,000 training samples are used to learn the low-dimensional embedding space, and 10,000 test samples are used for evaluation. To train a $D$-dimensional embedding, we first remove the normalization layer and then stack on the top another linear layer whose dimension is $D$. To thoroughly explore the local structure in the original data, we merge the clusters with a lower unfolding rate ($\eta = 0.2$). The learning process is stopped when the number of clusters reaches to 10. Though we stop the learning process as such, it should be noted that the stop criterion is not confined. For quantitative analysis, we compute the nearest-neighbor classification error and trustworthiness as in \cite{maaten2009learning}.

In Table~\ref{TB_1NN_Classification}, we show the 1-nearest neighbour classification error on MNIST test dataset. We copy the best results of the compared methods from \cite{maaten2009learning}. As we can see, our method outperforms all three other methods across three different embedding dimensions. These results illustrates that our method can obtain low-dimensional embedding with better generalization ability.

For visualization, it is important to retain the original data structure in the embedding space. For quantitative comparison, we report the trustworthiness of learned low-dimensional embedding in Table~\ref{TB_Trustworthiness}. Larger value means better preservation of original data structure. As we can see, our method is not as good as parametric t-SNE. These results are explainable. During training, we merely pay attention to the local structure among samples from different clusters, while omitting the relations among samples within one cluster. Therefore, the algorithm will learn embeddings that discriminate clusters well but possibly disorder the samples in each cluster. We believe this can be solved by introducing a loss to confine the within-cluster structure. We leave this as a future work for limited space.

\begin{table}[t]
\caption{1-nearest neighbor classification error on low-dimensional embedding of MNIST dataset.}
\vspace{-10pt}
\center
\begin{tabular}{@{}lcccc@{}}
  \toprule
  Method   &2D &10D &30D   \\   
  \midrule  
  PCA \cite{wold1987principal}           &0.782  &0.430  &0.108  \\
  NCA \cite{roweis2004neighbourhood}     &0.568  &0.088  &0.073  \\
  Autoencoder \cite{hinton2006reducing}  &0.668  &0.063
&\textbf{0.027} \\
  Param. t-SNE \cite{maaten2009learning} &0.099  &0.046  &\textbf{0.027}  \\  
  OURS       &\textbf{0.067}  &\textbf{0.019}  & \textbf{0.027} \\
  \bottomrule
\end{tabular}
\label{TB_1NN_Classification}
\end{table}

\begin{table}[t]
\caption{Trustworthiness T(12) on low-dimensional embedding of MNIST dataset.}
\vspace{-10pt}
\center
\begin{tabular}{@{}lcccc@{}}
  \toprule
  Method   &2D &10D &30D   \\   
  \midrule  
  PCA \cite{wold1987principal}          &0.744  &0.991  &0.998  \\
  NCA \cite{roweis2004neighbourhood}    &0.721  &0.968  &0.971  \\
  Autoencoder \cite{hinton2006reducing} &0.729  &0.996
&\textbf{0.999}\\
  Param. t-SNE \cite{maaten2009learning}&\textbf{0.927}  &\textbf{0.997}  &\textbf{0.999}  \\  
  Ours      &0.768  &0.936  &0.975  \\
  \bottomrule
\end{tabular}
\label{TB_Trustworthiness}
\end{table}

\subsection{Visualizing Learned Deep Representations}
\label{Ap_Displaying}
\begin{figure*}[t]
   \begin{subfigure}{0.33\linewidth}
   \centering
    \includegraphics[trim=6cm 0cm 5cm 4cm, clip=false, scale=0.2]{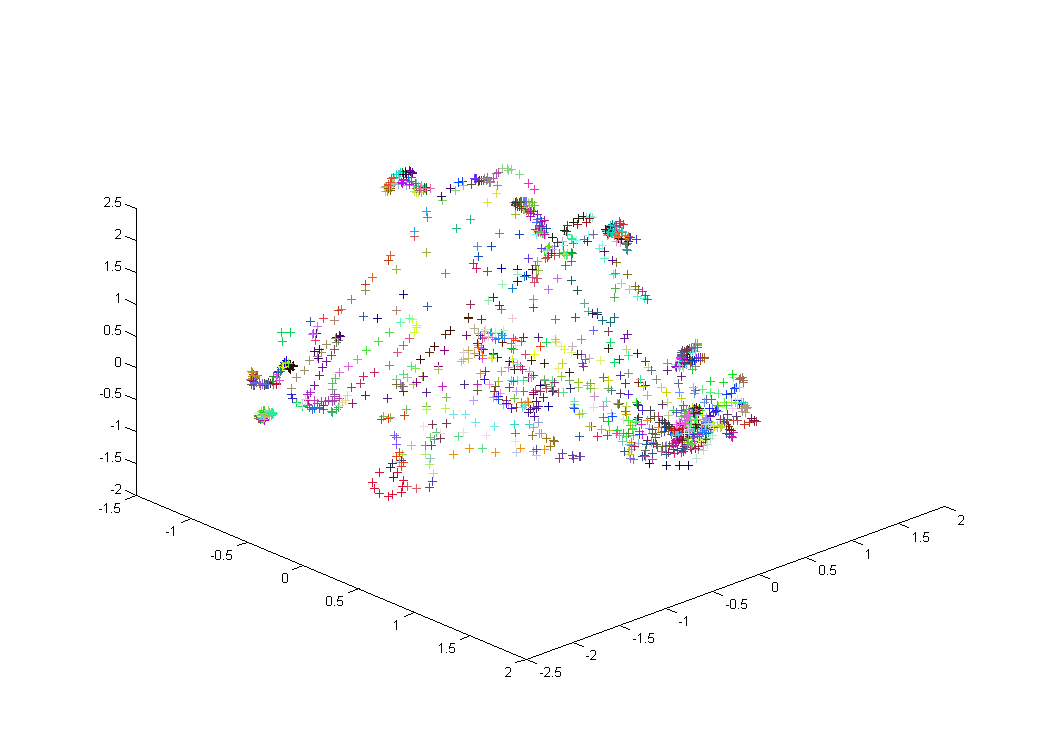}
    \vspace{-5pt}
    \caption{Initial stage (421)}
           \vspace{20pt}
   \end{subfigure}
   \begin{subfigure}{0.33\linewidth}
   \centering
    \includegraphics[trim=6cm 0cm 3cm 4cm, clip=false, scale=0.2]{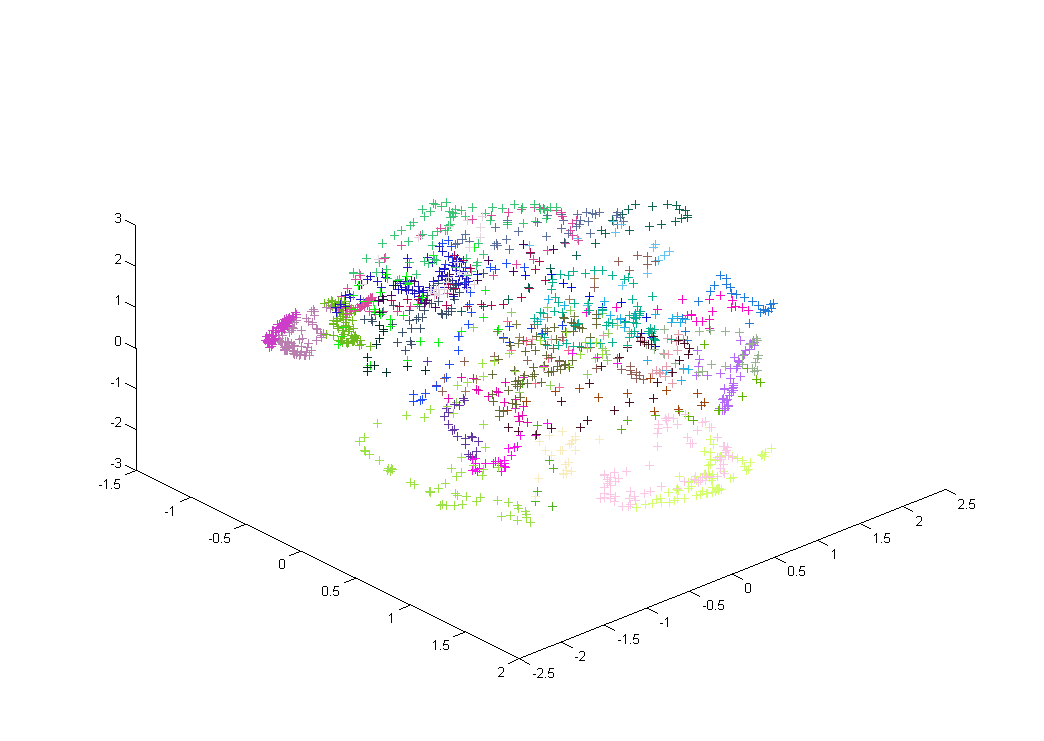}
    \vspace{-5pt}
    \caption{Middle stage (42)}
       \vspace{20pt}
   \end{subfigure}
   \begin{subfigure}{0.33\linewidth}
   \centering
   \includegraphics[trim=6cm 0cm 5cm 4cm, clip=false, scale=0.2]{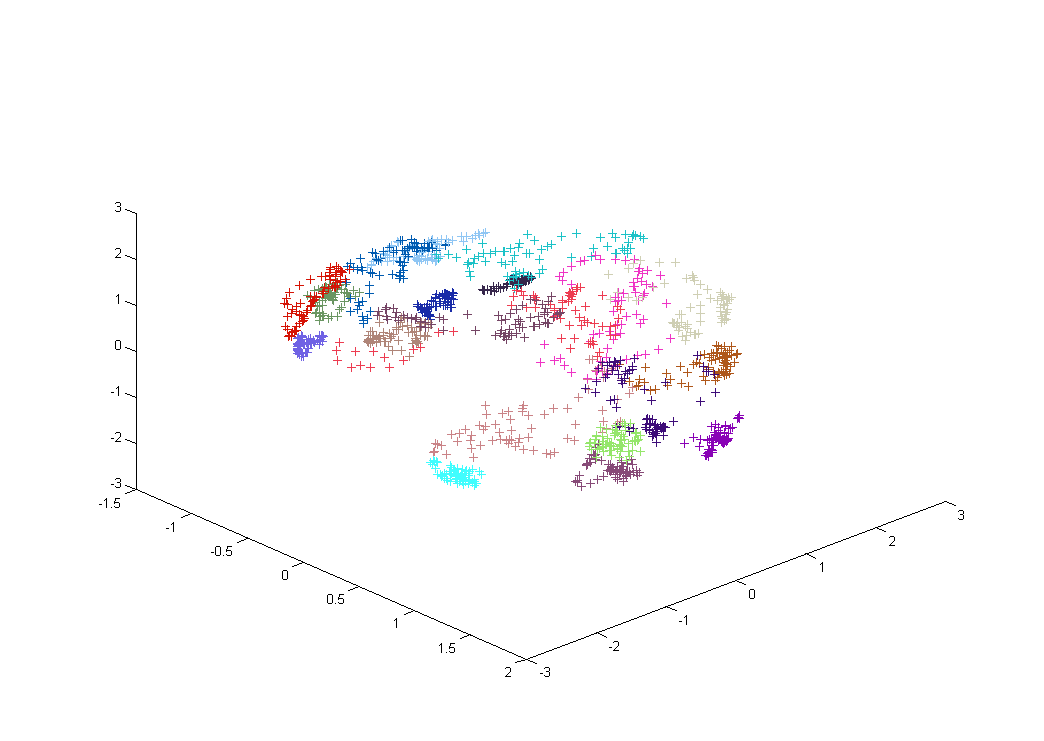}
   \vspace{-5pt}
   \caption{Final stage (20)}
   \vspace{20pt}
   \end{subfigure}
   
   \begin{subfigure}{0.33\linewidth}
   \centering
    \includegraphics[trim=6cm 0cm 5cm 4cm, clip=false, scale=0.2]{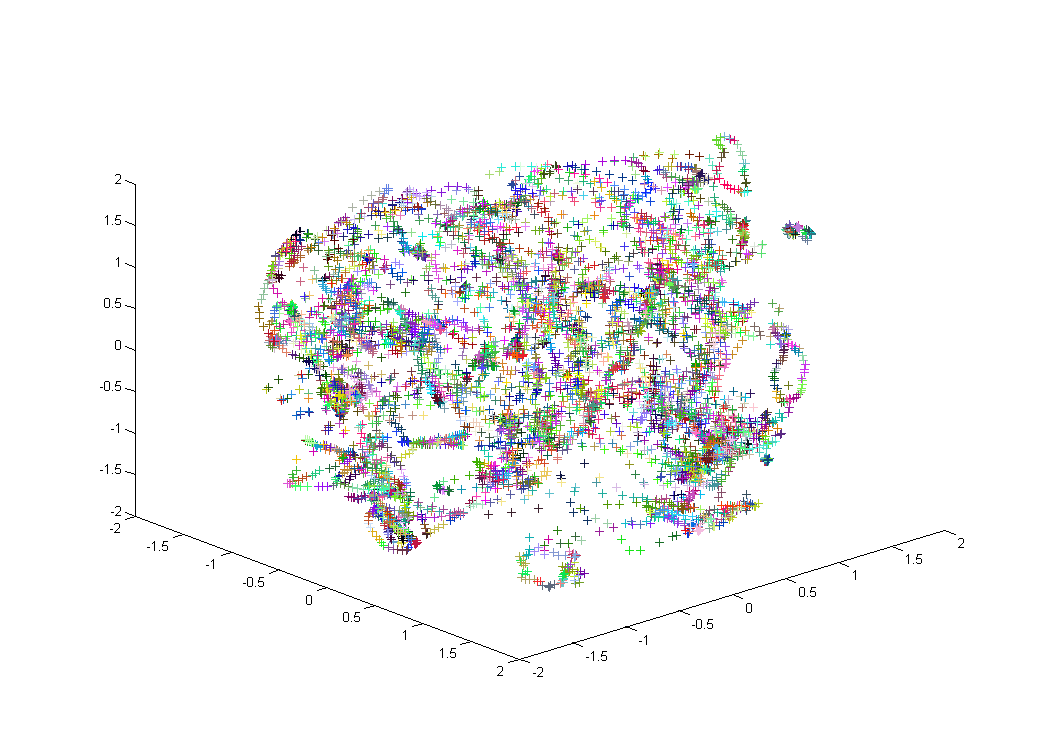}
    \vspace{-5pt}
    \caption{Initial stage (2162)}
           \vspace{20pt}
   \end{subfigure}
   \begin{subfigure}{0.33\linewidth}
   \centering
    \includegraphics[trim=6cm 0cm 3cm 4cm, clip=false, scale=0.2]{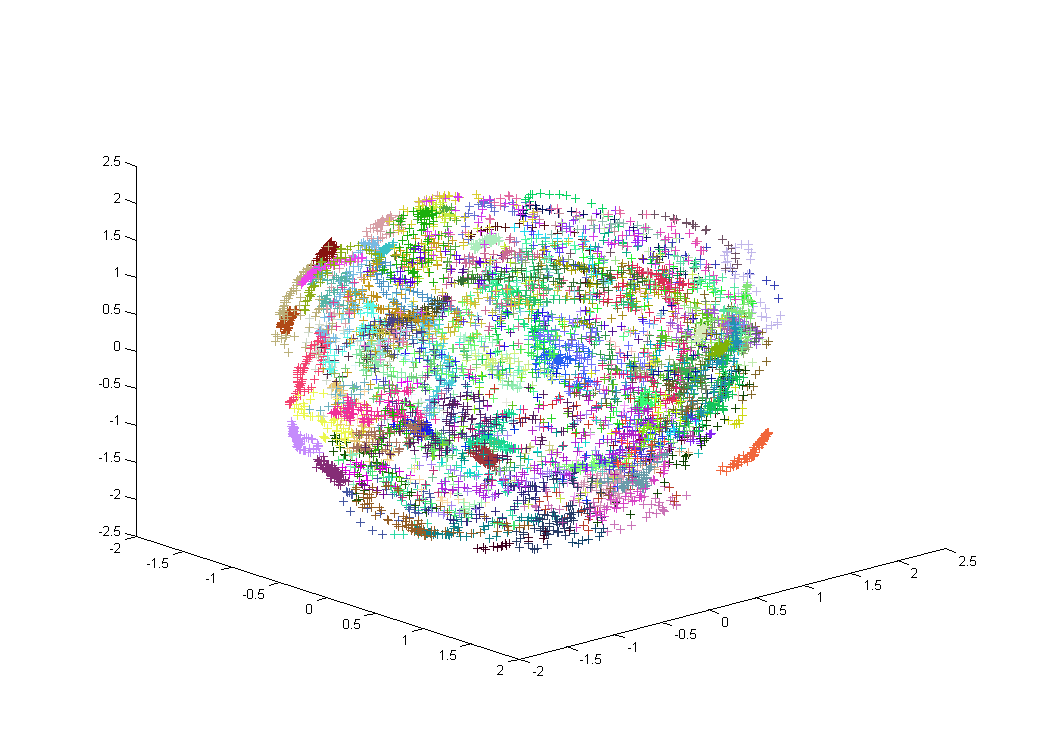}
    \vspace{-5pt}
    \caption{Middle stage (216)}
           \vspace{20pt}
   \end{subfigure}
   \begin{subfigure}{0.33\linewidth}
   \centering
       \includegraphics[trim=6cm 0cm 5cm 4cm, clip=false, scale=0.2]{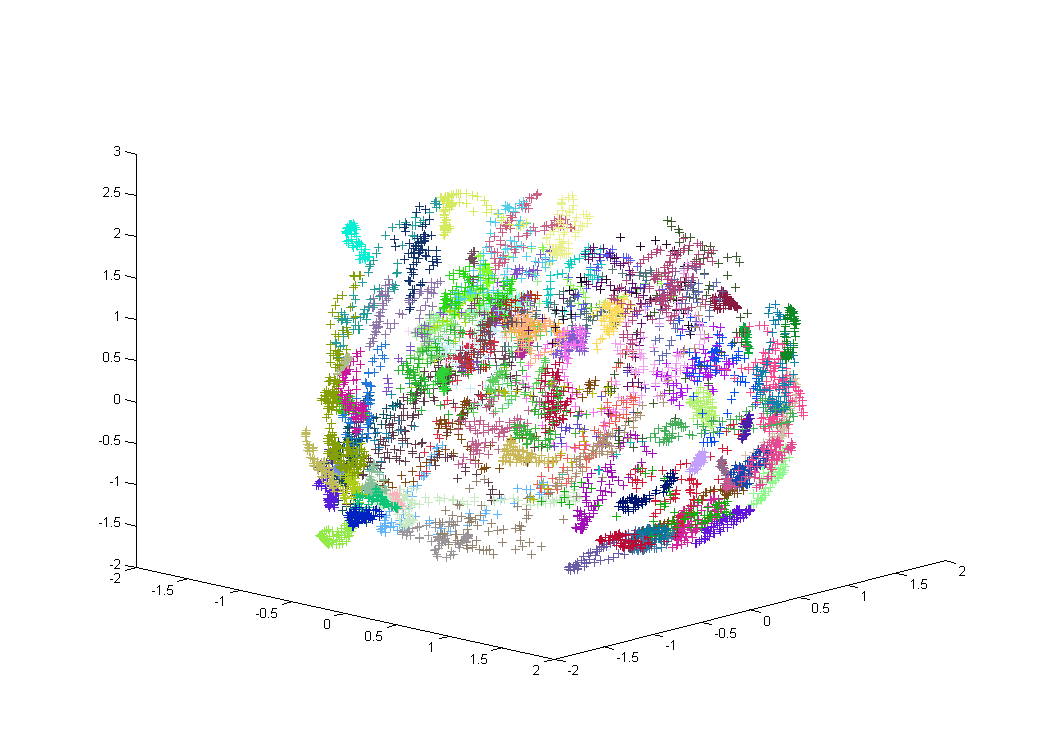}
   \vspace{-5pt}
   \caption{Final stage (100)}
          \vspace{20pt}
   \end{subfigure}

   \begin{subfigure}{0.33\linewidth}
   \centering
    \includegraphics[trim=6cm 0cm 5cm 4cm, clip=false, scale=0.2]{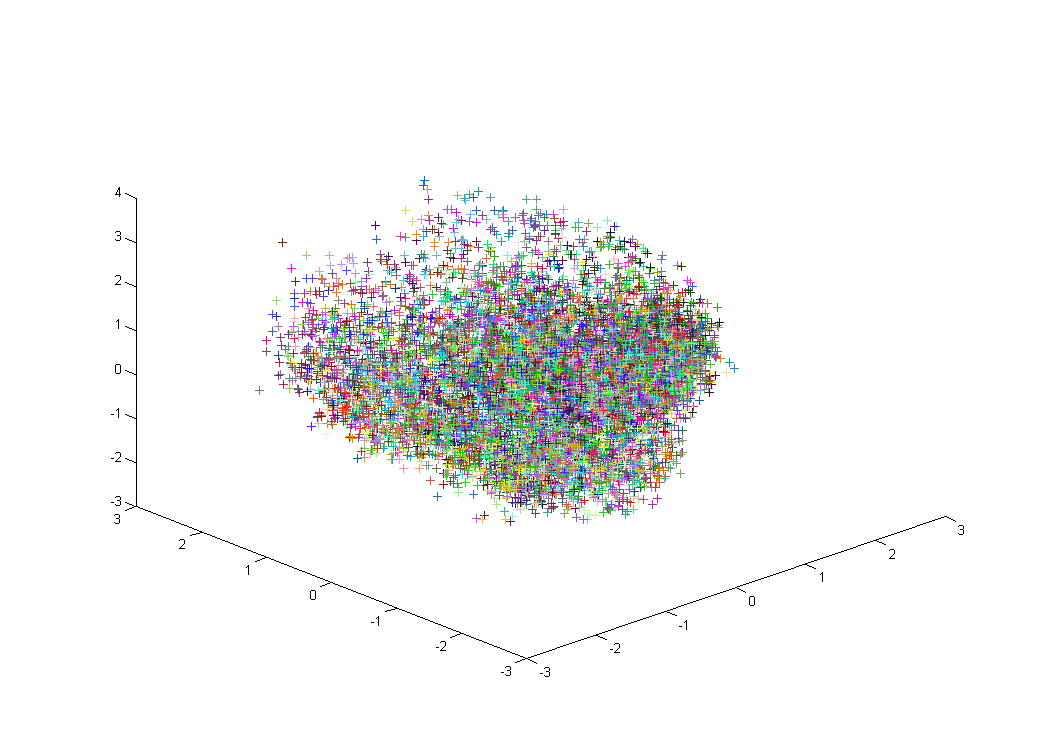}
    \vspace{-5pt}
    \caption{Initial stage (2232)}
    \vspace{20pt}
   \end{subfigure}
   \begin{subfigure}{0.33\linewidth}
   \centering
    \includegraphics[trim=6cm 0cm 3cm 4cm, clip=false, scale=0.2]{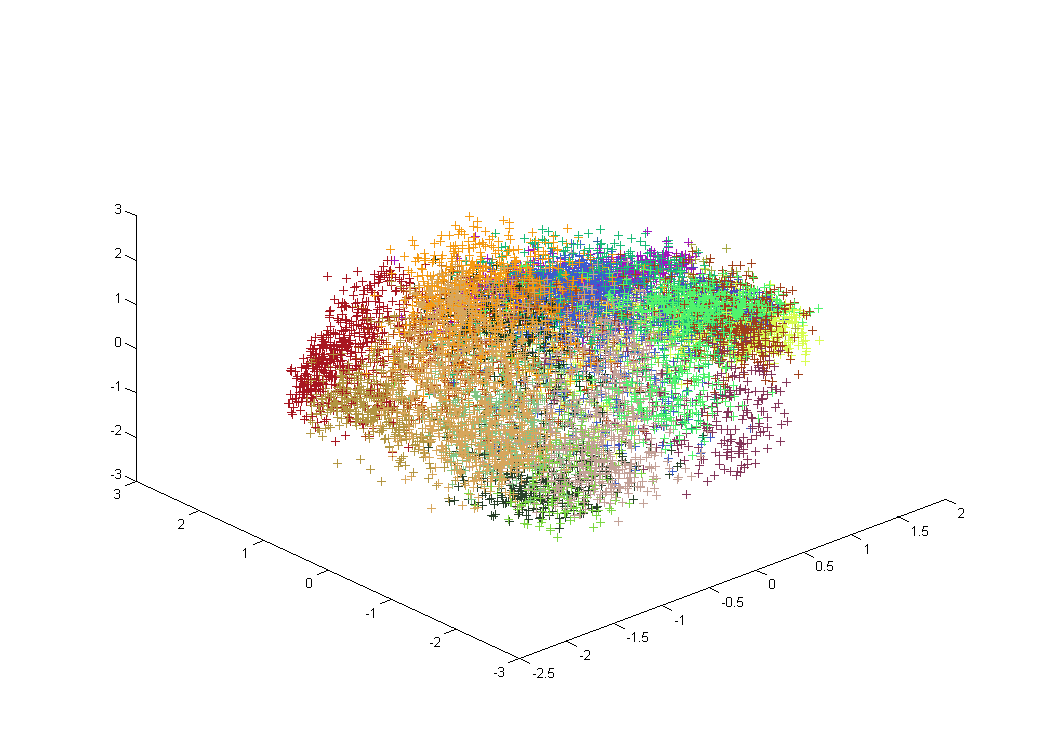}
    \vspace{-5pt}
    \caption{Middle stage (22)}
    \vspace{20pt}
   \end{subfigure}
   \begin{subfigure}{0.33\linewidth}
   \centering
       \includegraphics[trim=6cm 0cm 5cm 4cm, clip=false, scale=0.2]{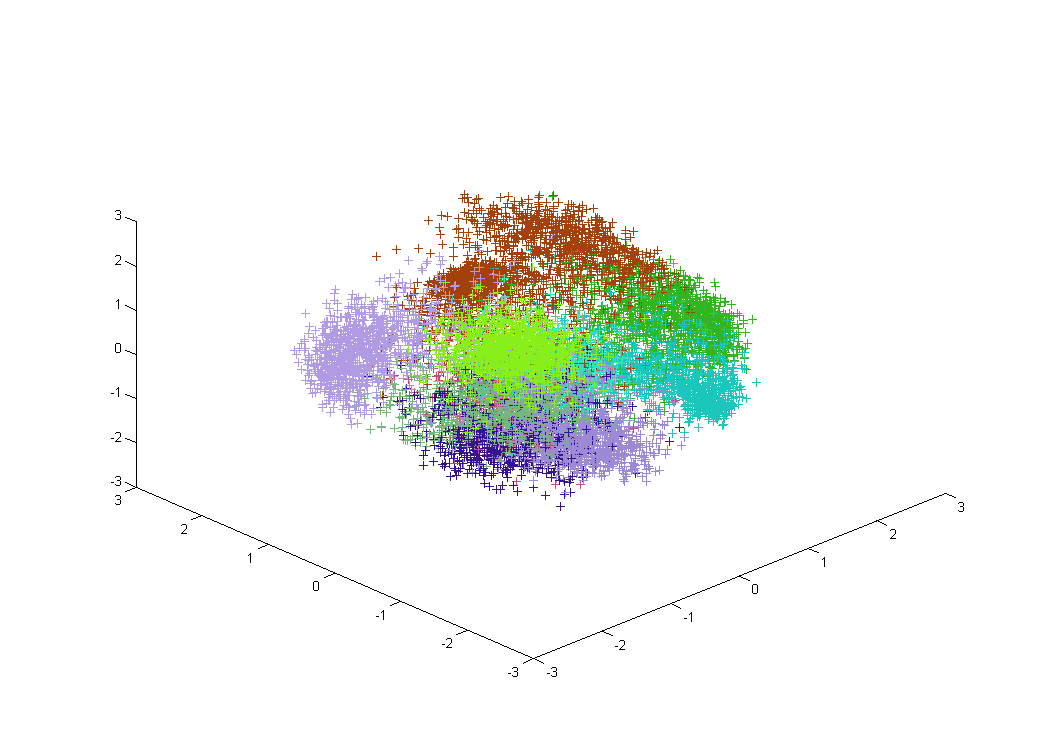}
   \vspace{-5pt}
   \caption{Final stage (10)}
   \vspace{20pt}
   \end{subfigure}

   \begin{subfigure}{0.33\linewidth}
   \centering
    \includegraphics[trim=6cm 0cm 5cm 4cm, clip=false, scale=0.2]{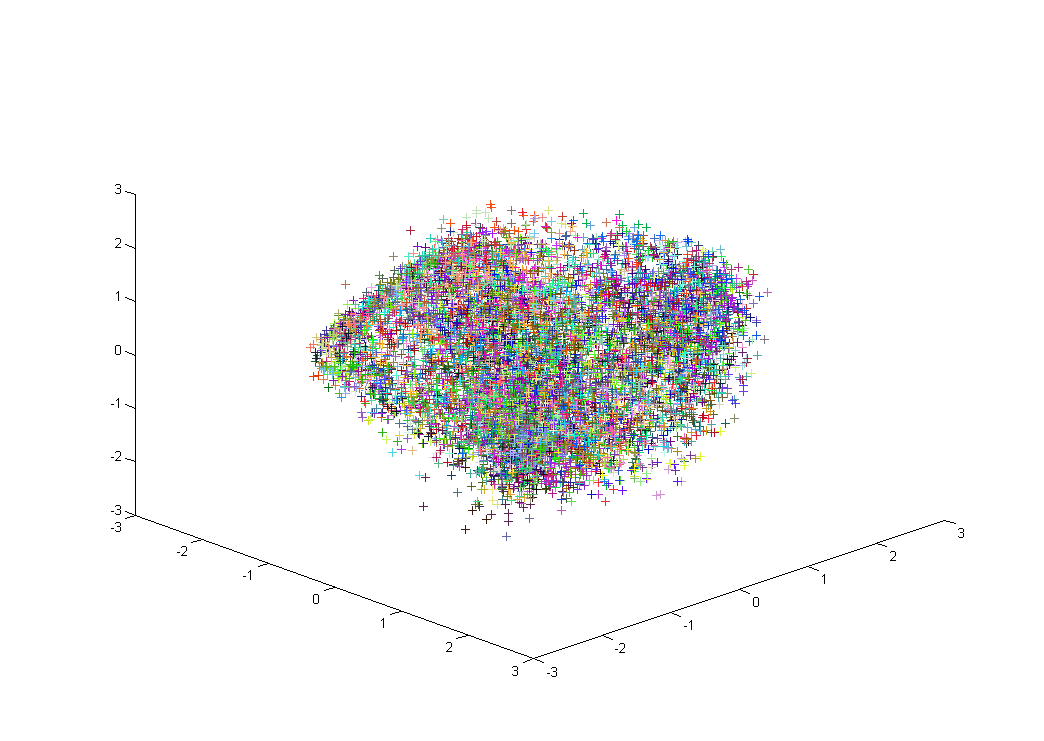}
    \vspace{-5pt}
    \caption{Initial stage (1762)}
    \vspace{20pt}
   \end{subfigure}
   \begin{subfigure}{0.33\linewidth}
   \centering
    \includegraphics[trim=6cm 0cm 3cm 4cm, clip=false, scale=0.2]{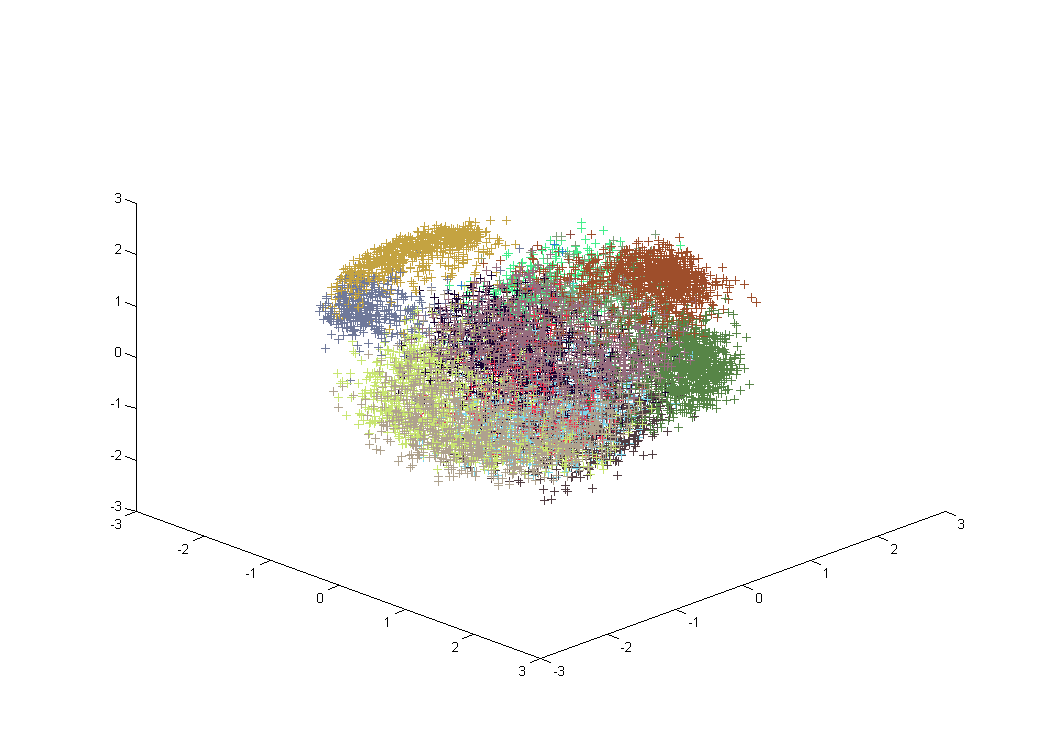}
    \vspace{-5pt}
    \caption{Middle stage (22)}
    \vspace{20pt}
   \end{subfigure}
   \begin{subfigure}{0.33\linewidth}
   \centering
       \includegraphics[trim=6cm 0cm 5cm 4cm, clip=false, scale=0.2]{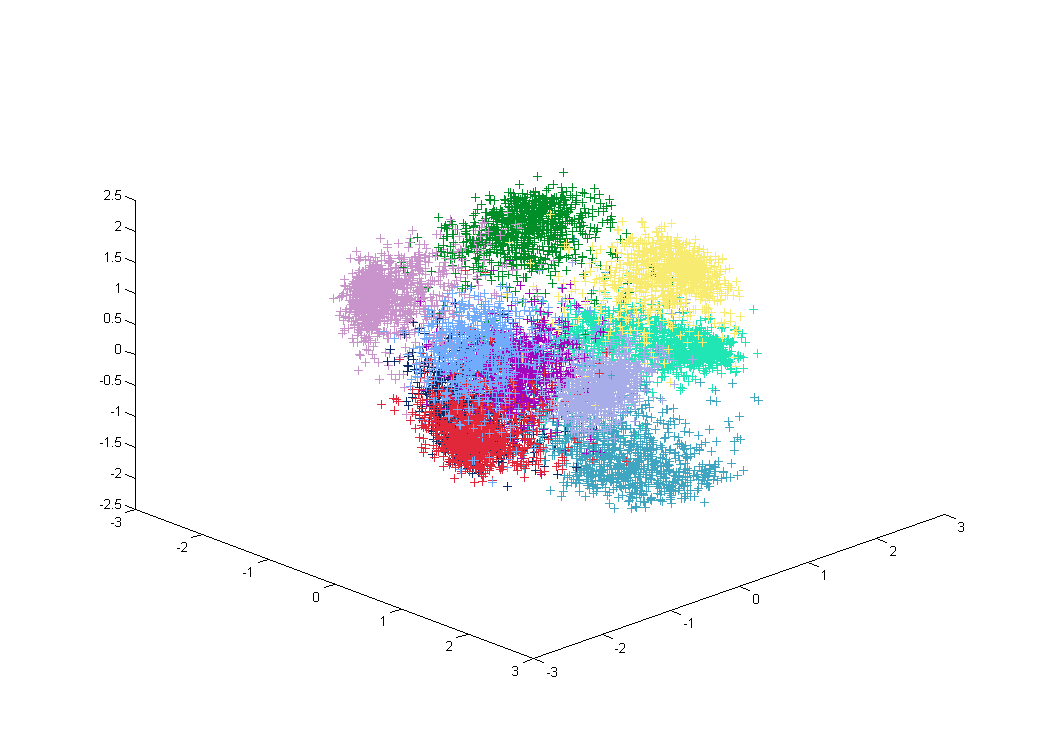}
   \vspace{-5pt}
   \caption{Final stage (10)}
   \vspace{20pt}
   \end{subfigure}

   \begin{subfigure}{0.33\linewidth}
   \centering
    \includegraphics[trim=6cm 0cm 5cm 4cm, clip=false, scale=0.2]{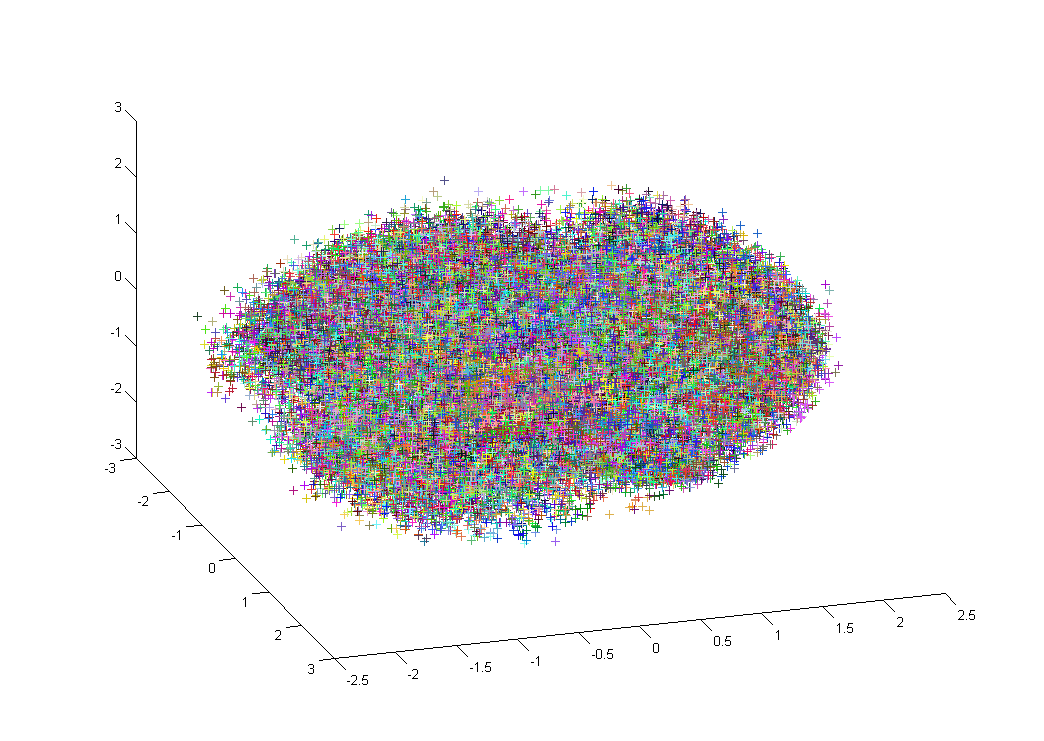}
    \vspace{-5pt}
    \caption{Initial stage (11521)}
       \vspace{20pt}
   \end{subfigure}
   \begin{subfigure}{0.33\linewidth}
   \centering
    \includegraphics[trim=6cm 0cm 3cm 4cm, clip=false, scale=0.2]{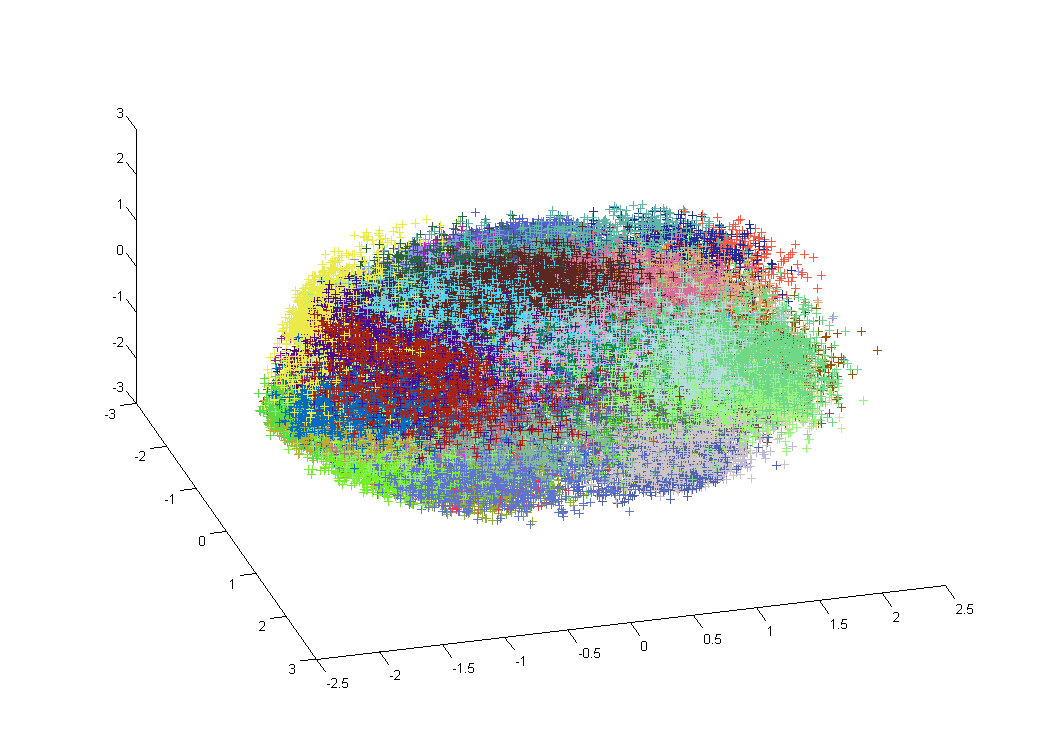}
    \vspace{-5pt}
    \caption{Middle stage (115)}
       \vspace{20pt}
   \end{subfigure}
   \begin{subfigure}{0.33\linewidth}
   \centering
       \includegraphics[trim=6cm 0cm 5cm 4cm, clip=false, scale=0.2]{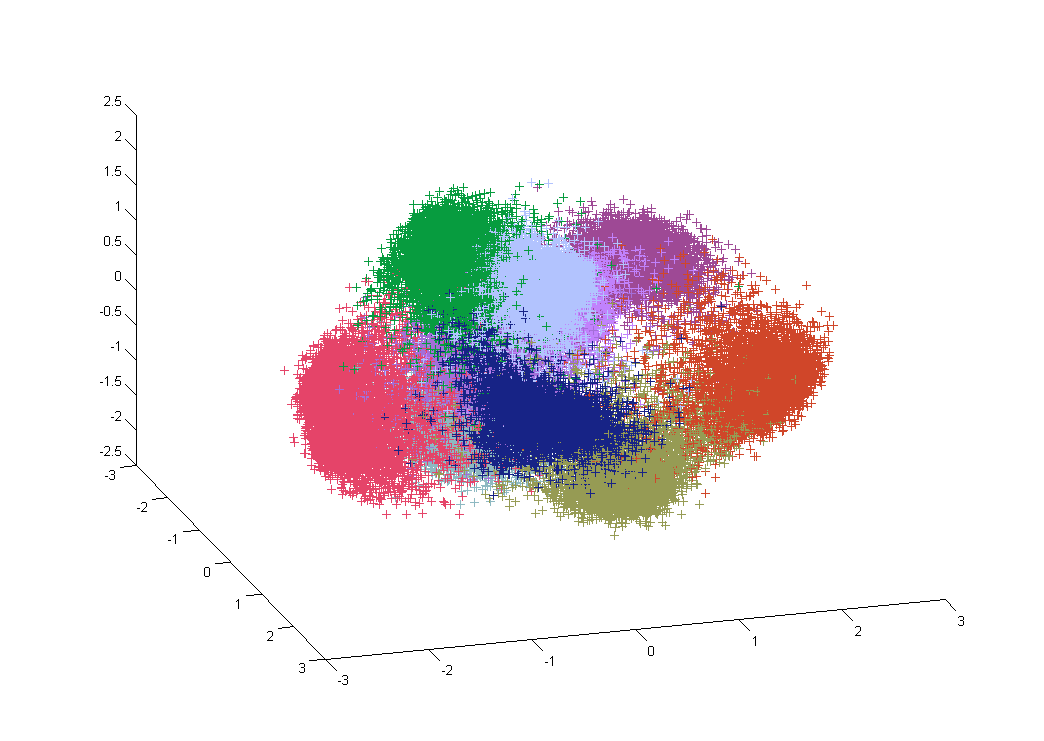}
   \vspace{-5pt}
   \caption{Final stage (10)}
      \vspace{20pt}
   \end{subfigure}
      \vspace{-5pt}
   \caption{Learned representations at different stages on five datasets. From top to bottom, they are \textit{COIL20}, \textit{COIL100}, \textit{USPS} and \textit{MNIST-test} and \textit{MNIST-full}. The first column are image intensities. For \textit{MNIST-test}, we show another view point different from Fig.1 in the main paper.}
   \label{Fig_PCA_Display_5}
\end{figure*}

\begin{figure*}
   \begin{subfigure}{0.33\linewidth}
   \centering
    \includegraphics[trim=6cm 0cm 5cm 4cm, clip=false, scale=0.2]{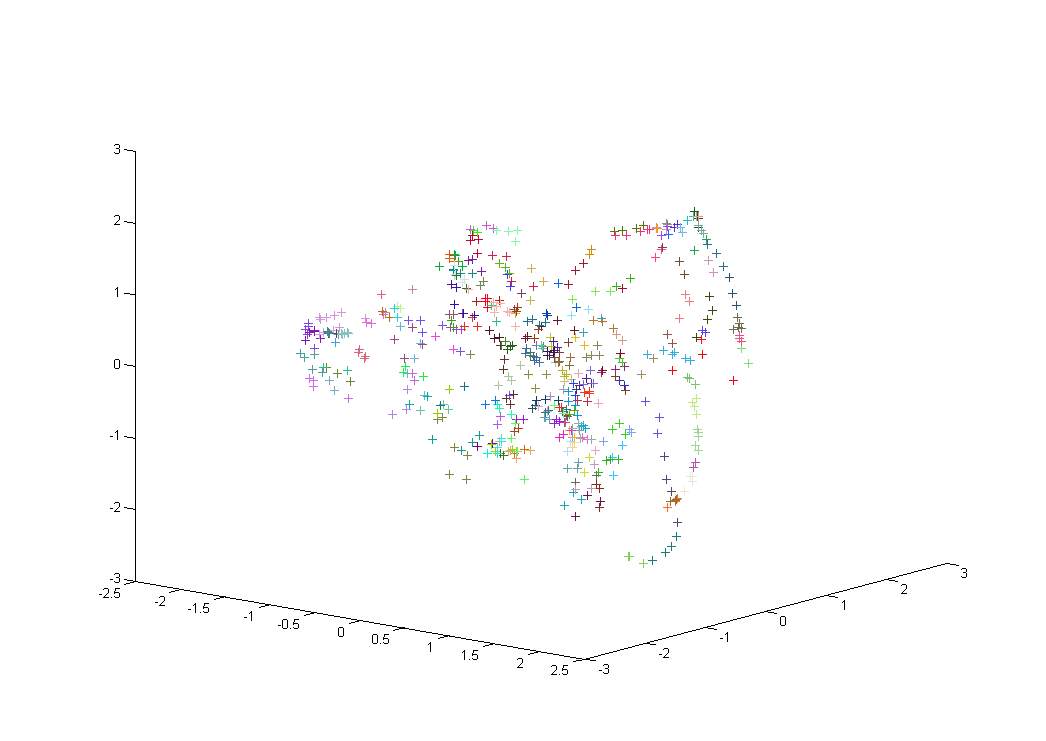}
    \vspace{-5pt}
    \caption{Initial stage (188)}
          \vspace{20pt}
   \end{subfigure}
   \begin{subfigure}{0.33\linewidth}
   \centering
    \includegraphics[trim=6cm 0cm 3cm 4cm, clip=false, scale=0.2]{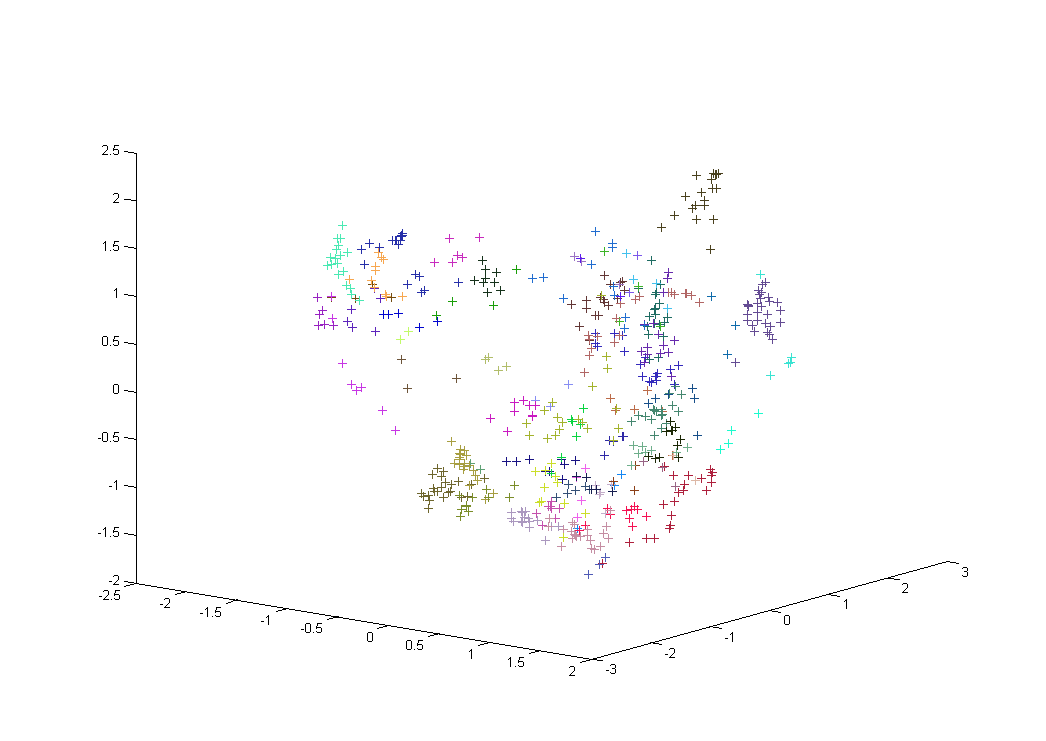}
    \vspace{-5pt}
    \caption{Middle stage (60)}
          \vspace{20pt}
   \end{subfigure}
   \begin{subfigure}{0.33\linewidth}
   \centering
       \includegraphics[trim=6cm 0cm 5cm 4cm, clip=false, scale=0.2]{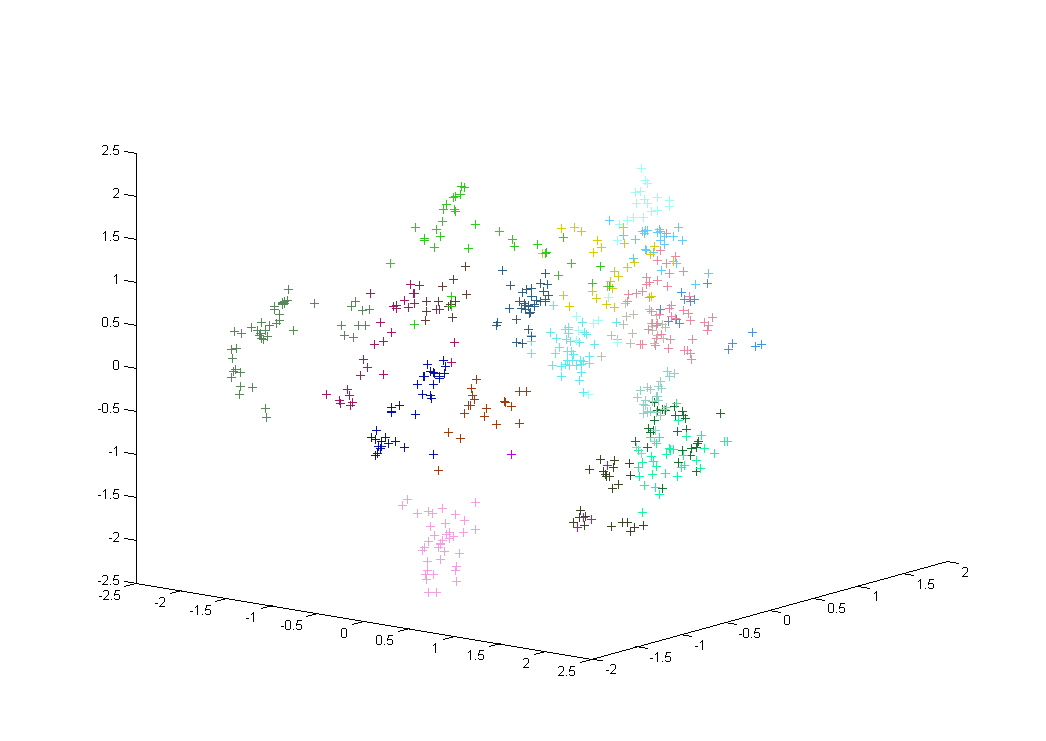}
   \vspace{-5pt}
   \caption{Final stage (20)}
         \vspace{20pt}
   \end{subfigure}

   \begin{subfigure}{0.33\linewidth}
   \centering
    \includegraphics[trim=6cm 0cm 5cm 4cm, clip=false, scale=0.2]{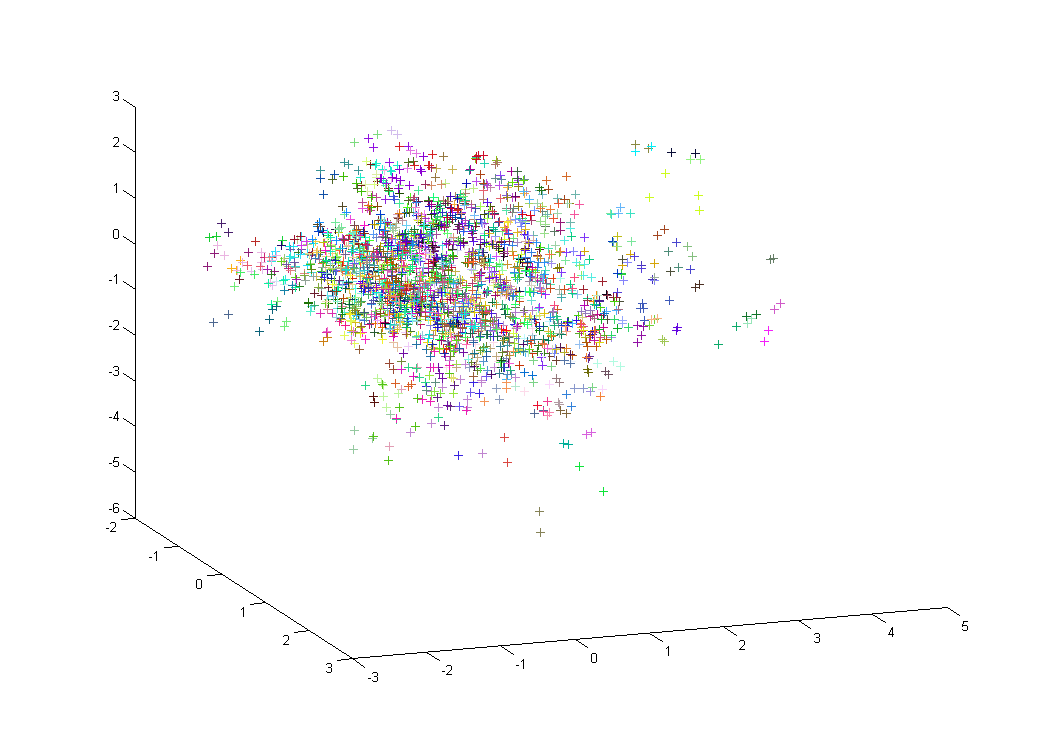}
    \vspace{-5pt}
    \caption{Initial stage (775)}
          \vspace{20pt}
   \end{subfigure}
   \begin{subfigure}{0.33\linewidth}
   \centering
    \includegraphics[trim=6cm 0cm 3cm 4cm, clip=false, scale=0.2]{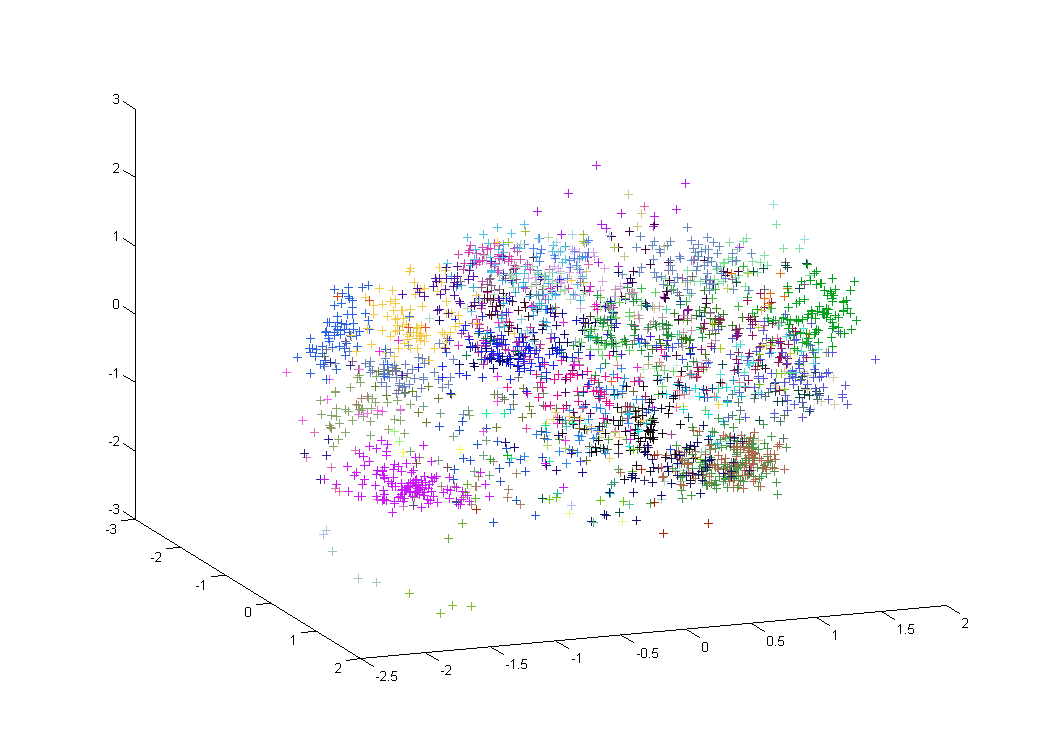}
    \vspace{-5pt}
    \caption{Middle stage (128)}
          \vspace{20pt}
   \end{subfigure}
   \begin{subfigure}{0.33\linewidth}
   \centering
       \includegraphics[trim=6cm 0cm 5cm 4cm, clip=false, scale=0.2]{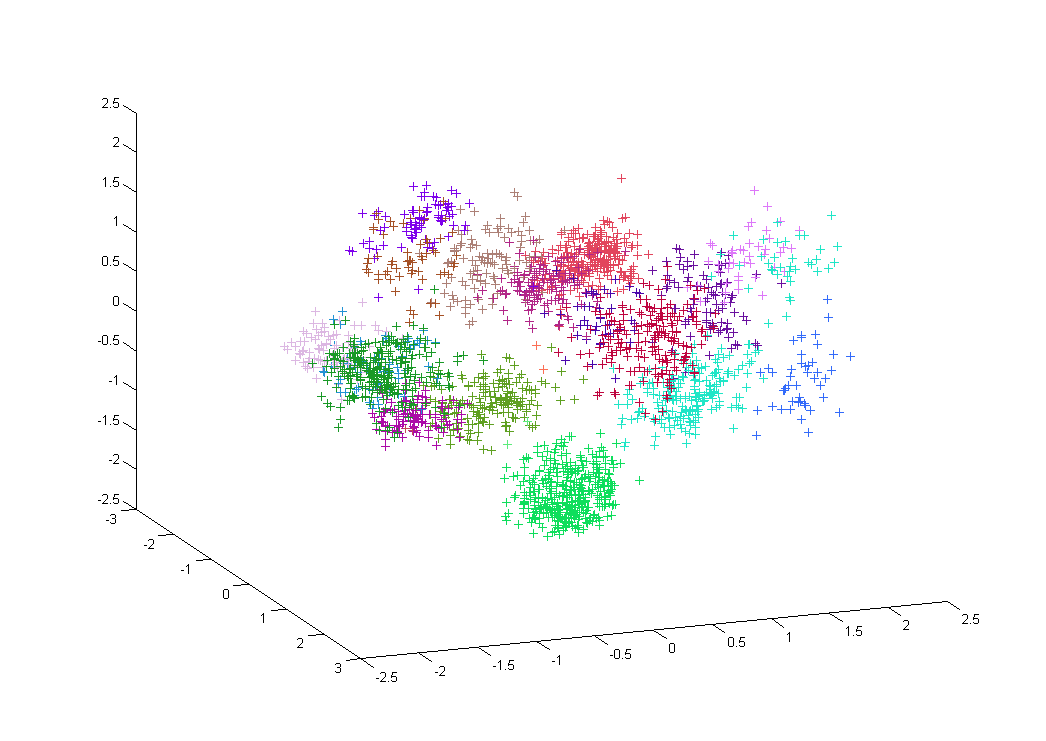}
   \vspace{-5pt}
   \caption{Final stage (20)}
         \vspace{20pt}
   \end{subfigure}

   \begin{subfigure}{0.33\linewidth}
   \centering
    \includegraphics[trim=6cm 0cm 5cm 4cm, clip=false, scale=0.2]{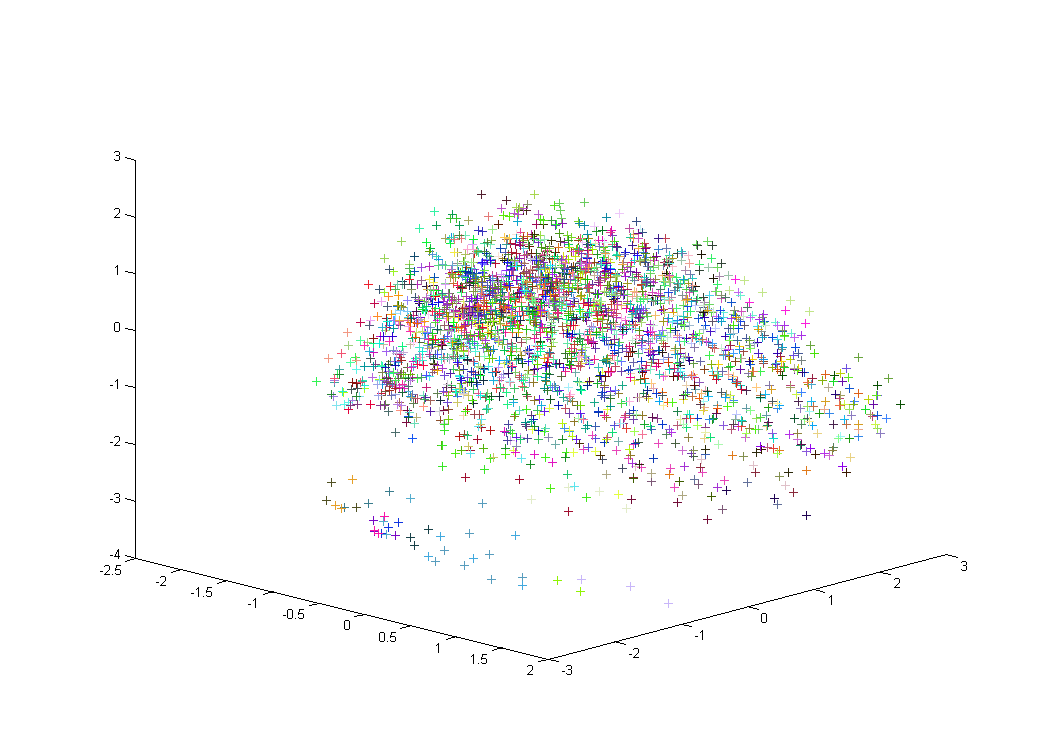}
    \vspace{-5pt}
    \caption{Initial stage (775)}
          \vspace{20pt}
   \end{subfigure}
   \begin{subfigure}{0.33\linewidth}
   \centering
    \includegraphics[trim=6cm 0cm 3cm 4cm, clip=false, scale=0.2]{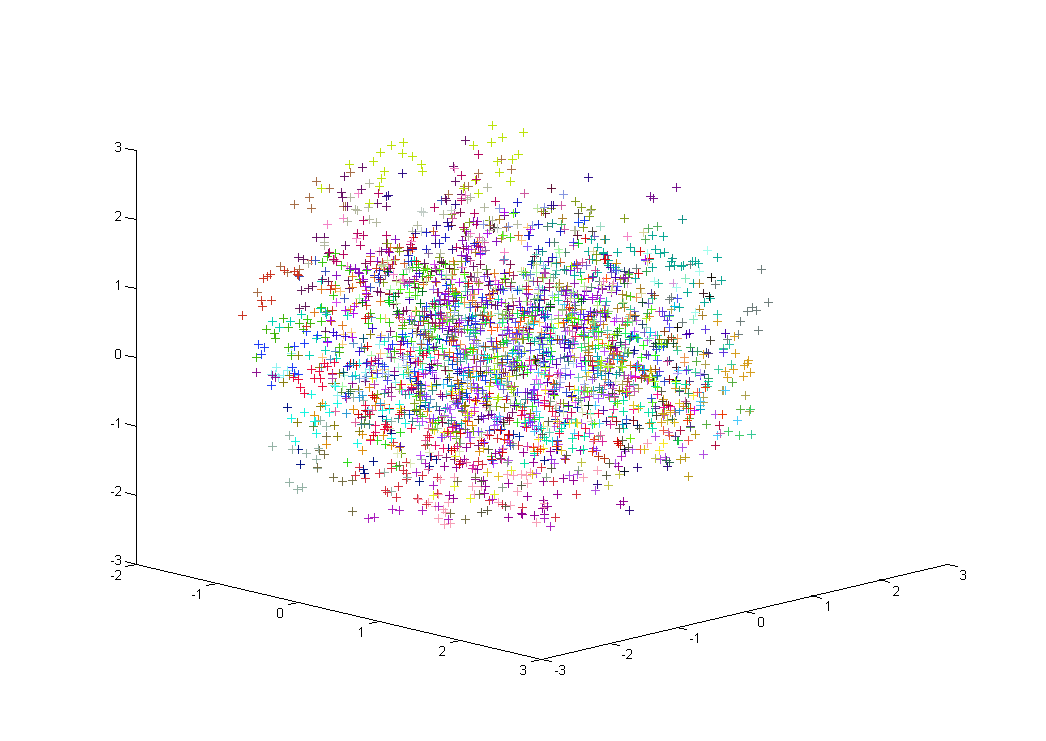}
    \vspace{-5pt}
    \caption{Middle stage (200)}
          \vspace{20pt}
   \end{subfigure}
   \begin{subfigure}{0.33\linewidth}
   \centering
       \includegraphics[trim=6cm 0cm 5cm 4cm, clip=false, scale=0.2]{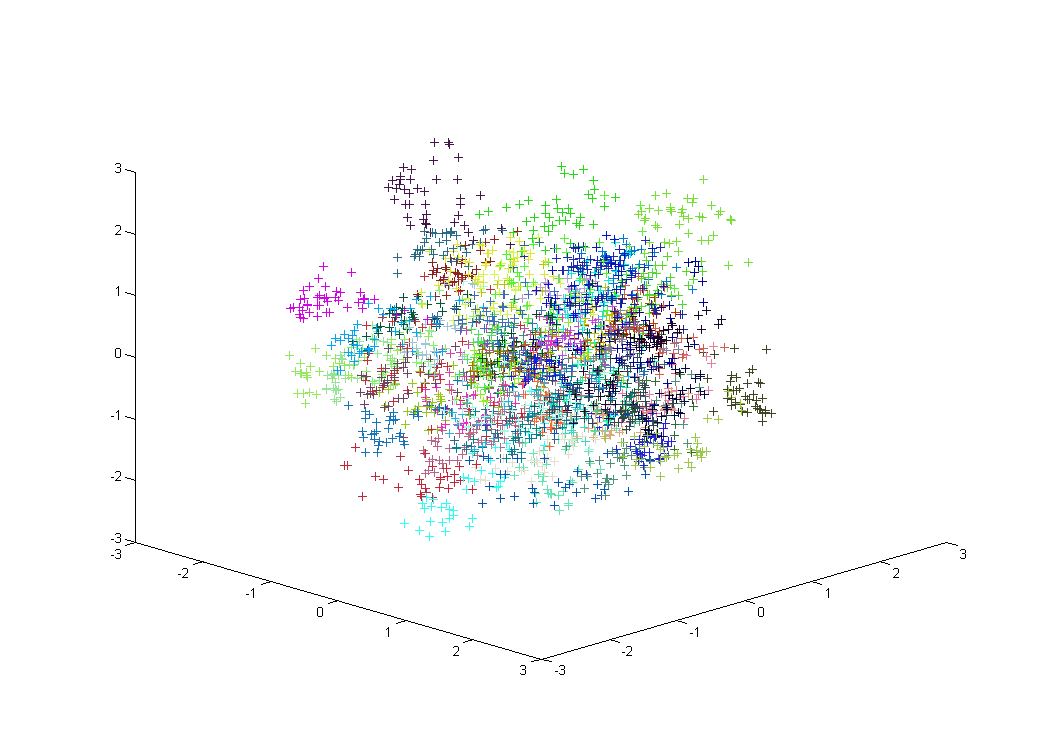}
   \vspace{-5pt}
   \caption{Final stage (68)}
         \vspace{20pt}
   \end{subfigure}

   \begin{subfigure}{0.33\linewidth}
   \centering
    \includegraphics[trim=6cm 0cm 5cm 4cm, clip=false, scale=0.2]{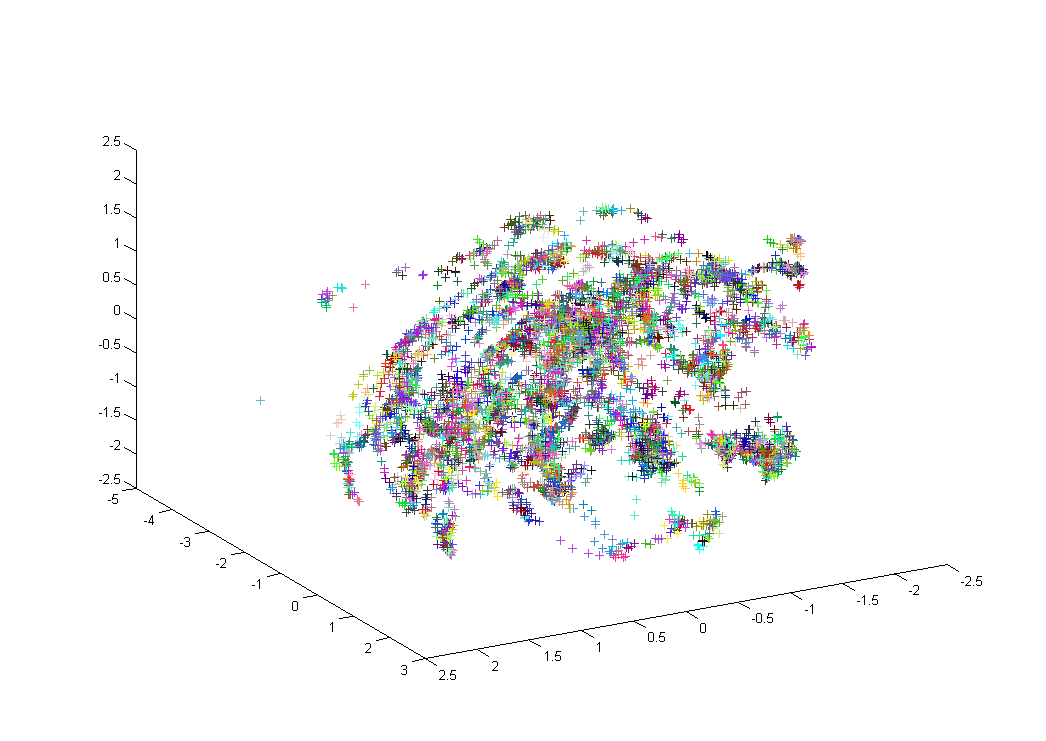}
    \vspace{-5pt}
    \caption{Initial stage (2814)}
          \vspace{20pt}
   \end{subfigure}
   \begin{subfigure}{0.33\linewidth}
   \centering
    \includegraphics[trim=6cm 0cm 3cm 4cm, clip=false, scale=0.2]{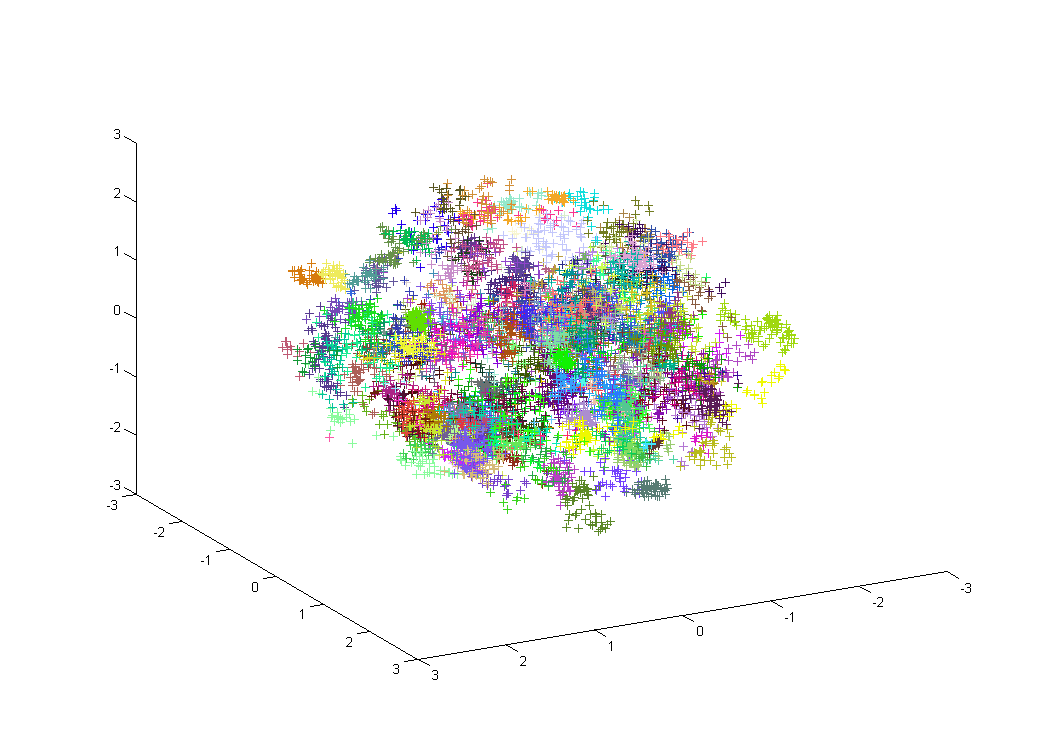}
    \vspace{-5pt}
    \caption{Middle stage (300)}
          \vspace{20pt}
   \end{subfigure}
   \begin{subfigure}{0.33\linewidth}
   \centering
       \includegraphics[trim=6cm 0cm 5cm 4cm, clip=false, scale=0.2]{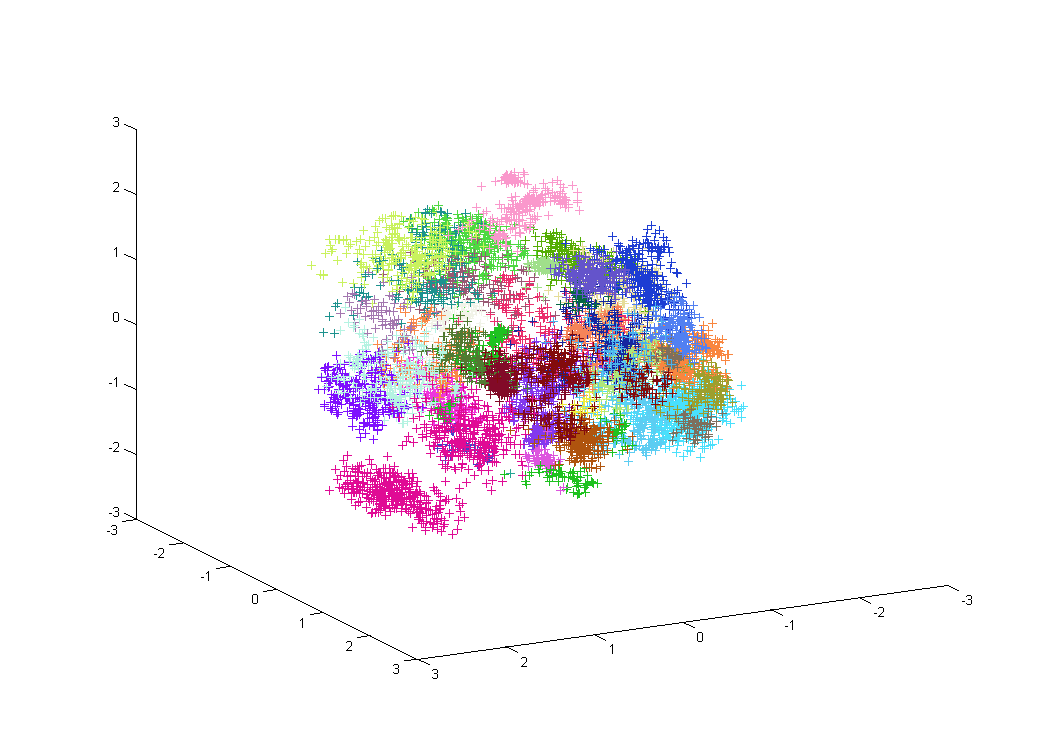}
   \vspace{-5pt}
   \caption{Final stage (41)}
      \vspace{20pt}
   \end{subfigure}
   \vspace{-5pt}
   \caption{Learned representations as different stages on four datasets. From top to bottom, they are \textit{UMist}, \textit{FRGC}, \textit{CMU-PIE} and \textit{YTF}. The first column are image intensities.}
   \label{Fig_PCA_Display_4}
\end{figure*}

We show the first three principle components of learned representations in Fig.~\ref{Fig_PCA_Display_5} and Fig.~\ref{Fig_PCA_Display_4} at different stages. For comparison, we show the image intensities at the first column. We use different colors for representing different clusters that we predict during the algorithm. At the bottom of each plot, we give the number of clusters at the corresponding stage. At the final stage, the number of cluster is same to the number of categories in the dataset. After a number of iterations, we can learn more discriminative representations for the datasets, and thus facilitate more precise clustering results.

\end{document}